%% file: main.tex
\documentclass[acmtog,nonacm]{acmart}

\acmSubmissionID{399}

\AtBeginDocument{%
  }

\setcopyright{none}
\acmDOI{XXXXXXX.XXXXXXX}

\usepackage{cleveref}
\usepackage{multirow}
\usepackage{subcaption}

\usepackage{graphicx}
\usepackage{tikz}
\usetikzlibrary{arrows.meta} 
\usepackage[abs]{overpic} %
\usepackage{xcolor}       %
\usetikzlibrary{shadows}

\usepackage[title]{appendix}

\usepackage{cleveref}
\usepackage{multirow}
\usepackage{subcaption}
\usepackage{mwe}

\usepackage{graphicx}
\usepackage{tikz}
\usetikzlibrary{arrows.meta} 
\usepackage[abs]{overpic} %
\usepackage{xcolor}       %
\usetikzlibrary{shadows}

\input{preamble}

\begin{document}

\title{Continuous Control of Editing Models via Adaptive-Origin Guidance}

\author{Alon Wolf}
\affiliation{%
  \institution{Tel Aviv University, Decart.ai}
  \city{Tel Aviv}
  \country{Israel}
}
\email{alonwolfy@gmail.com}

\author{Chen Katzir}
\affiliation{%
  \institution{Decart.ai}
  \city{Tel Aviv}
  \country{Israel}
}
\email{chen@decart.ai}

\author{Kfir Aberman}
\affiliation{%
  \institution{Decart.ai}
  \city{Tel Aviv}
  \country{Israel}
}
\email{kfiraberman@gmail.com}

\author{Or Patashnik}
\affiliation{%
  \institution{Tel Aviv University}
  \city{Tel Aviv}
  \country{Israel}
}
\email{orpatashnik@gmail.com}

\authorsaddresses{}

\renewcommand{\shortauthors}{Wolf et al.}

\input{figures/0_teaser/teaser}

\input{0_abstract}

\maketitle

\input{1_intro}

\input{2_related}

\input{3_method_v2}

\input{4_experiments_2}

\input{5_conclusion}

\begin{acks}
We would like to express our deep gratitude to Yiftach Edelstein for his invaluable insights and extensive feedback throughout the development of this work. We also thank Yoav Baron, Shelly Golan, Saar Huberman, and Rishubh Parihar for their early feedback and helpful suggestions. Finally, we are grateful to the entire Decart.ai team for their continued support and for providing a stimulating research environment.
\end{acks}

\bibliographystyle{ACM-Reference-Format}
\bibliography{main}

\clearpage
\input{more_results/additional_results1}

\input{figures/3_results-videos/results-videos}

\clearpage
\section*{\huge{Appendix}}
\vspace{2pt}
\appendix
\input{supp/details}
\input{supp/additional_results}

\end{document}

%% file: preamble.tex
\usepackage{xcolor}
\usepackage{amsmath}
\usepackage{enumitem}
\usepackage{multirow}
\usepackage{url}
\usepackage{hyperref}
\definecolor{darkgreen}{RGB}{0,100,0} 
\definecolor{linkgrayblue}{RGB}{70,90,110}
\definecolor{dustyrose}{RGB}{150,110,120}

\newif\ifdraft
\draftfalse
\ifdraft
\newcommand{\awc}[1]{{\color{blue}[\textbf{AW:} #1]}}
\newcommand{\kac}[1]{{\color{purple}[\textbf{KA:} #1]}}
\newcommand{\opc}[1]{{\color{red}[\textbf{OP:} #1]}}

\else
\newcommand{\awc}[1]{}
\newcommand{\kac}[1]{}
\newcommand{\opc}[1]{}

\fi

\newcommand{\boldparagraph}[1]{\paragraph{\textbf{#1}}}
\newcommand{\REC}{$\langle\texttt{id}\rangle$}

\newcommand{\LONGNAME}{\text{Adaptive Origin Guidance} }
\usepackage{cleveref}

%% file: figures/0_teaser/teaser.tex
\begin{teaserfigure}
  \includegraphics[width=\textwidth]{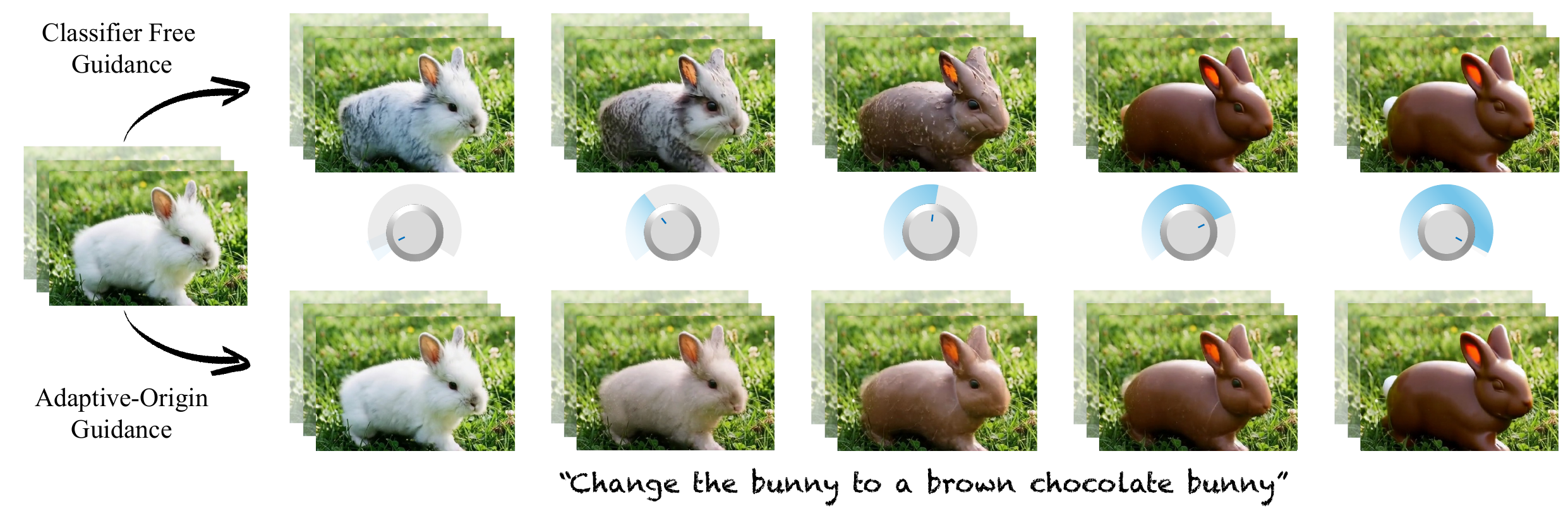}
  \vspace{-28pt}
  \caption{Given an input video (left) and a text instruction (“change the bunny to a brown chocolate bunny”), standard Classifier-Free Guidance (top row) does not recover the input content as the guidance scale decreases. Instead of a smooth transition back to the source, it yields arbitrary, inconsistent modifications (e.g., the gray bunny on the left).
  In contrast, our Adaptive-Origin Guidance (bottom row) enables a controllable gradual, semantically consistent progression editing from the original video to the fully edited result by adaptively interpolating the guidance origin. }
  \label{fig:teaser}
\end{teaserfigure}

%% file: 0_abstract.tex
\begin{abstract}

Diffusion-based editing models have emerged as a powerful tool for semantic image and video manipulation.
However, existing models lack a mechanism for smoothly controlling the intensity of text-guided edits.
In standard text-conditioned generation, Classifier-Free Guidance (CFG) impacts prompt adherence, suggesting it as a potential control for edit intensity in editing models.
However, we show that scaling CFG in these models does not produce a smooth transition between the input and the edited result.
We attribute this behavior to the unconditional prediction, which serves as the guidance origin and dominates the generation at low guidance scales, while representing an arbitrary manipulation of the input content.
To enable continuous control, we introduce Adaptive-Origin Guidance (AdaOr), a method that adjusts this standard guidance origin with an identity-conditioned adaptive origin, using an identity instruction corresponding to the identity manipulation.
By interpolating this identity prediction with the standard unconditional prediction according to the edit strength, we ensure a continuous transition from the input to the edited result.
We evaluate our method on image and video editing tasks, demonstrating that it provides smoother and more consistent control compared to current slider-based editing approaches.
Our method incorporates an identity instruction into the standard training framework, enabling fine-grained control at inference time without per-edit procedure or reliance on specialized datasets.
Additional results and videos are available at \href{https://adaor-paper.github.io/}{\color{dustyrose}{https://adaor-paper.github.io/}}.

\end{abstract}

%% file: 1_intro.tex
\vspace{-8pt}
\section{Introduction}

Recent years have witnessed significant progress in diffusion models for the creation and manipulation of visual content~\cite{ho2020denoising, song2022denoising, rombach2022highresolution, wan2025wanopenadvancedlargescale}. In particular, in the realm of image and video editing, these models enable complex semantic manipulations guided by natural language~\cite{hertz2022prompttoprompt, labs2025flux1kontextflowmatching, decart2025lucyedit}. However, while text prompts offer an intuitive interface for specifying what should be changed, they provide limited control over how much the change should be applied. 
For instance, when editing a portrait to add a beard, one may wish to smoothly vary the result from a clean-shaven appearance to light stubble and eventually to a full beard, rather than committing to a single, fixed outcome.
Supporting such gradual transitions between the source and the edited result, however, remains challenging for existing diffusion-based editing models.

While recent research has explored methods for achieving edit strength control, these approaches often rely on per-edit-type procedures~\cite{gandikota2025sliderspace, gandikota2023conceptslidersloraadaptors} or require extensive data collection~\cite{parihar2025kontinuous}. 
As a result, they tend to lack robustness across different inputs, edits, and model architectures, and their applicability to video editing remains largely unexplored.

In text-conditioned generation, Classifier-Free Guidance (CFG) is a crucial mechanism for ensuring visual quality and alignment between the prompt and the generated content~\cite{ho2022classifierfree}. By scaling the guidance, the model is forced to adhere more strictly to the conditioning signal, effectively increasing the influence of the prompt on the final output. 
Given the central role of CFG in modulating prompt influence in both text-to-image and text-to-video generation, a natural question is whether adjusting the CFG scale can provide smooth control over edit strength in diffusion-based editing models.
We show that this is not the case: in editing models, lowering the CFG scale does not produce a smooth transition back to the unedited input (see \Cref{fig:comparison_with_cfg,fig:teaser}).

The key observation of this work is that the limitation of CFG in controlling editing strength arises from the dominance of the unconditional prediction at low guidance scales and from the nature of this prediction in editing models.
We refer to this term as the \textit{guidance origin}.
In instruction-based editing settings, the unconditional prediction typically corresponds to an arbitrary manipulation of the input rather than faithful reconstruction. Consequently, when the guidance scale is varied, low guidance values do not induce small semantic changes around the input. 
Instead, the denoising process becomes dominated by the unconditional prediction serving as the guidance origin, leading to arbitrary deviations from the source content and steering the edit along uncontrolled directions.

To enable smooth control over edit strength in diffusion-based editing models, we introduce \textit{\LONGNAME}. We first propose an identity instruction, an instruction that corresponds to the identity manipulation: reproducing the input content without any semantic modification. Building on this, we introduce a guidance mechanism where the term that dominates the prediction at low scales (i.e., the origin) is adjusted according to the desired edit strength. Specifically, we interpolate between the identity prediction and the standard unconditional prediction. By assigning greater weight to the identity term at lower edit strengths and transitioning to the standard term at higher strengths, our method enables smooth, continuous control over manipulation intensity.

We evaluate our method on both image and video editing models, demonstrating its effectiveness across various architectures and manipulation tasks. Our results show that the proposed guidance strategy provides smooth, high-quality transitions that are significantly more consistent than standard CFG-based scaling. Furthermore, we compare our approach against existing slider-based editing methods~\cite{parihar2025kontinuous, gandikota2023conceptslidersloraadaptors, kamenetsky2025saedit}, demonstrating superior performance in balancing semantic change with structural preservation. 
Our guidance mechanism enables precise control over manipulation intensity without requiring per-edit optimization or the collection of specialized datasets.

%% file: 2_related.tex
\vspace{-4pt}
\section{Related Work}

\paragraph{\textbf{Instruction-driven Image and Video Editing}}

In recent years, the ability to generate and edit visual content using natural language has advanced rapidly~\cite{saharia2022photorealistic, nichol2022glide, ramesh2022hierarchical, patashnik2021styleclip}. In particular, progress in text-conditioned diffusion models has enabled high-quality semantic manipulations that previously required substantial manual effort~\cite{hertz2022prompttoprompt, meng2022sdedit, brooks2023instructpix2pix, avrahami2022blended, tumanyan2023plug}. Early works primarily focused on image editing, with a prominent line of research steering the denoising process in a training-free manner using feature injection~\cite{hertz2022prompttoprompt, tumanyan2023plug, patashnik2023localizing, alaluf2023crossimage, cao2023masactrl, mokady2022nulltext, garibi2024renoiserealimageinversion, Avrahami_2025_CVPR, parihar2025zero}, latent optimization~\cite{epstein2023selfguidance, parmar2023zeroshot}, or alternative sampling strategies~\cite{meng2022sdedit, hubermanspiegelglas2023edit, kulikov2025flowedit, rout2025semantic}. Several of these approaches have later been generalized to video editing~\cite{tokenflow2023, yatim2025dynvfxaugmentingrealvideos, ku2024anyv2v, wu2023tune, gao2025lora}.

\input{figures/1_cfgvsours/0_2_cfg_vs_ours}

More recently, instruction-driven diffusion models have demonstrated state-of-the-art performance in image editing~\cite{labs2025flux1kontextflowmatching, wu2025qwen, flux-2-2025, team2025zimage}. These models take as input an image and a text instruction, and produce a corresponding edited image~\cite{brooks2023instructpix2pix}. A central challenge in training such models lies in data collection, as paired examples of an image and its edited version are rarely available in the wild. As a result, training-free editing methods are often used to synthetically construct datasets for supervised training. With the rapid advancement of video generation models~\cite{wan2025wanopenadvancedlargescale, hacohen2024ltx, hacohen2026ltx, kong2024hunyuanvideo}, several video editing models have also been introduced~\cite{decart2025lucyedit, vace}, including Lucy~\cite{decart2025lucyedit}, an instruction-driven video editing model.

\paragraph{\textbf{Continuous Control for Image Editing}}
While natural language provides a form of semantic control that is difficult to achieve using other modalities, it is inherently coarse. As a result, achieving fine-grained control over generated or edited content using language alone is often challenging. In particular, continuous control over edit strength is difficult to express through text. To address this limitation, recent works introduce slider-based controls into generative models, enabling explicit manipulation of edit strength via a continuous parameter~\cite{gandikota2025sliderspace, gandikota2023conceptslidersloraadaptors, parihar2025kontinuous, dravid2024interpretingweightspacecustomized, parihar2024precisecontrol, li2023w-plus-adapter}.

Among these approaches, some rely on per-edit-type procedures, such as optimization or direction construction~\cite{gandikota2023conceptslidersloraadaptors, kamenetsky2025saedit, gandikota2025sliderspace}, which incur additional computational cost for each edit type. Other methods operate in the text embedding space, identifying editing directions and traversing them with varying step sizes corresponding to different edit strengths~\cite{dalva2024fluxspace, kamenetsky2025saedit, baumann2025attributecontrol}. Another line of work collects datasets containing edits at multiple strengths and trains models that are explicitly conditioned on the edit strength~\cite{cheng2024marble, lightlab@Magar2025, parihar2025kontinuous}. Since collecting such datasets is challenging, most of these methods focus on a narrow class of edits, such as material changes. Some works~\cite{parihar2025kontinuous, xu2025numerikontrol} construct datasets that are not restricted to a specific domain, but require substantial computational resources for data generation and model training. 
In contrast, our method does not rely on a custom dataset. As a result, it supports the diverse range of edit types already handled by the backbone editing model. 
Notably, existing approaches focus exclusively on image editing and do not address video.

Another related line of work is image morphing~\cite{zhang2024diffmorpher, cao2025freemorph}. These methods take two images as input and generate a continuous sequence that morphs between them. When combined with image editing, they can interpolate between the original and edited images. However, they often exhibit noticeable jumps, as they operate in latent spaces that lack sufficient continuity, and they remain largely underexplored in the video domain.

\vspace{-3pt}
\paragraph{\textbf{Guidance in Diffusion Models}}

To achieve high-fidelity alignment with the conditioning prompt, text-conditioned diffusion models rely on inference-time mechanisms to steer the generative process.
Early methods relied on external classifiers to guide the generation process using the gradient of the classifier with respect to the input to modify the diffusion model’s denoising prediction~\cite{dhariwal2021diffusion}. While effective, this approach introduces additional complexity by requiring external models at inference.

Classifier-Free Guidance (CFG)~\cite{ho2022classifierfree} removes the need for a separate classifier by using the diffusion model itself as an “implicit classifier”. This is achieved by modifying the training procedure so that the model is trained not only as a conditional generator, but also as an unconditional one. Specifically, unconditional behavior is learned by training the model to denoise arbitrary inputs when the null condition (typically the empty string, denoted as $\varnothing$) is provided. Having access to both conditional and unconditional predictions allows these outputs to be combined at inference time in a way that is equivalent to using a classifier, enabling the diffusion model to act as this implicit classifier.

While the original derivation of CFG was grounded in the gradient of the log-likelihood of an implicit classifier, follow-up works have increasingly analyzed this mechanism through different lenses, proposing various interpretations and improvements~\cite{yehezkel2025navigating, karras2024guiding, chung2025cfg, bradley2024classifier, katzir2024noisefree, hyung2025spatiotemporal}.

Recently, instruction-based editing models have become increasingly popular for both image and video editing~\cite{labs2025flux1kontextflowmatching, wu2025qwen, decart2025lucyedit}. These models rely on two conditioning signals: the input image or video to be edited and the edit instruction.
Similarly to text-conditioned models, these models also employ CFG to enforce high prompt adherence. 
While earlier approaches~\cite{brooks2023instructpix2pix} utilized dual guidance by independently dropping both the text and the input image, recent state-of-the-art editing models typically restrict guidance to the instruction alone.

%% file: figures/1_cfgvsours/0_2_cfg_vs_ours.tex
\begin{figure}[t]
    \centering
    \small
    \setlength{\tabcolsep}{0pt} %
    
    \begin{tabular}{cc ccccc}
        \multicolumn{7}{c}{\textit{``Add an elegant italian masquerade mask on the woman's eyes''}} \\
        
        \raisebox{35pt}{\rotatebox[origin=t]{90}{{CFG}}} & { } &
        \includegraphics[width=0.19\linewidth]{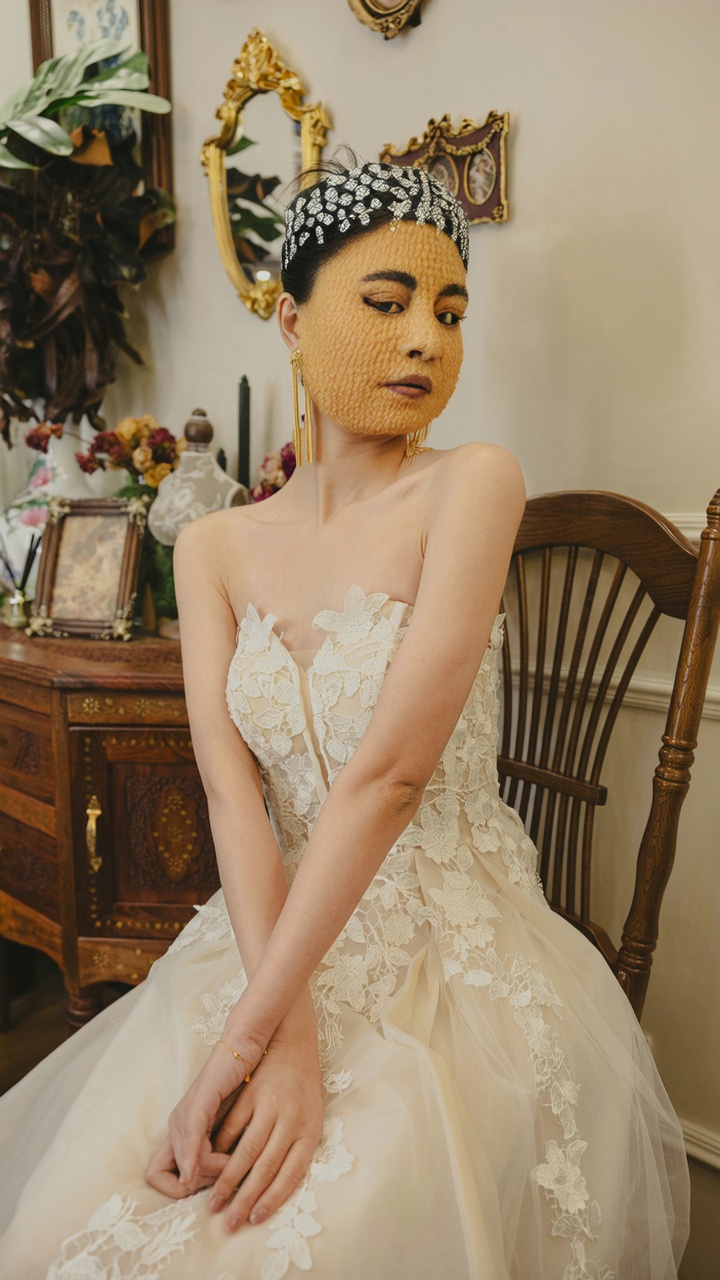} &
        \includegraphics[width=0.19\linewidth]{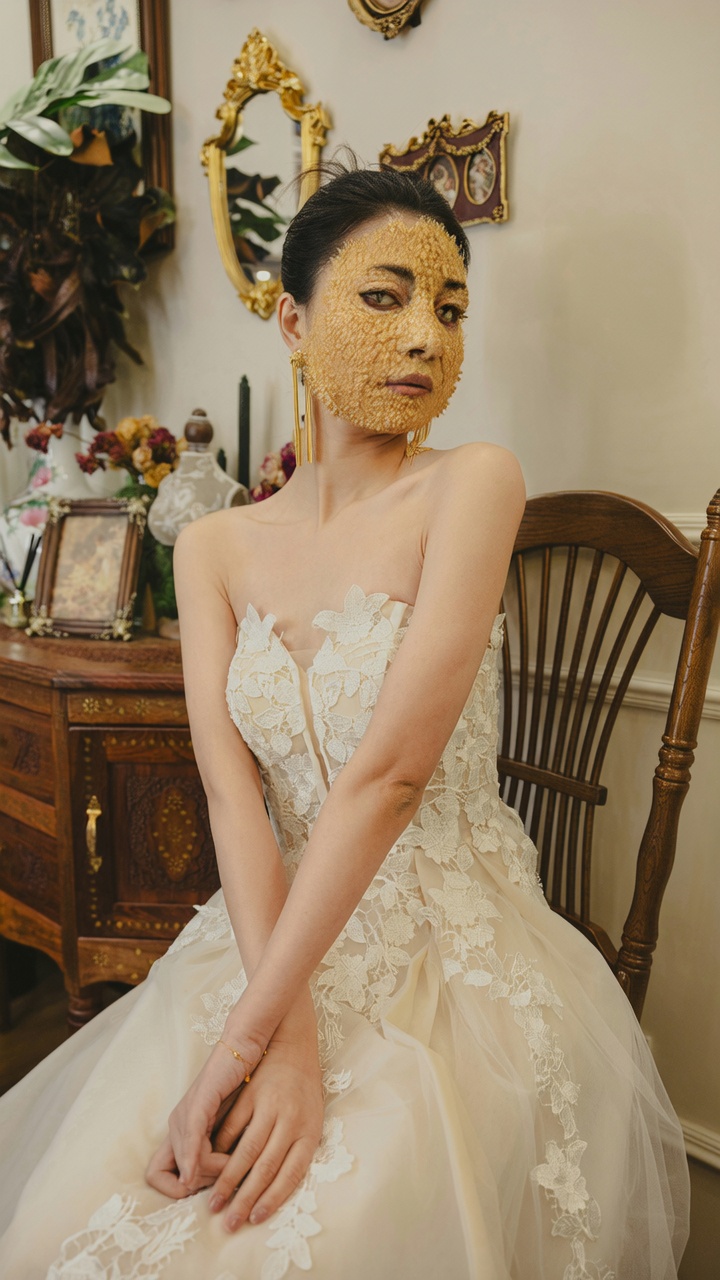} &
        \includegraphics[width=0.19\linewidth]{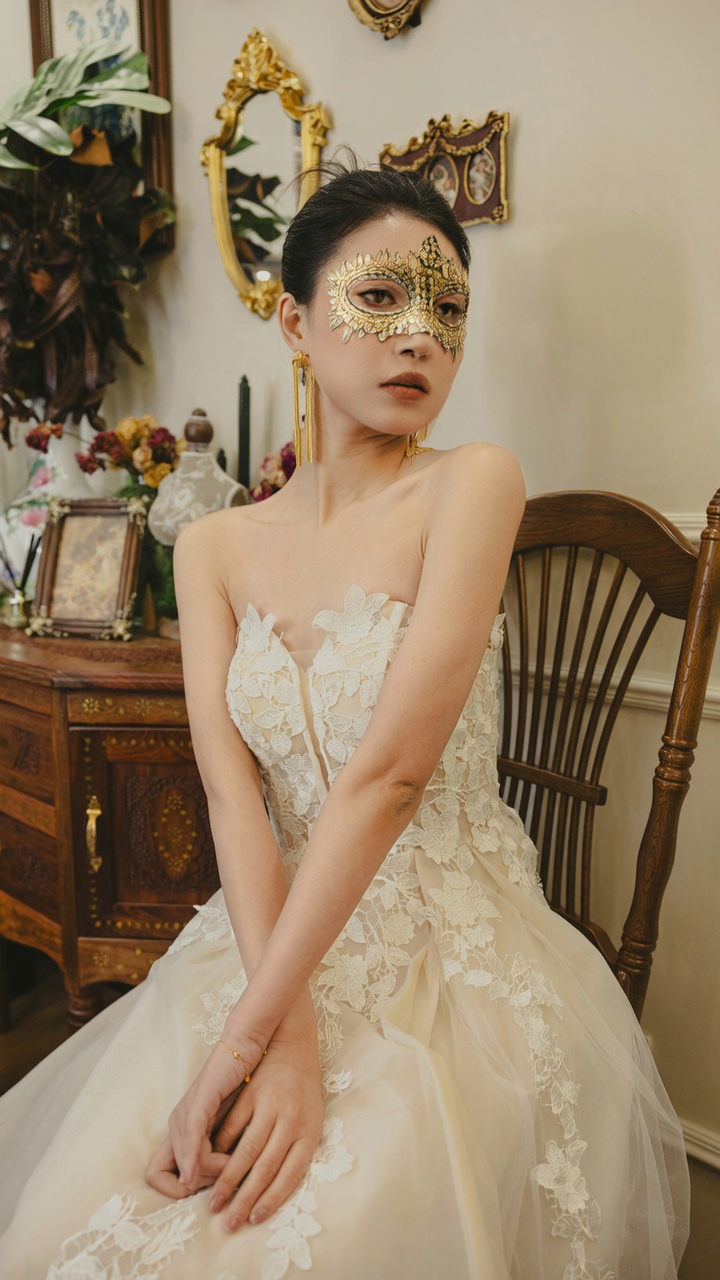} &
        \includegraphics[width=0.19\linewidth]{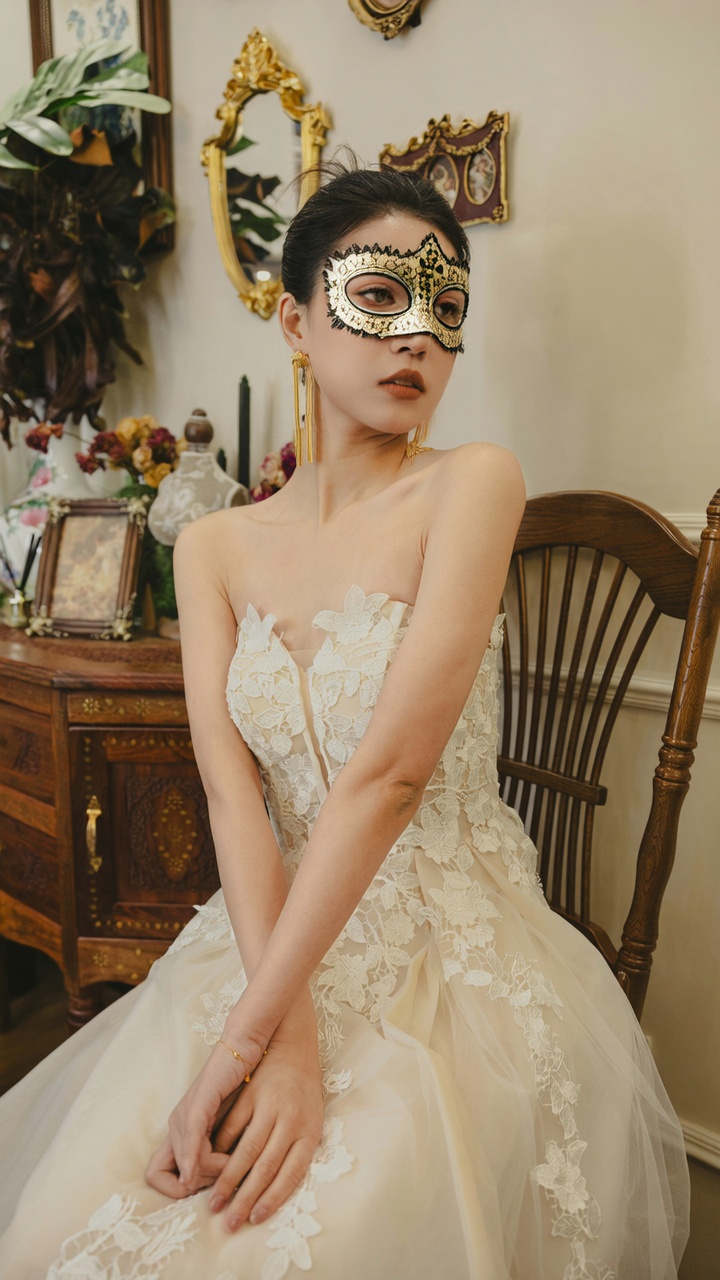} &
        \includegraphics[width=0.19\linewidth]{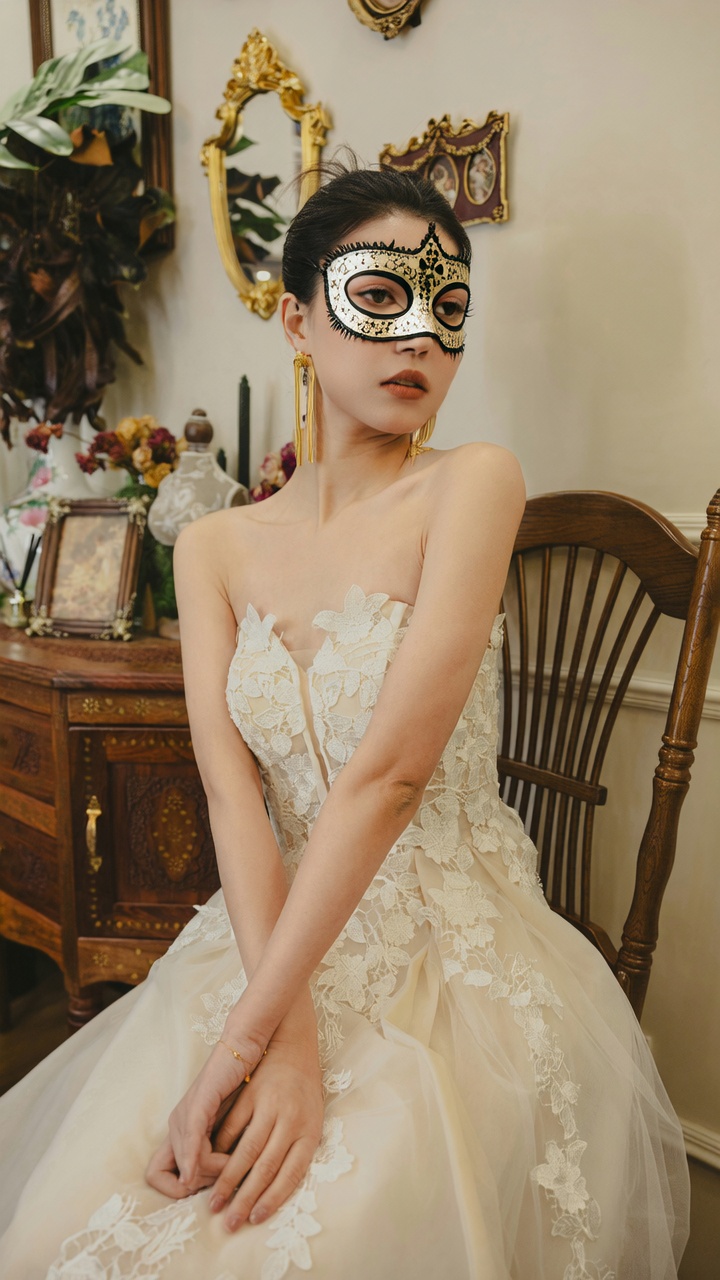} \\

        \raisebox{35pt}{\rotatebox[origin=t]{90}{{AdaOr (ours)}}} & { } &
        \includegraphics[width=0.19\linewidth]{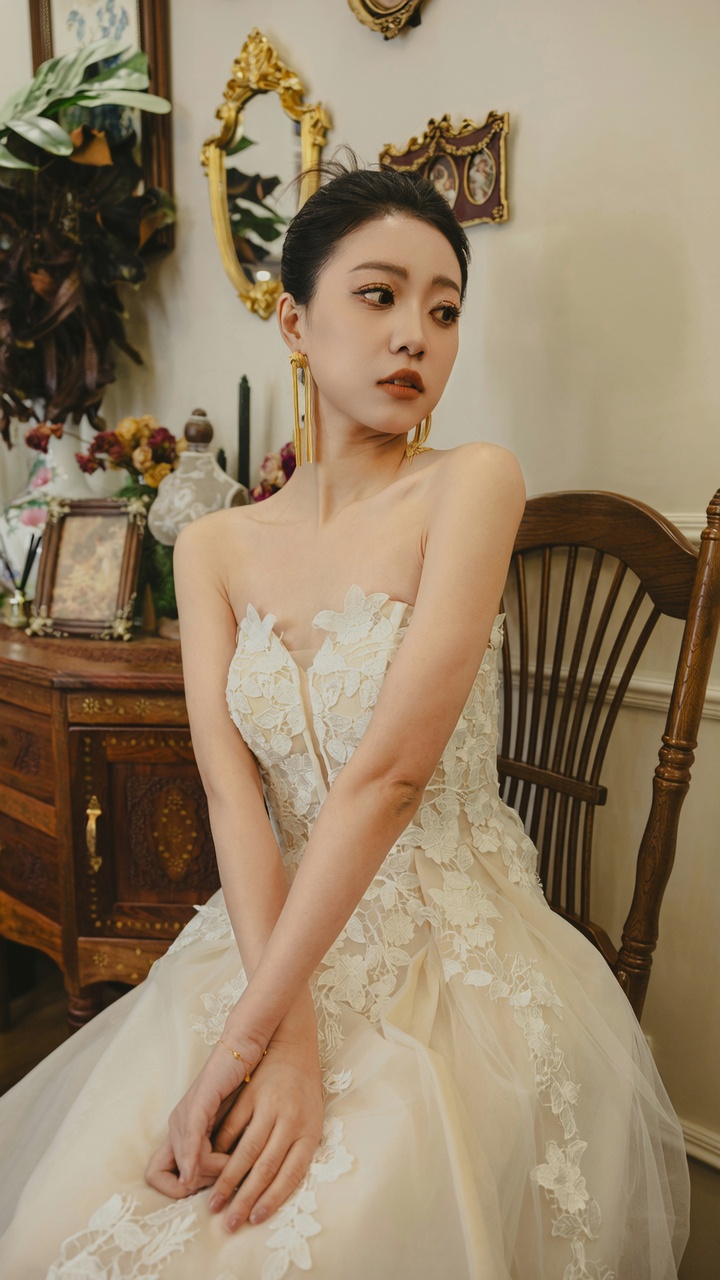} &
        \includegraphics[width=0.19\linewidth]{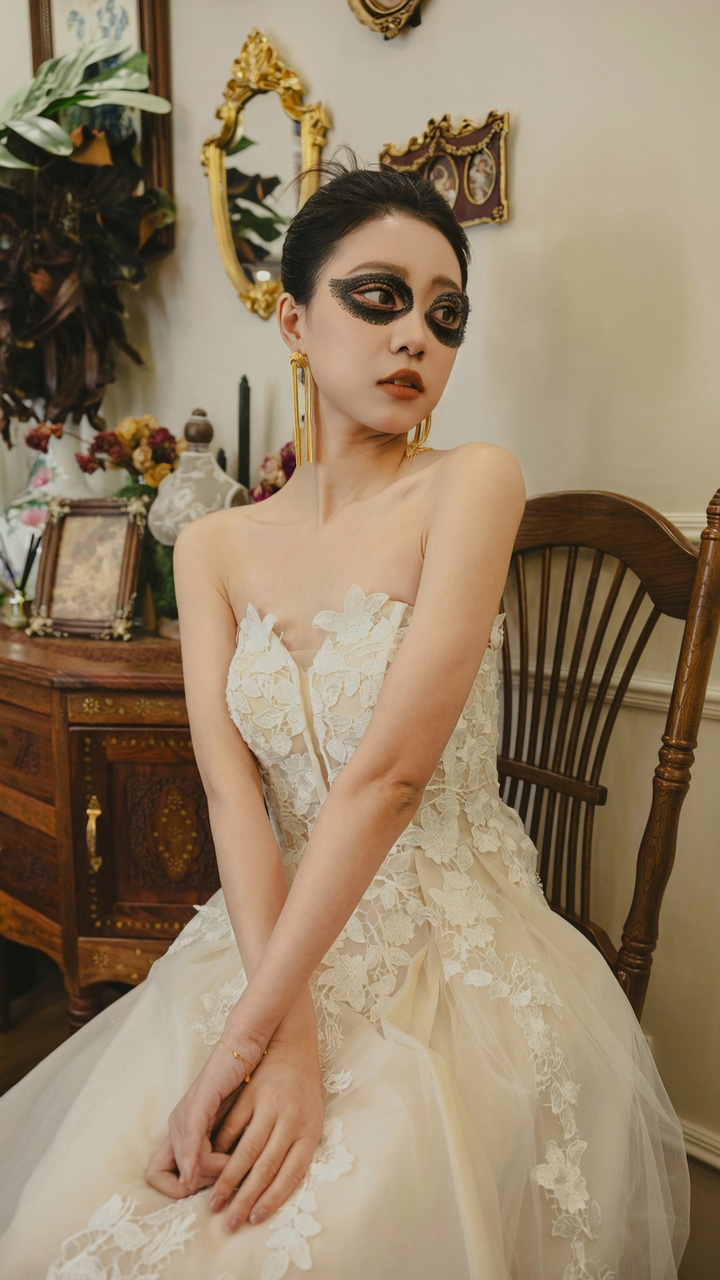} &
        \includegraphics[width=0.19\linewidth]{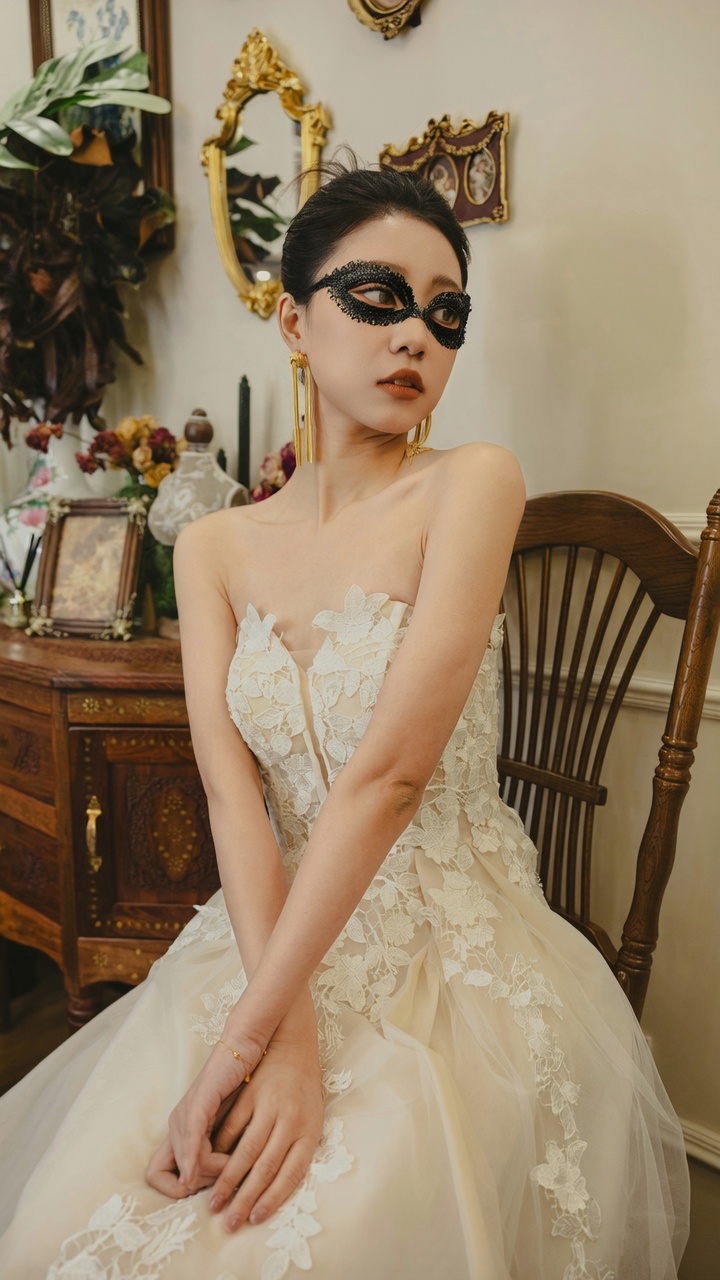} &
        \includegraphics[width=0.19\linewidth]{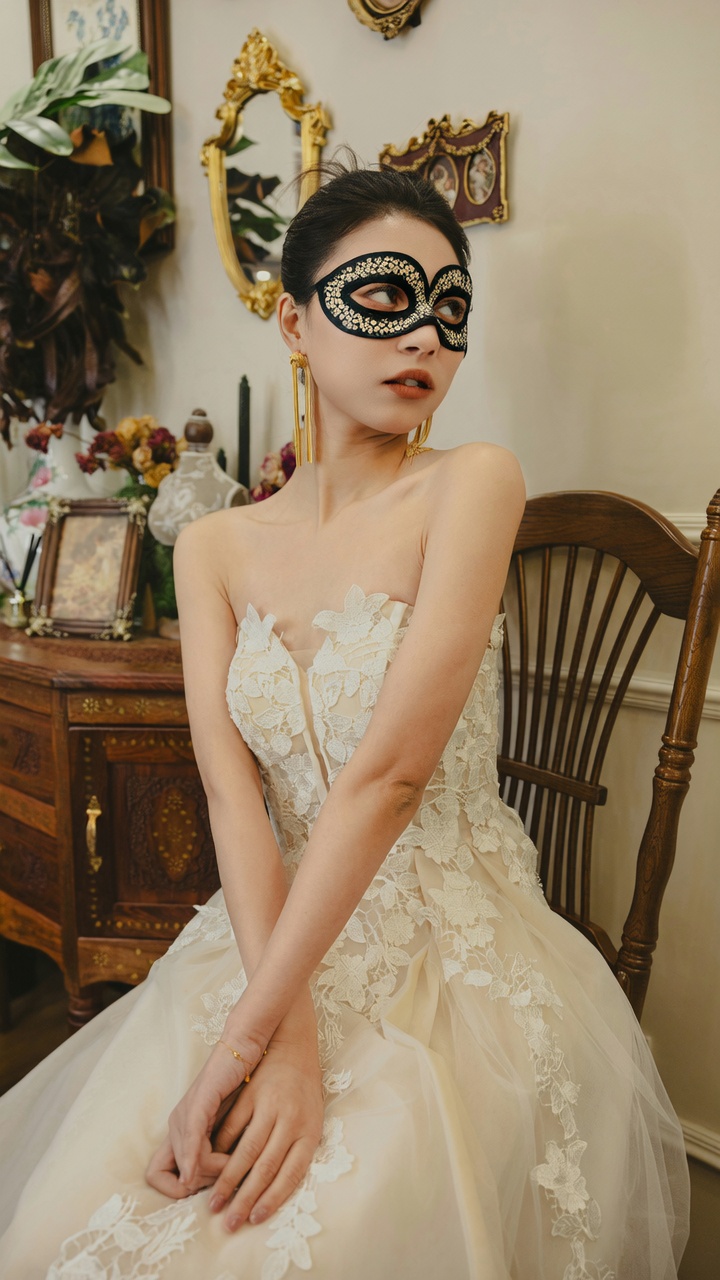} &
        \includegraphics[width=0.19\linewidth]{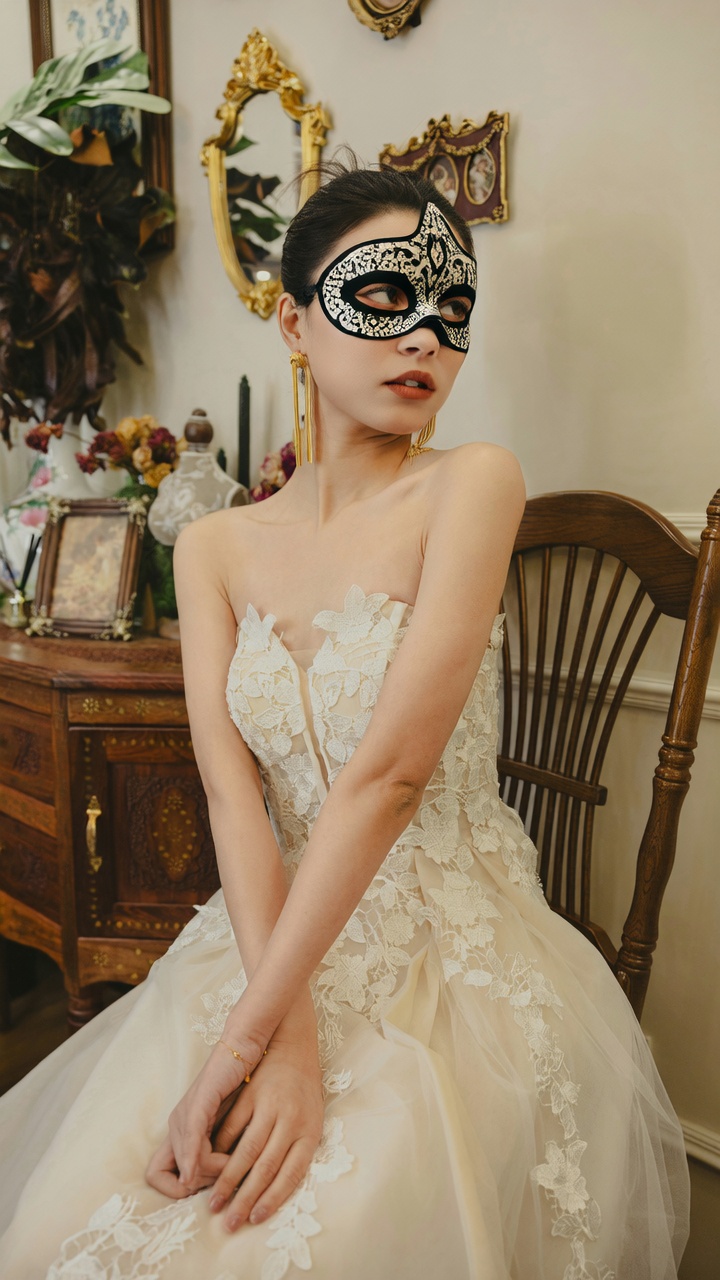} \\

        && \multicolumn{5}{l}{Edit Intensity $\xrightarrow{\hspace{180pt}}$} \\
        
    \end{tabular}
    \vspace{-8pt}
    \caption{\textbf{Comparison with standard CFG scaling.}
    We compare the progression of the edit as the intensity increases from 0 (left) to 1 (right).
    Standard CFG (top) exhibits arbitrary modifications at lower scales (e.g., the gold face paint).
    In contrast, our method (bottom) enables a smooth, continuous transition, gradually introducing the mask features while maintaining the structural integrity of the face throughout the interpolation.
    }
    \vspace{-12pt}
    \label{fig:comparison_with_cfg}
\end{figure}

%% file: 3_method_v2.tex
\section{Method}

\input{figures/1_method/method}

\subsection{Preliminaries}
\label{sec:preliminaries}

Classifier-free guidance (CFG) combines conditional and unconditional noise predictions to steer the generation process. Given a noisy latent variable $\mathbf{z}_t$ at diffusion timestep $t$, the guided noise prediction is defined as:
\begin{equation*}
    \epsilon^{w}(\mathbf{z}_t; c, t)
    \;=\;
    \epsilon(\mathbf{z}_t; \varnothing, t)
    \;+\;
    w \bigl(
        \epsilon(\mathbf{z}_t; c, t)
        -
        \epsilon(\mathbf{z}_t; \varnothing, t)
    \bigr),
\end{equation*}
where $\epsilon(\cdot)$ denotes the noise prediction network of the diffusion model, $c$ is the conditioning signal (e.g., a text prompt), and $\varnothing$ denotes the unconditional (null) input. The scalar $w \geq 0$ is the guidance scale controlling the strength of conditioning, with larger values encouraging closer adherence to $c$.

Geometrically, noise predictions can be viewed as vectors inducing transitions between successive noise distributions, namely, transitions from the marginal noise distribution $p_t$ at time $t$ to the slightly less noisy distribution $p_{t-1}$.
Under this view, CFG decomposes into an \textit{origin} given by the unconditional score, $\epsilon(\mathbf{z}_t; \varnothing, t)$, which moves the latent from $p_t$ toward the manifold of $p_{t-1}$, and a \textit{steering component} $\epsilon(\mathbf{z}_t;c,t)-\epsilon(\mathbf{z}_t;\varnothing,t)$ that acts on this manifold to bias the trajectory toward the conditional distribution, as illustrated in \Cref{fig:method1}.

For \textit{instruction-based editing models}, which are conditioned on an input image or video $c_I$ and an edit instruction $c_T$, CFG is commonly used to enforce prompt adherence. In practice, guidance is typically applied only to the instruction, and models are trained to handle a null instruction $\varnothing$ by randomly dropping $c_T$ during training while still denoising the target output conditioned on $c_I$. At inference, the guided prediction at timestep $t$ is:
\begin{equation*}
    \epsilon^w(\mathbf{z_t};c_I, c_T, t) =  \epsilon(\mathbf{z_t};c_I,\varnothing, t) + w (\epsilon(\mathbf{z_t};c_I, c_T, t) - \epsilon(\mathbf{z_t};c_I,\varnothing, t)),
\end{equation*}
with $w$ controlling the strength of instruction guidance.

\subsection{Adaptive-Origin Guidance}

Motivated by the fact that the guidance scale $w$ in CFG controls prompt adherence, our goal is to leverage it to control edit strength. However, as demonstrated in \Cref{fig:comparison_with_cfg,fig:teaser}, this approach does not work na\"ively.
We first observe that for small values of $w$, the noise prediction $\epsilon^w(\mathbf{z}_t; c_I, c_T, t)$ is dominated by the origin term $\epsilon(\mathbf{z}_t; c_I, \varnothing, t)$, rather than by the steering direction $\epsilon(\mathbf{z}_t; c_I, c_T, t) - \epsilon(\mathbf{z}_t; c_I, \varnothing, t)$. In other words, as $w \to 0$, the prediction collapses to the guidance origin.

This raises the question: what is the semantic meaning of the null prediction in an editing model? We observe that the null instruction $\varnothing$ effectively functions as an ``any edit'' command. This parallels the role of the null prompt in standard text-conditioned models, which represents the distribution of ``any natural image''. In the context of editing, however, since the model is conditioned on an input image, the null term corresponds to the marginal distribution of valid edits. As a result, it projects the input onto a generic manifold of edited images. 
We demonstrate such generic edits at small CFG scales in \Cref{fig:teaser}, where the bunny turns gray instead of brown, and in \Cref{fig:comparison_with_cfg}, where the face is painted gold and a hair decoration is added.

This observation explains why standard CFG fails to provide continuous control over edit strength. 
Intuitively, lowering the guidance scale ($w \to 0$) should reduce the edit magnitude and generate the input image.
However, because the standard origin represents an ``any edit'' point, it drives the output toward a generic manifold of edited images (see \Cref{fig:method1}). As a result, small values of $w$ produce arbitrary edits rather than weak ones, causing the output to lose input-specific structure in favor of generic dataset features.

To achieve continuous control over edit strength, we propose an adaptive-origin guidance that interpolates between a ``no-edit'' origin and the standard null origin, as shown in \Cref{fig:method2}. This design ensures that the edited image remains faithful to the input image at low edit strengths, while recovering the model's standard editing behavior at higher strengths.
To obtain this ``no-edit'' origin, we introduce a dedicated instruction token, \REC, which corresponds to the identity transformation. We then define the adaptive origin as a function of the edit strength $\alpha \in [0,1]$:
\begin{equation*}
    \mathcal{O}(\alpha) = s(\alpha) \epsilon(\mathbf{z_t};c_I,c_T=\varnothing, t) + (1-s(\alpha)) \epsilon(\mathbf{z_t};c_I,c_T=\text{\REC}, t),
\end{equation*}
where $s(\alpha)$ is a monotonically increasing scheduler satisfying $s(0) = 0$ and $s(1)=1$. Integrating this origin into the update step, our final guided prediction becomes:
\begin{equation*}
    \epsilon^{w, \alpha}(\mathbf{z_t};c_I, c_T, t) =  \mathcal{O}(\alpha) + \alpha \cdot w (\epsilon(\mathbf{z_t};c_I, c_T, t) - \epsilon(\mathbf{z_t};c_I,\varnothing, t)),
\end{equation*}
where $w$ is set as the standard CFG scale of the model. The resulting guidance term is illustrated in \Cref{fig:method2,fig:method3}.
We highlight the two boundary conditions of this formulation. First, at $\alpha = 0$, the prediction reduces to $\epsilon(\mathbf{z_t}; c_I, \text{\REC}, t)$, ensuring the input image remains intact, consistent with a zero-strength edit. Second, at $\alpha = 1$, the equation recovers the standard CFG formulation: $\epsilon(\mathbf{z_t}; c_I, \varnothing, t) + w(\epsilon(\mathbf{z_t}; c_I, c_T, t) - \epsilon(\mathbf{z_t}; c_I, \varnothing, t))$, thereby reproducing the model's default editing behavior.

\vspace{-3pt}
\boldparagraph{Learning the \REC{} Instruction}

Our formulation relies on the availability of a dedicated instruction that corresponds to the identity editing transformation. To acquire this, we introduce a new token, \REC, to the text encoder's vocabulary and incorporate it into the training of the editing model. Specifically, we train the model using the standard flow matching objective on a paired editing dataset, which we augment with identity pairs. For these identity samples, we provide identical source and target images paired with the \REC{} instruction, effectively teaching the model to associate this token with faithful reconstruction.

\vspace{-3pt}
\boldparagraph{Implementation Details}

We employ the Lucy-Edit~\cite{decart2025lucyedit} architecture as our backbone for both image and video editing, treating images as single-frame videos. To learn the identity instruction, we modify the standard training batch construction by introducing a stochastic mixing strategy. Given a source-target training pair $(I_{src}, I_{tgt})$ and a text instruction $T$, we construct the effective training triplet as follows: with a probability of $10\%$, we drop the text condition ($T = \varnothing$) to support unconditional prediction; with another $10\%$ probability, we align the \REC{} token with the identity mapping by setting the target equal to the source ($I_{tgt} = I_{src}$) and replacing the text instruction with the token \REC. In the remaining $80\%$ of cases, we use the standard editing triplet $(I_{src}, I_{tgt}, T)$. We train the full model using this strategy for 3,000 steps following the standard Lucy-Edit protocol.
We define the adaptive origin scheduler $s$ as $s(\alpha) = \sqrt{\alpha}$, and ablate this design choice in \Cref{sec:experiments}.

In Appendix \ref{app:results_with_qwen}, we further show continuous image editing results with Qwen-Image-Edit~\cite{wu2025qwen}.

\input{figures/2_othersvsours/others_vs_ours}

\subsection{\REC{} Editing Discussion}
\label{sec:rec-analysis}

Previous works have demonstrated that the null prediction in the CFG formulation can be replaced by alternative terms to improve guidance behavior, for example by using a weaker model prediction to increase diversity~\cite{karras2024guiding}.
This raises a question for continuous editing: can we replace the null prediction $\epsilon(\mathbf{z_t}; c_I, \varnothing, t)$ with the \REC{} prediction $\epsilon(\mathbf{z_t}; c_I, \text{\REC}, t)$, that is, use it both as the guidance origin and in the steering direction?
In this section, we show that this seemingly straightforward replacement leads to fundamental instabilities and is therefore unsuitable for continuous editing.
In \Cref{sec:experiments}, we further validate this finding empirically.

We observe that at the final denoising step ($t=0$), the conditional distribution $p_t(\mathbf{z}\mid c_I, \text{\REC})$ collapses to the input image $c_I$, with all probability mass concentrated at that point. Formally, this corresponds to a Dirac delta distribution, $\delta(\mathbf{z} - c_I)$. At earlier timesteps $t>0$, since the latent is corrupted by Gaussian noise with variance $\sigma_t^2$, this point mass corresponds to a Gaussian distribution centered at $c_I$:
\vspace{-5pt}
\begin{equation*}
\vspace{-2pt}
p_t(\mathbf{z_t}\mid c_I,\text{\REC}) \propto \exp\left(-\frac{\lVert \mathbf{z_t} - c_I\rVert^2}{2\sigma_t^2}\right).
\end{equation*}
Recall that the noise prediction of a diffusion model is proportional to the scaled score function, $-\sigma_t \nabla_{\mathbf{z_t}} \log p_t(\mathbf{z_t})$. Applying this relation to the Gaussian distribution induced by \REC{}, we obtain:
\begin{equation*}
\vspace{-2pt}
\label{eq:rec_analytical_pred}
\epsilon(\mathbf{z_t}; c_I, \text{\REC}, t) \approx \frac{\mathbf{z_t} - c_I}{\sigma_t}.
\vspace{-2pt}
\end{equation*}
This relationship exposes an inherent instability in the \REC{} prediction. 
The term corresponds to a score that encourages $\mathbf{z_t}$ to move toward the input $c_I$. This behavior is appropriate when the desired edit strength is low and the latent remains close to the input $c_I$.
However, as the edit strength increases and $\mathbf{z_t}$ deviates from the input toward a new concept, the numerator becomes non-zero. As the timestep approaches zero, the noise level $\sigma_t$ vanishes, causing the magnitude of the \REC{} prediction to diverge to infinity. When used in the CFG steering term, this effect is further amplified by the guidance scale, leading to instability near the end of the denoising process for stronger edits.

In contrast, our adaptive origin guidance avoids this behavior by explicitly conditioning on edit strength. When the desired edit strength is low and the latent remains close to the input $c_I$, we rely on the \REC{} term as the guidance origin. As the edit strength increases and $\mathbf{z_t}$ departs from the input, our schedule gradually transitions the origin to the standard null prediction. This null prediction models the broader manifold of natural images and remains well behaved near the end of the denoising process, thereby preventing the guidance explosion.

%% file: figures/1_method/method.tex
\begin{figure*}[t!]
    \centering
    \begin{subfigure}[b]{0.32\linewidth}
        \centering
        \includegraphics[width=\linewidth]{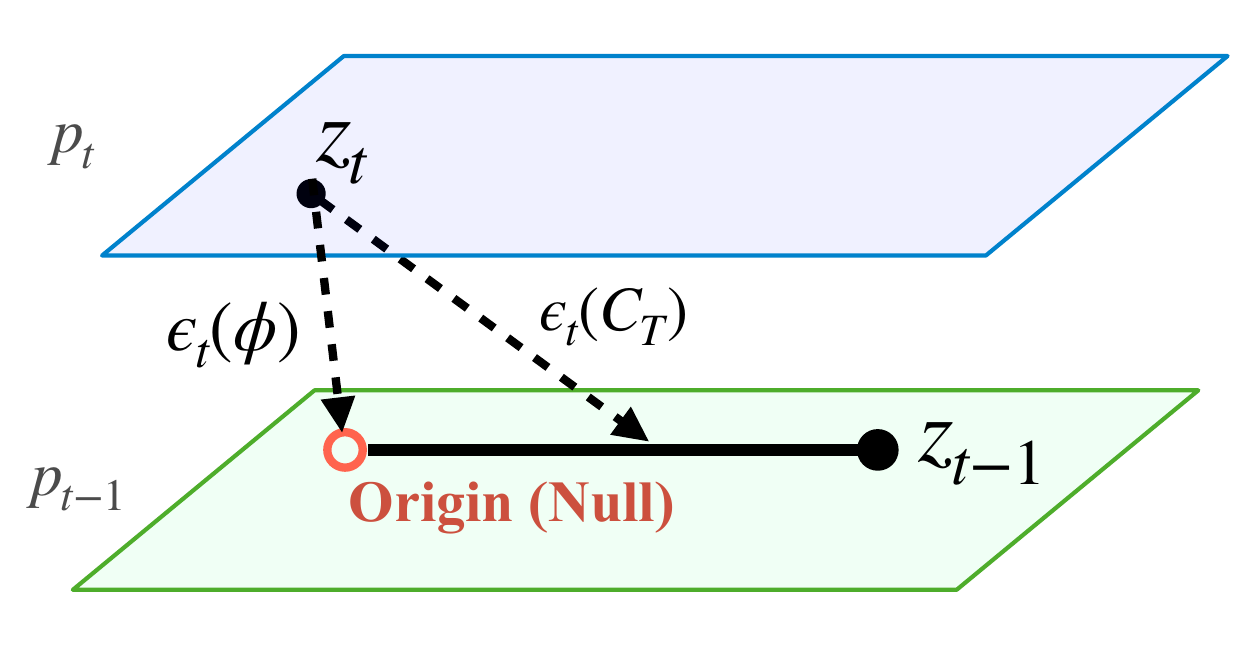}
        \caption{\textbf{Standard CFG}:\\ Null-condition as origin}
        \label{fig:method1}
    \end{subfigure}%
    \hfill
    \begin{subfigure}[b]{0.32\linewidth}
        \centering
        \includegraphics[width=\linewidth]{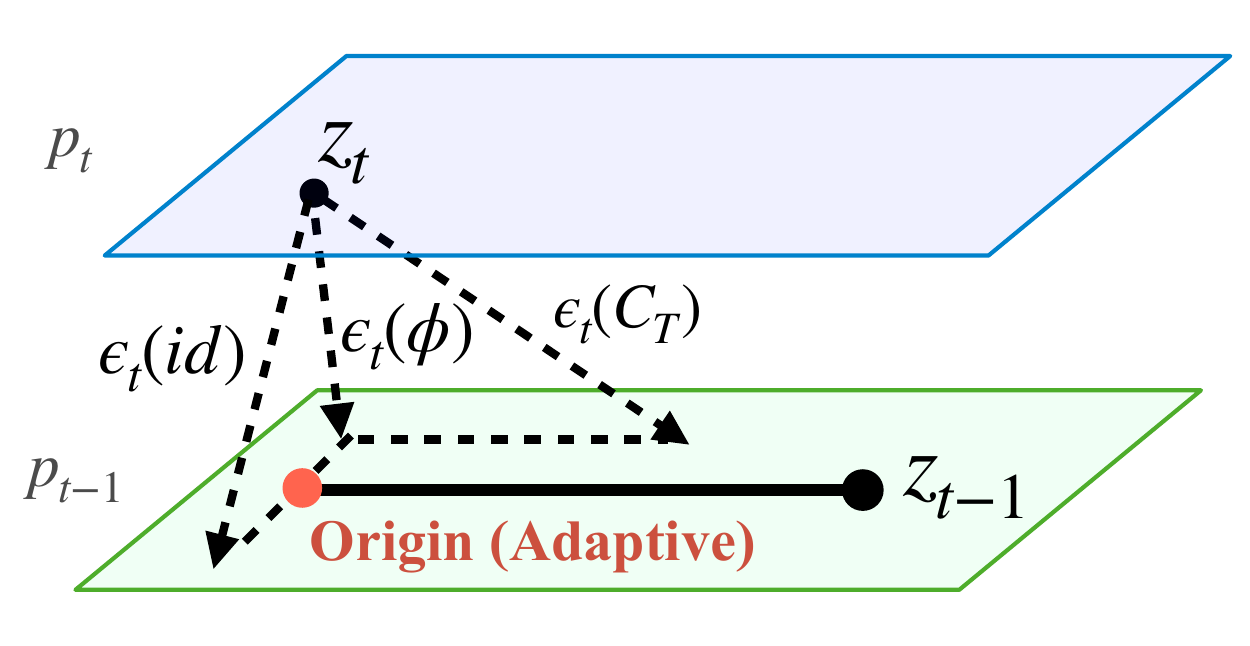}
        \caption{\centering \textbf{Adaptive Origin}:\\ Null-identity interpolated origin}
        \label{fig:method2}
    \end{subfigure}%
    \hfill
    \begin{subfigure}[b]{0.32\linewidth}
        \centering
        \includegraphics[width=\linewidth]{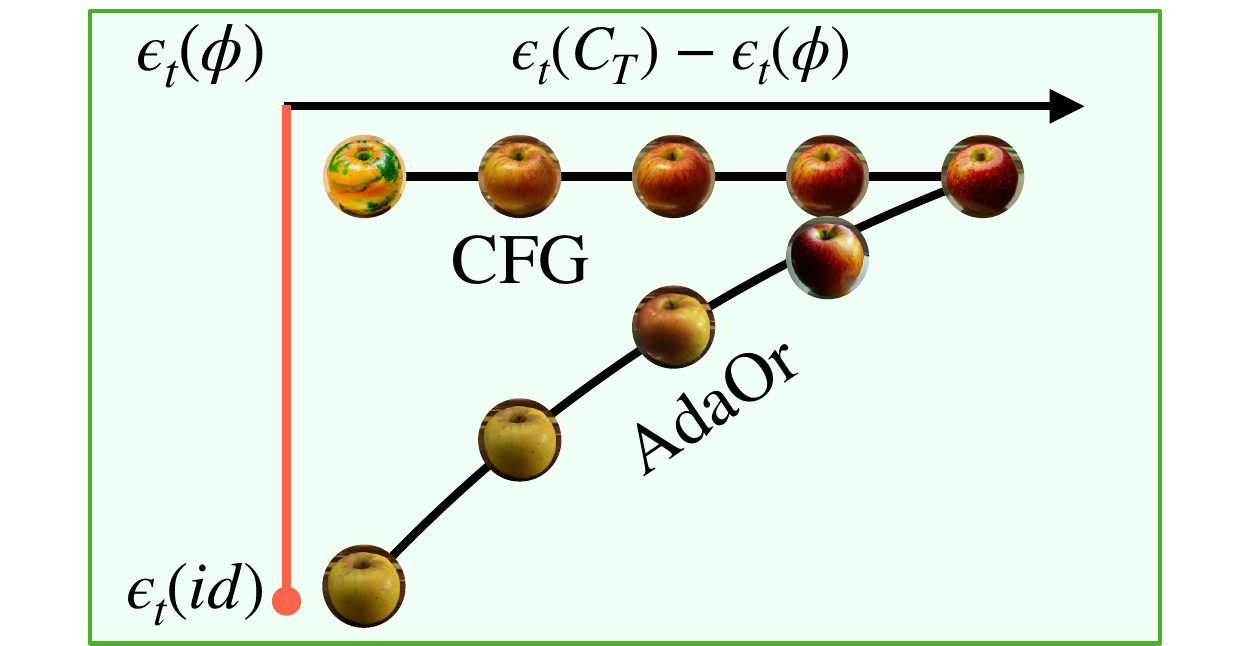}
        \caption{\centering \textbf{Edit Progression Comparison}: \\ Varying CFG scale vs AdaOr edit strength}
        \label{fig:method3}
    \end{subfigure}

    \vspace{-5pt}
    \caption{
    \textbf{Geometric interpretation of standard Classifier-Free Guidance and Adaptive Origin Guidance.} In both (a) and (b), we illustrate a single denoising step in which the latent $\mathbf{z}_t$ lies on the manifold of the marginal distribution $p_t$. The origin prediction first denoises $\mathbf{z}_t$ toward the manifold of the less noisy distribution $p_{t-1}$, after which the trajectory is steered on this manifold toward better alignment with the conditioning signal.
    In (a) \textit{standard CFG}, the origin is given by $\epsilon_t(\varnothing)$, and the guidance direction is $\epsilon_t(c_T) - \epsilon_t(\varnothing)$.
    In (b) \textit{Adaptive Origin Guidance (ours)}, the origin is interpolated between the identity prediction $\epsilon_t(\text{\REC})$ and the standard null prediction, $\epsilon_t(\varnothing)$, as a function of the edit strength. This ensures faithful reconstruction at low strengths while smoothly recovering standard CFG behavior at higher strengths.
    In (c), we show the edit progression as a function of edit strength. While standard CFG originates from the unconditional prediction (representing arbitrary edits), Adaptive Origin Guidance originates from the identity prediction, creating a trajectory that smoothly transitions from the input image to the target edit.
    }
    \vspace{-5pt}
    \label{fig:method}
\end{figure*}

%% file: figures/2_othersvsours/others_vs_ours.tex
\begin{figure*}
    \centering
    \setlength{\tabcolsep}{0pt}
    \begin{tabular}{cccc cc cccccc}
        \multicolumn{12}{c}{\textit{\input{figures/2_othersvsours/67/instruction.txt}}} \\
        \raisebox{26pt}{\rotatebox[origin=t]{90}{{Kontinuous}}} & { } &
        \raisebox{26pt}{\rotatebox[origin=t]{90}{{Kontext}}} & { } &
        \includegraphics[width=0.125\linewidth]{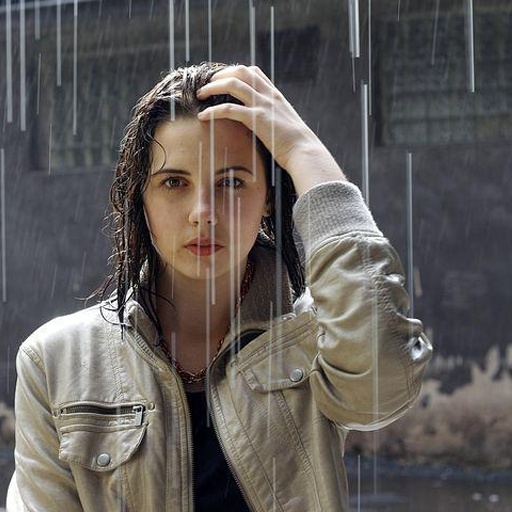} & { } &
        \includegraphics[width=0.125\linewidth]{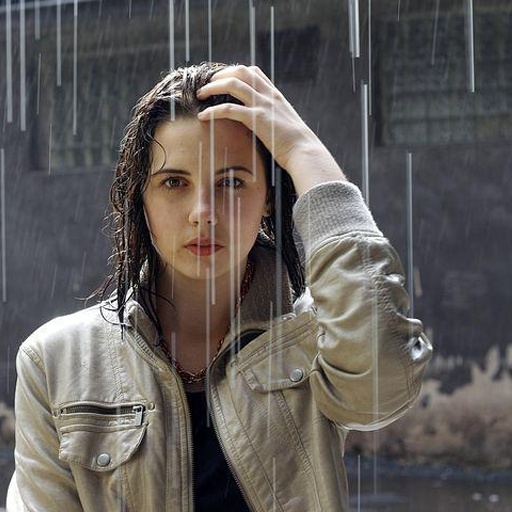} &
        \includegraphics[width=0.125\linewidth]{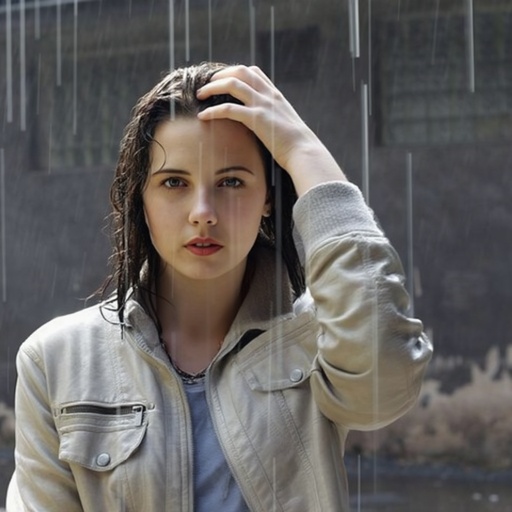} &
        \includegraphics[width=0.125\linewidth]{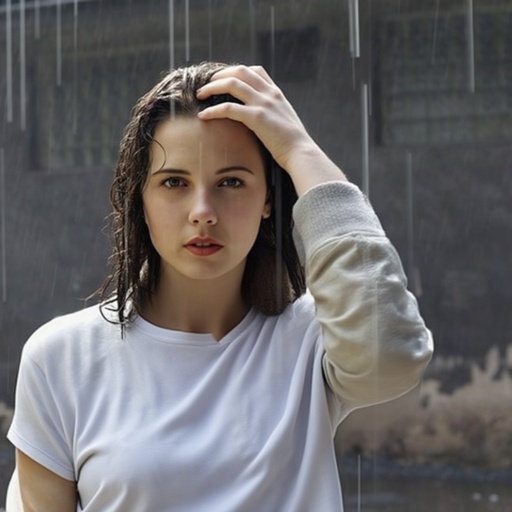} &
        \includegraphics[width=0.125\linewidth]{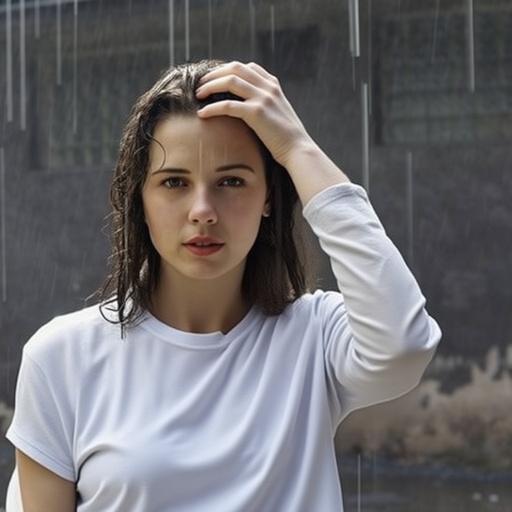} &
        \includegraphics[width=0.125\linewidth]{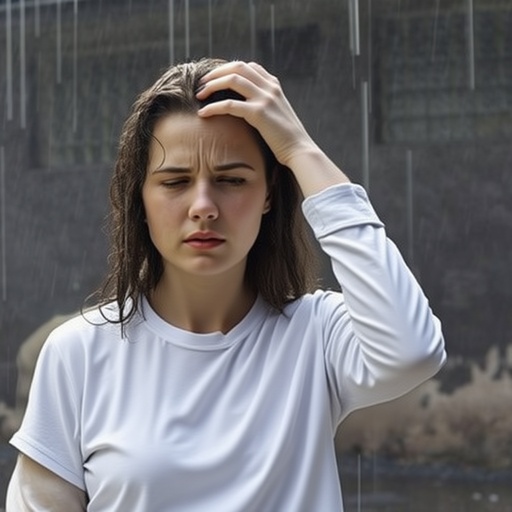} &
        \includegraphics[width=0.125\linewidth]{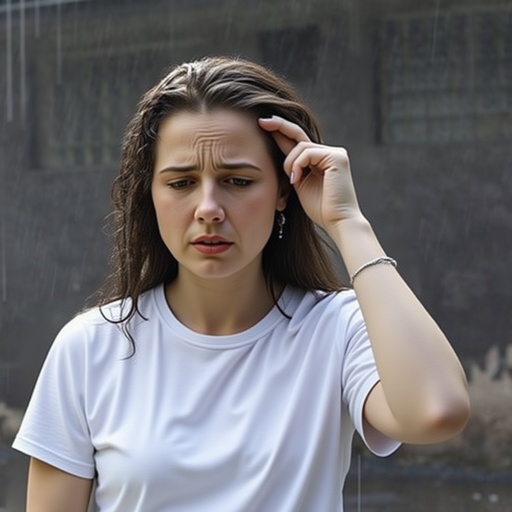} \\
        && \raisebox{26pt}{\rotatebox[origin=t]{90}{{FreeMorph}}} & { } &
        \includegraphics[width=0.125\linewidth]{figures/2_othersvsours/67/src.jpg} & { } &
        \includegraphics[width=0.125\linewidth]{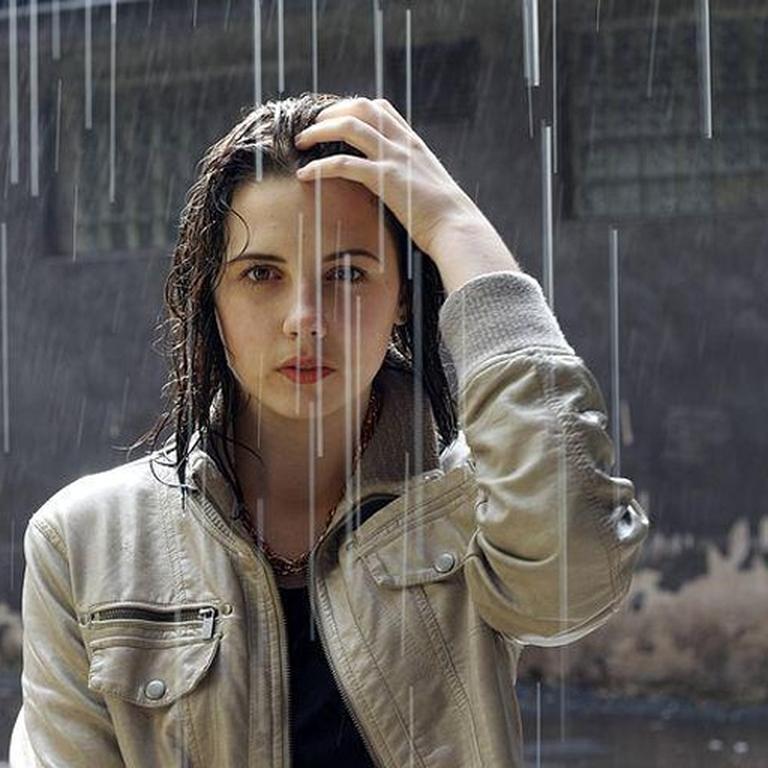} &
        \includegraphics[width=0.125\linewidth]{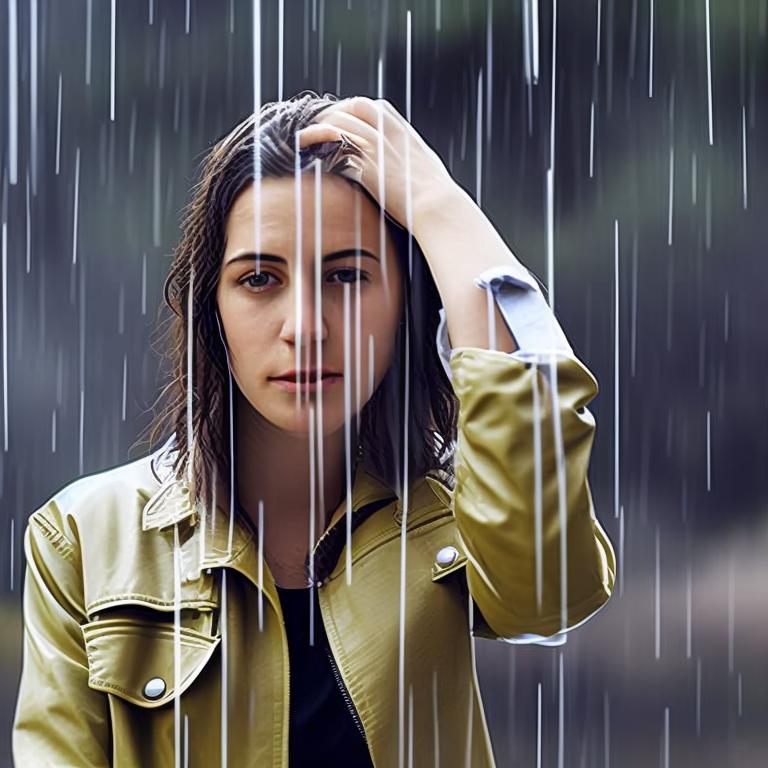} &
        \includegraphics[width=0.125\linewidth]{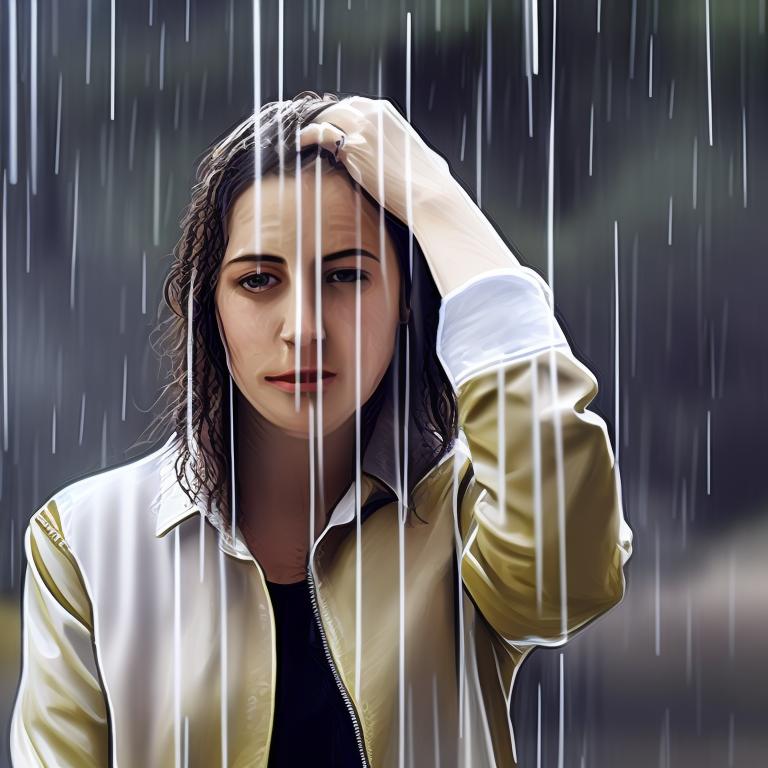} &
        \includegraphics[width=0.125\linewidth]{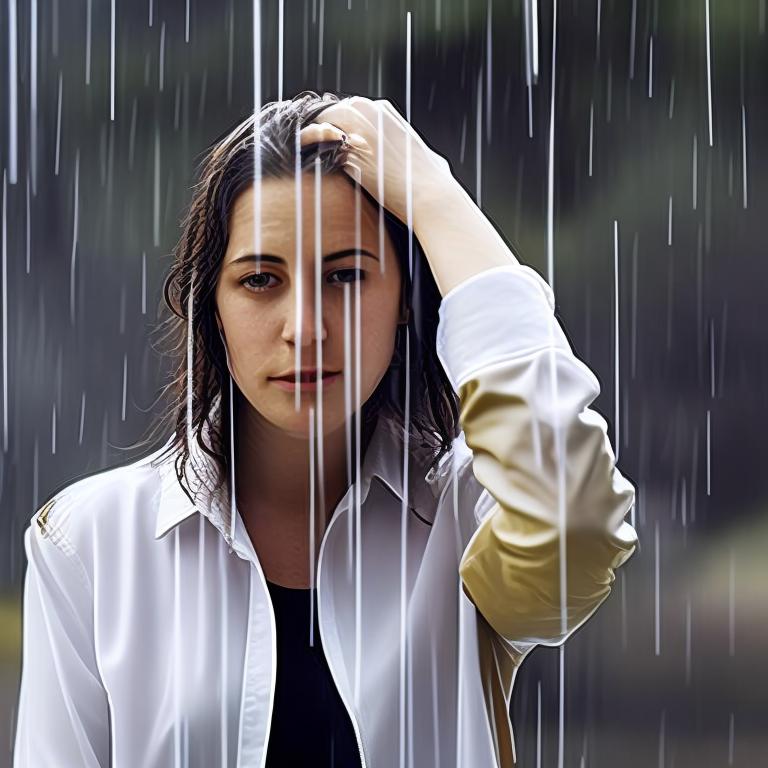} &
        \includegraphics[width=0.125\linewidth]{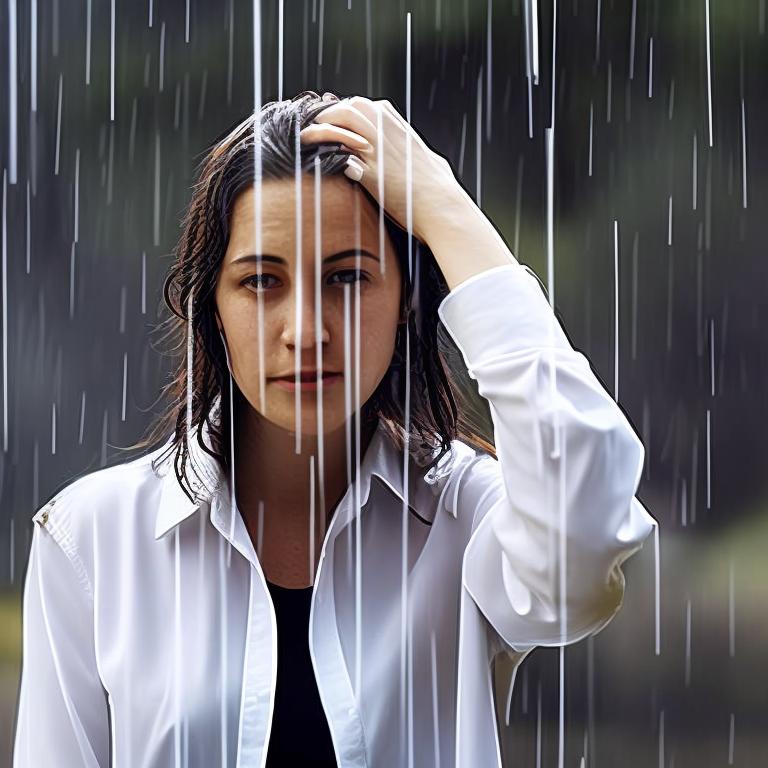} &
        \includegraphics[width=0.125\linewidth]{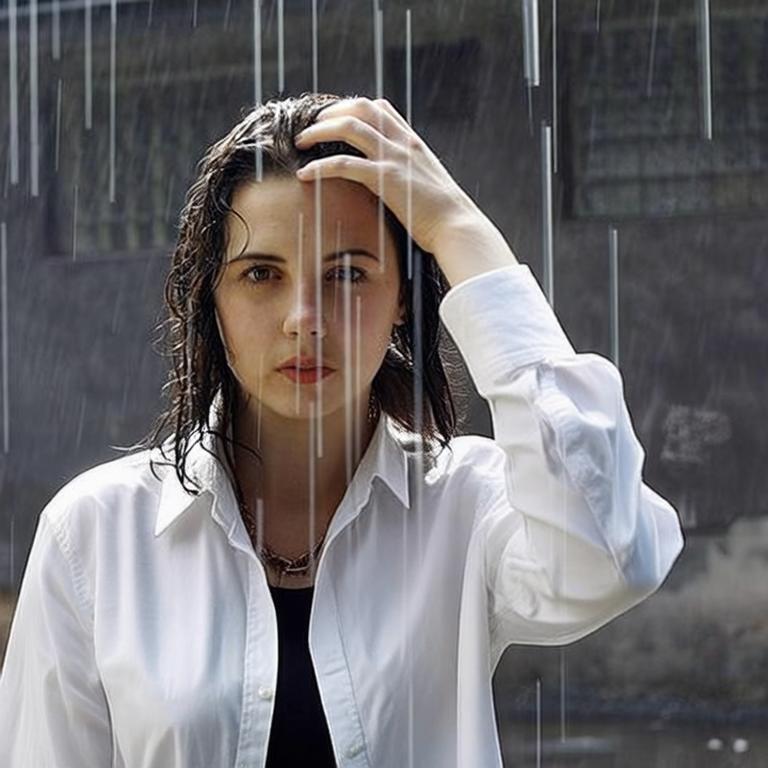} \\
        && \raisebox{26pt}{\rotatebox[origin=t]{90}{{AdaOr (Ours)}}} & { } &
        \includegraphics[width=0.125\linewidth]{figures/2_othersvsours/67/src.jpg} & { } &
        \includegraphics[width=0.125\linewidth]{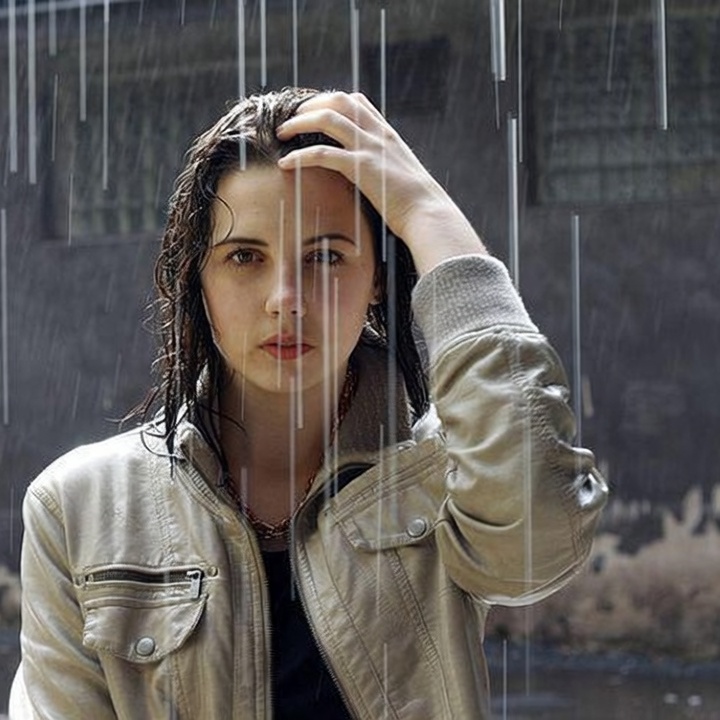} &
        \includegraphics[width=0.125\linewidth]{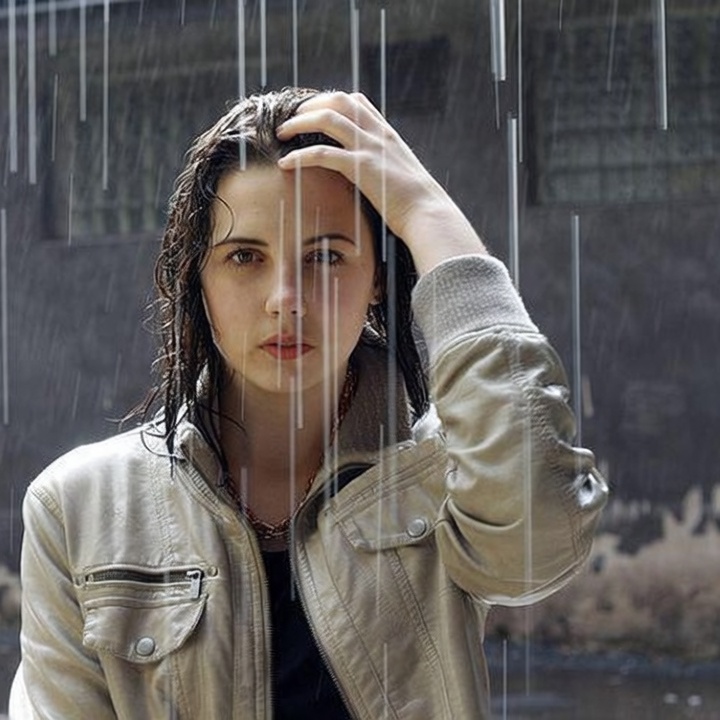} &
        \includegraphics[width=0.125\linewidth]{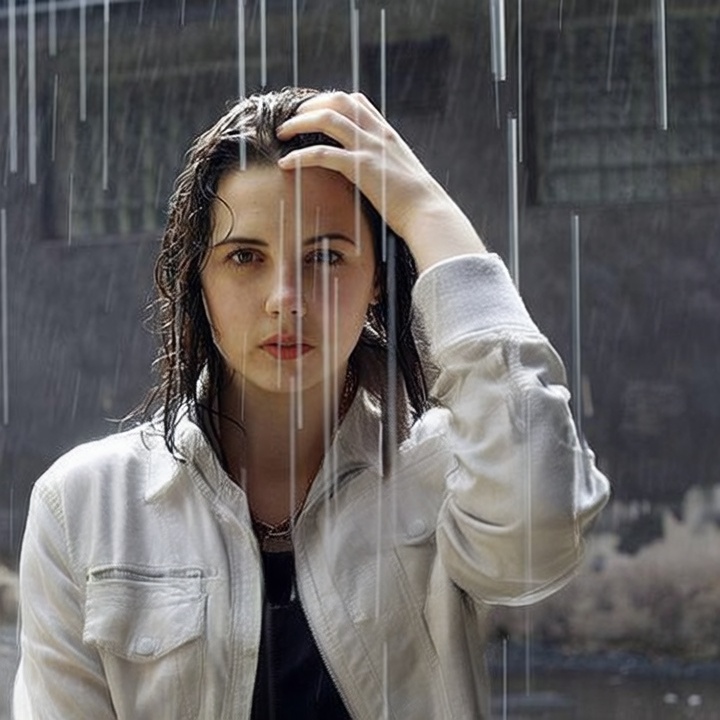} &
        \includegraphics[width=0.125\linewidth]{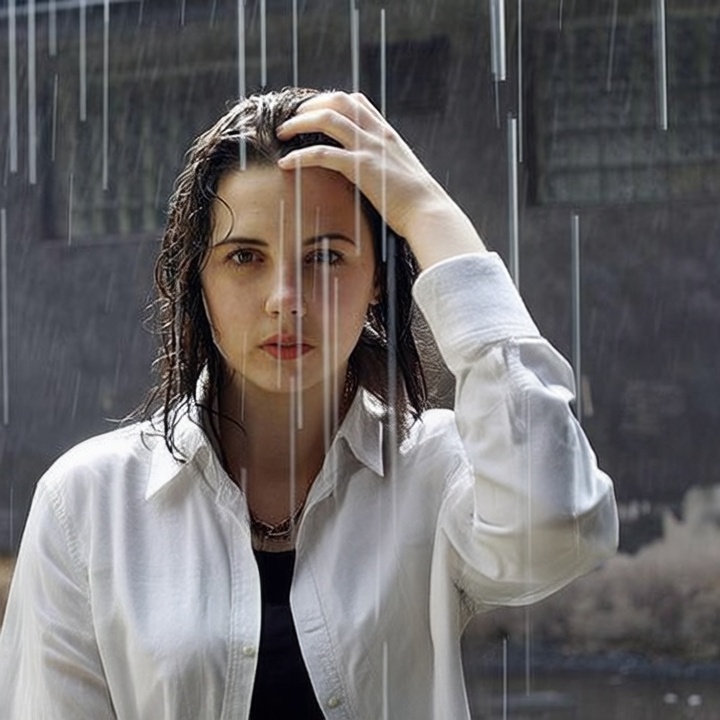} &
        \includegraphics[width=0.125\linewidth]{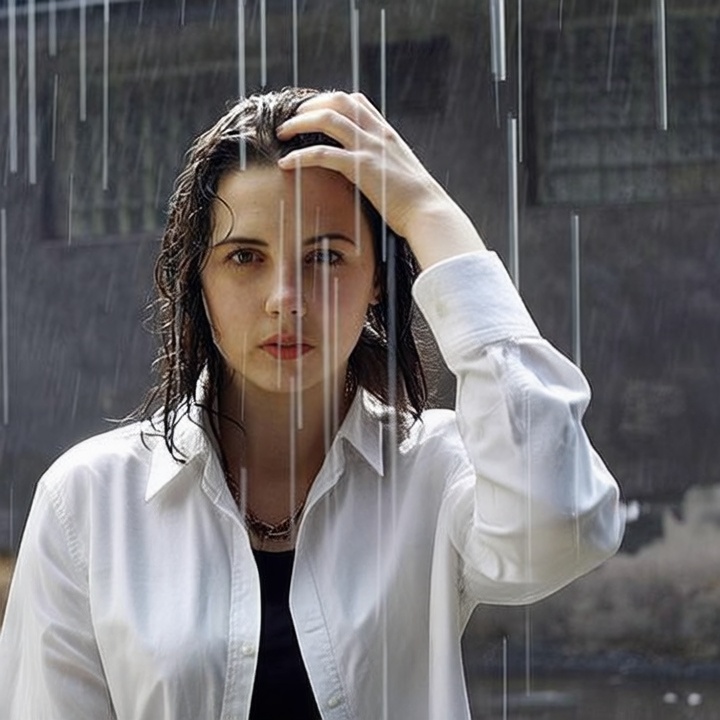} &
        \includegraphics[width=0.125\linewidth]{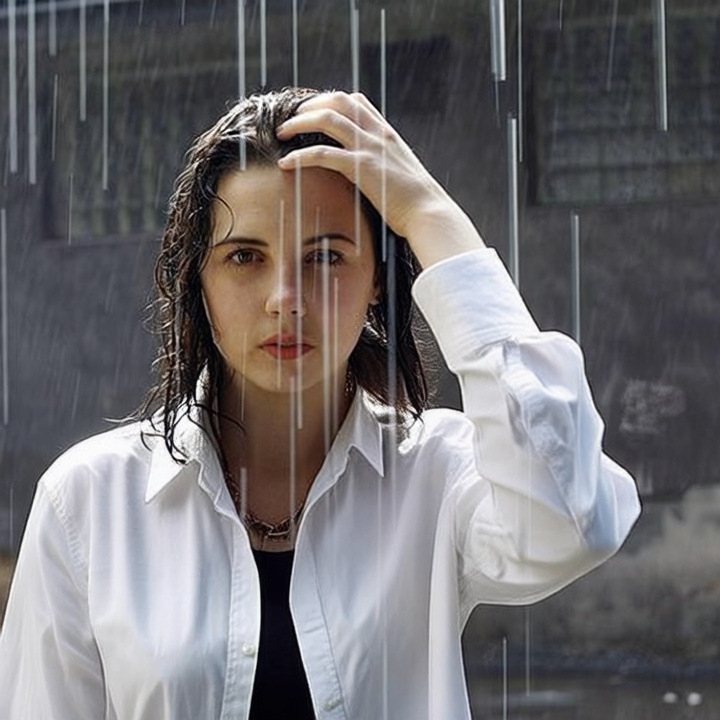} \\
        &&&& Input && 0.0 & 0.2 & 0.4 & 0.6 & 0.8 & 1.0
    \end{tabular}
    \vspace{-8pt}
    \caption{\textbf{Qualitative comparison.}
We compare our method (bottom) against Kontinuous Kontext (top) and FreeMorph \cite{cao2025freemorph} (middle).
Kontinuous Kontext \cite{parihar2025kontinuous} suffers from semantic entanglement, removing the rain and altering the woman's expression as the edit strength increases.
FreeMorph fails to generate plausible intermediate states, producing severe artifacts (e.g., the distorted sleeve and hand at strength 0.2).
In contrast, AdaOr (Ours) produces a smooth, linear transition that effectively applies the edit while strictly preserving the input content and the subject's appearance.}
\vspace{-10pt}
    \label{fig:comparison}
\end{figure*}

%% file: figures/2_othersvsours/67/instruction.txt
``Transform the woman from wearing a jacket to a white shirt.''

%% file: 4_experiments_2.tex
\section{Experiments}
\label{sec:experiments}

In this section, we present an extensive evaluation of our method for continuous editing across both video and image domains. Given the dynamic nature of the video results, we encourage readers to refer to the accompanying video and our \href{https://adaor-paper.github.io/}{\color{dustyrose}{project page}} for full visual demonstrations.
In the following, we compare our method qualitatively and quantitatively against existing continuous editing baselines and perform ablation studies to validate our design choices.

\vspace{-5pt}
\subsection{Comparisons}
Since existing continuous editing methods are primarily designed for image generation and lack validation on video models, our analysis focuses on the image domain. We evaluate AdaOr against four baselines representing different continuous editing approaches.

We first compare against FreeMorph~\cite{cao2025freemorph}, which morphs between two input images. Since it requires defined endpoints rather than a text instruction, we adapt it for our task by providing the original input image and our edited result at maximum strength ($1.0$) as the source and target, respectively.
Second, we evaluate Kontinuous Kontext~\cite{parihar2025kontinuous}, which fine-tunes an image editing model on a synthetic dataset to accept a continuous scalar input for strength control.
Next, we compare with Concept Sliders~\cite{gandikota2023conceptslidersloraadaptors}, a method that learns a LoRA-based slider for each edit type.
Finally, we compare with SAEdit~\cite{kamenetsky2025saedit}, which leverages Sparse Autoencoders (SAE) trained on human-centered prompts to identify editing directions within the text encoder's latent space.
For all baselines, we utilize the official implementations and recommended settings.

\vspace{-4pt}
\paragraph{\textbf{Qualitative Results}}
We present a qualitative comparison with FreeMorph and Kontinuous Kontext in \Cref{fig:comparison}, and with Concept Sliders and SAEdit in \Cref{fig:comparison_saedit}. As shown in \Cref{fig:comparison}, FreeMorph fails to generate plausible intermediate images, exhibiting severe structural artifacts such as the distortions in the woman's fingers. Furthermore, the required inversion process degrades the fidelity of the intermediate images. While Kontinuous Kontext yields smoother transitions of higher visual quality, it suffers from semantic entanglement, altering unrelated attributes such as the rain and the subject's expression. In contrast, our method produces smooth, high-fidelity transitions that strictly adhere to the input image, preserving the original context while effectively applying the requested edit.

As shown in \Cref{fig:comparison_saedit}, Concept Sliders alters the person’s identity, while SAEdit changes the man’s hair color but introduces only weak curl patterns. In contrast, AdaOr achieves the desired edit while maintaining a continuous sequence.
Additional qualitative results are provided on pages 10–11.

\input{figures/5_compare-sliders/comparison_sliders}

\vspace{-4pt}
\paragraph{\textbf{Quantitative Results}}
Following the evaluation protocol of Kontinuous Kontext, we utilize a subset of PIE-Bench~\cite{ju2023direct} to quantitatively evaluate FreeMorph and Kontinuous Kontext. More details in Appendix \ref{app:benchmarks}. Since Concept Sliders and SAEdit are not suitable for the broad scope of PIE-Bench (Concept Sliders requires training a specific LoRA per edit type, and SAEdit was trained on human-centered prompts), we evaluate these methods on the dedicated benchmark provided by SAEdit.

We evaluate our method using $N=6$ edits with uniform strengths, assessing performance across three dimensions: (i) smoothness: To ensure the editing process is not ``jumpy'', we measure second-order smoothness using $\delta_\text{smooth}$~\cite{parihar2025kontinuous} and introduce a \textit{Linearity} metric that assesses the uniformity of the editing pace by measuring the variance of perceptual changes between consecutive edit strengths; (ii) text alignment consistency: We introduce \textit{Normalized CLIP-Dir} to verify that every intermediate strength moves semantically towards the target prompt, measuring the average alignment between the local direction at each strength interval and the global text direction; and (iii) perceptual trajectory consistency: To ensure the edit follows a direct path in perceptual space, we measure the cosine similarity between the update at each strength interval and the total edit vector using the DreamSim~\cite{fu2023dreamsim} metric. Full mathematical definitions and implementation details are provided in Appendix \ref{app:metrics}.

The quantitative results are presented in \Cref{tab:quatn_results,tab:quatn_results_saedit}. As shown in the tables, our method consistently outperforms all baselines across all evaluated metrics. While Kontinuous Kontext achieves comparable smoothness, our method demonstrates superior text alignment consistency and perceptual trajectory consistency. Furthermore, unlike the competing approaches which are restricted to the image domain, we explicitly demonstrate the extensibility of our method to video editing.

\input{tables/baseline-comparison}

\input{tables/concept-comparisons}

\vspace{-2pt}
\paragraph{\textbf{User Study}}
We further evaluate our method through a user study comparing AdaOr against two strong baselines, Kontinuous Kontext and FreeMorph. The evaluation was conducted on random samples from the PIE-Bench dataset~\cite{ju2023direct}, using source images and corresponding editing instructions. For AdaOr and Kontinuous Kontext, we generated intermediate images at six strength values ranging from $0$ to $1$. FreeMorph was evaluated under two settings: one using a Lucy-generated edit at maximum strength to enable a direct comparison of interpolation behavior with AdaOr, and another using a third-party model~\cite{wu2025qwenimagetechnicalreport} to assess performance with an alternative endpoint.
A total of 36 participants each evaluated 10 randomly assigned tuples, comparing AdaOr’s transition sequences against those of a baseline. For each tuple, participants answered three questions: (1) which transition is more linear and smooth, (2) which has more natural-looking intermediate frames, and (3) which result is preferred overall. As shown in \Cref{fig:userstudy}, AdaOr outperforms both FreeMorph variants across all metrics. Finally, AdaOr achieves comparable intermediate quality and overall preference to Kontinuous Kontext, while outperforming it in transition smoothness.

\subsection{Ablation Studies}

Next, we perform ablation studies. First, we compare against standard CFG by removing the adaptive origin mechanism entirely. Second, we evaluate CFG-\REC, where we replace the unconditional prediction in CFG with the identity prediction. This is formulated as: $\epsilon(\mathbf{z_t}; c_I, \text{\REC}, t) + w (\epsilon(\mathbf{z_t}; c_I, c_T, t) - \epsilon(\mathbf{z_t}; c_I, \text{\REC}, t))$. 
Finally, we examine the role of our scheduling strategy by replacing the square root scheduler with a linear scheduler, $s(\alpha) = \alpha$.

\input{figures/user_study/user_study}

\input{figures/10_ablation_full/ablation_full}

We present the qualitative results of the ablation studies in \Cref{fig:ablation_full}. Consistent with previous observations, standard CFG generates arbitrary content at low scales, failing to produce meaningful small-strength edits. As predicted by the analysis in \Cref{sec:rec-analysis}, CFG-\REC{} fails to produce a smooth sequence. As the scale increases, the generation becomes visually exaggerated and distorted, diverging significantly from the target semantics rather than converging to the correct edit. 
While the linear scheduler produces valid edits, the magnitude of change is inconsistent across the editing strengths. In contrast, our full method successfully balances input preservation with editing fidelity, ensuring a smooth and consistent transition.

The quantitative ablation results are reported in the lower section of \Cref{tab:quatn_results}. As shown, our full method significantly outperforms standard CFG across all metrics. Notably, while CFG-\REC{} achieves a high linearity score, indicating uniform perceptual step sizes, its poor $\delta_\text{smooth}$ score reflects a lack of second-order smoothness, consistent with the jagged transitions observed qualitatively. Finally, although the linear scheduler attains performance comparable to our approach, our method exhibits superior smoothness.

\subsection{Limitations}
Our method adopts the training framework of the backbone editing model (e.g., Lucy-Edit). However, as we utilize only a subset of the original training data and a shorter training schedule, our model's representational scope is naturally more constrained. Using our model, we cannot perform continuous edits beyond the backbone's editing capabilities, and we may inherit tendencies toward unnecessary changes or failures in specific edit types. We illustrate two such failure cases in \Cref{fig:limitations}. Additionally, our adaptive origin mechanism introduces a slight computational overhead at inference time. While standard CFG requires two noise predictions per step, our method employs three predictions (unconditional, conditional, and \REC{}), resulting in a modest increase in computational cost.

\input{figures/11_limitations/limitations}

%% file: figures/5_compare-sliders/comparison_sliders.tex
\begin{figure}
    \centering
    \setlength{\tabcolsep}{0pt}
    \begin{tabular}{cc c cccc}
    \multicolumn{7}{c}{\textit{``Change to curly hair.''}} \\

    \raisebox{18pt}{\rotatebox[origin=t]{90}{\small{Concept Sliders}}} & { } &
    \includegraphics[width=0.195\linewidth]{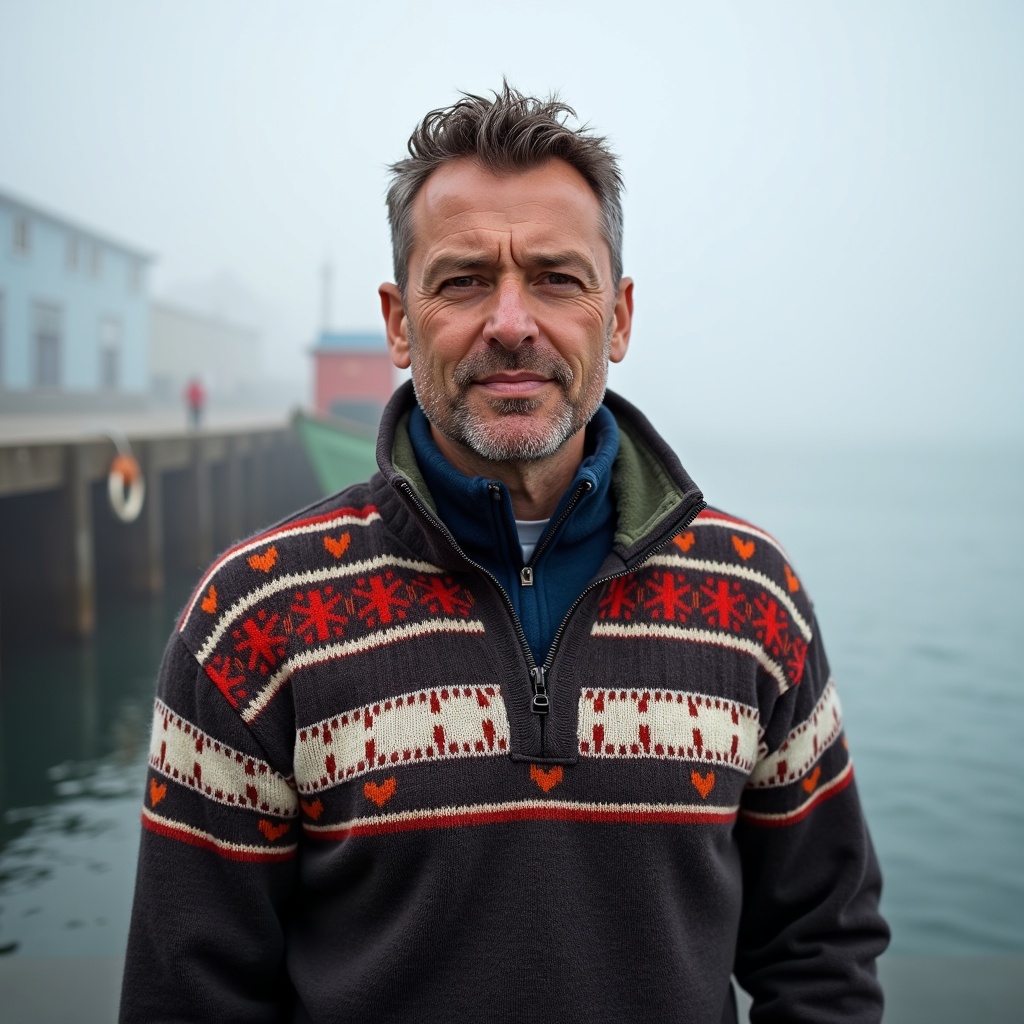} & 
    \includegraphics[width=0.195\linewidth]{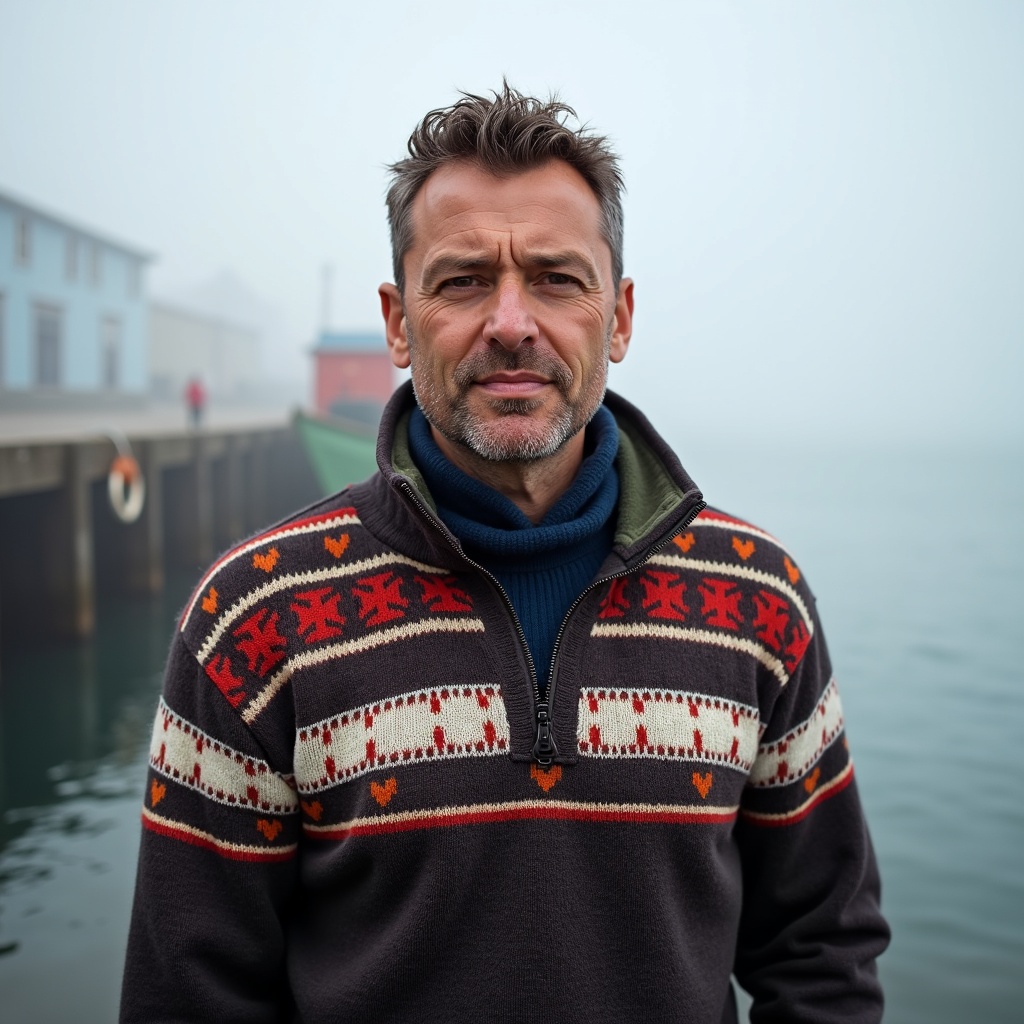} & 
    \includegraphics[width=0.195\linewidth]{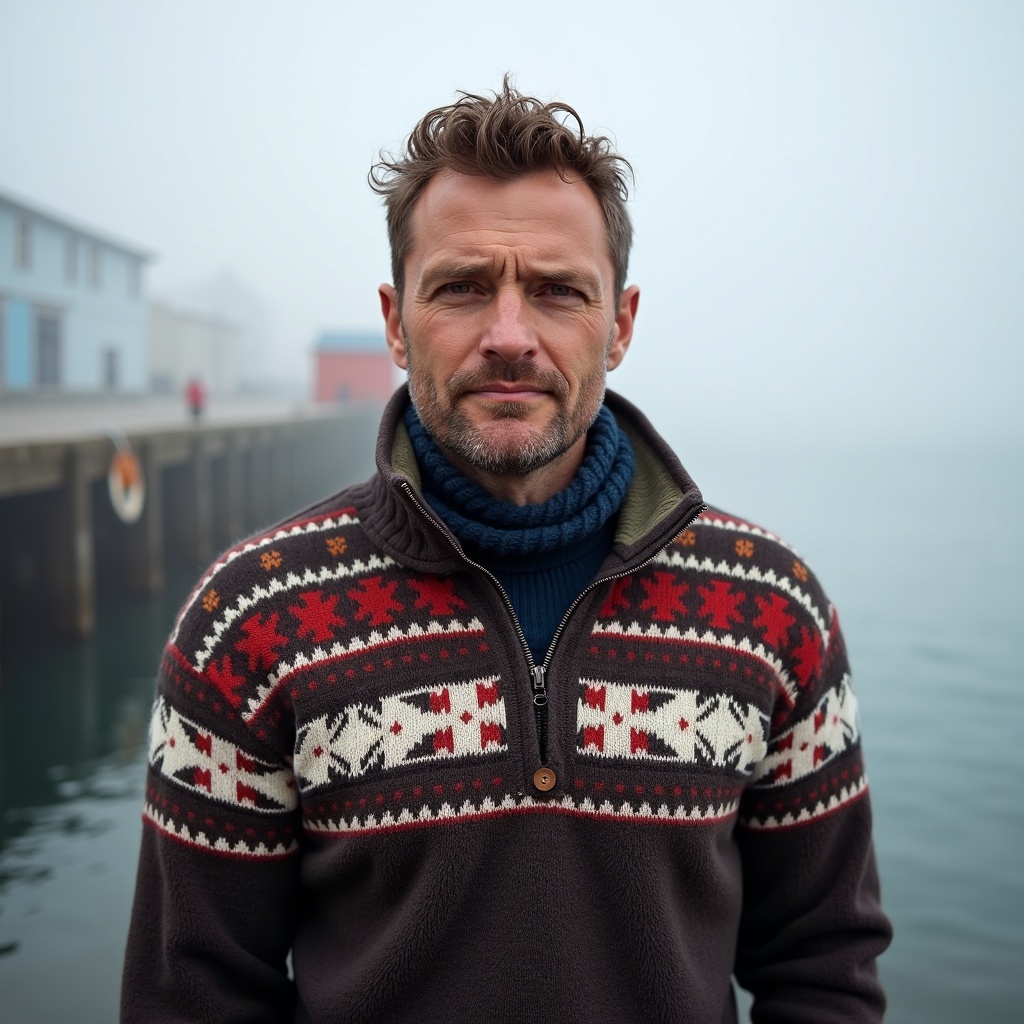} & 
    \includegraphics[width=0.195\linewidth]{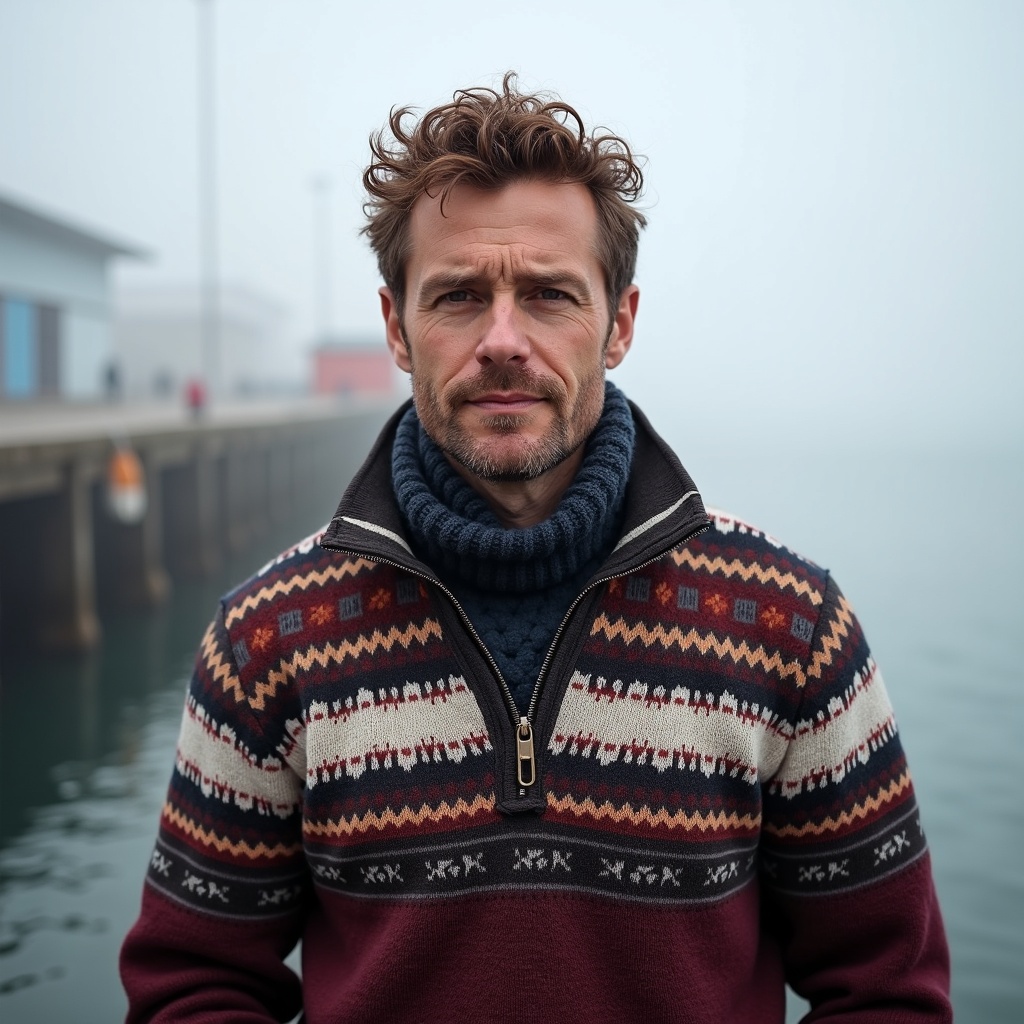} & 
    \includegraphics[width=0.195\linewidth]{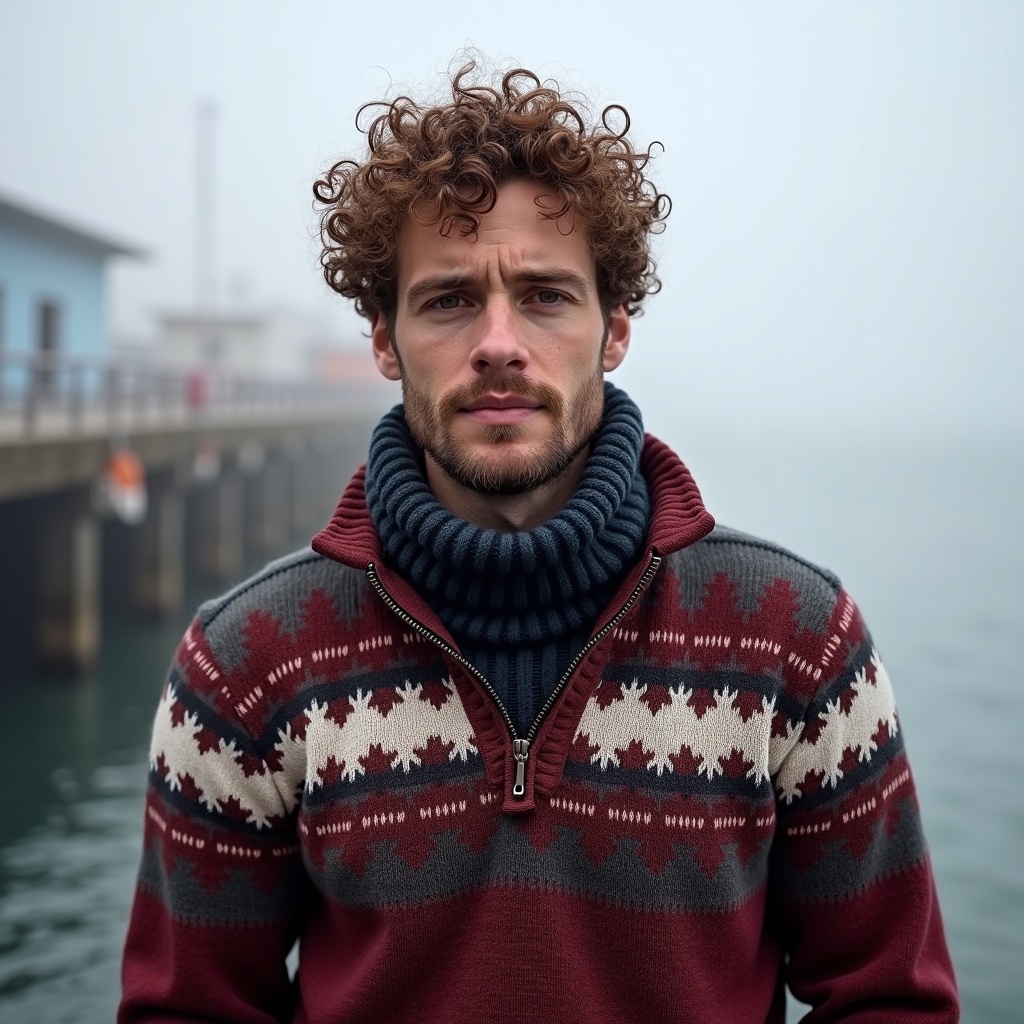} \\
    \raisebox{18pt}{\rotatebox[origin=t]{90}{\small{SAEdit}}} & { } &
    \includegraphics[width=0.195\linewidth]{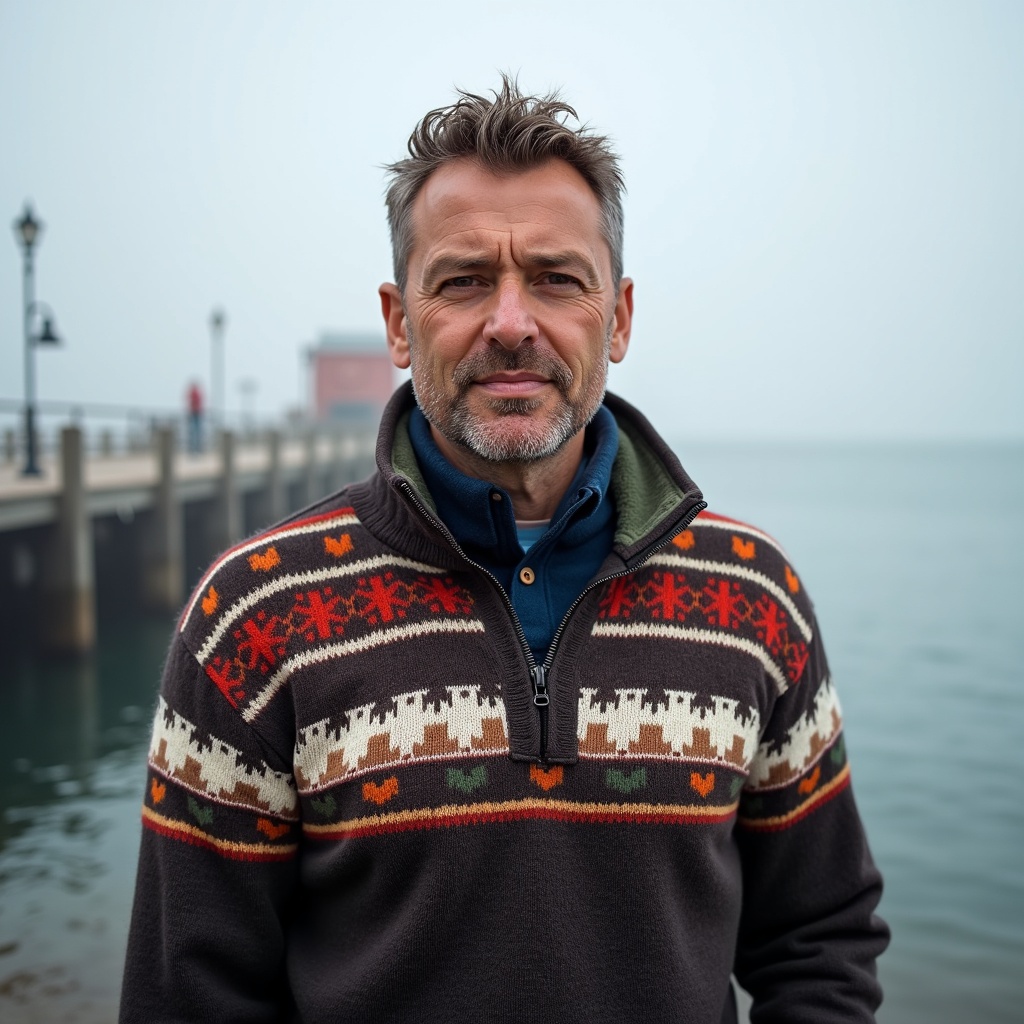} & 
    \includegraphics[width=0.195\linewidth]{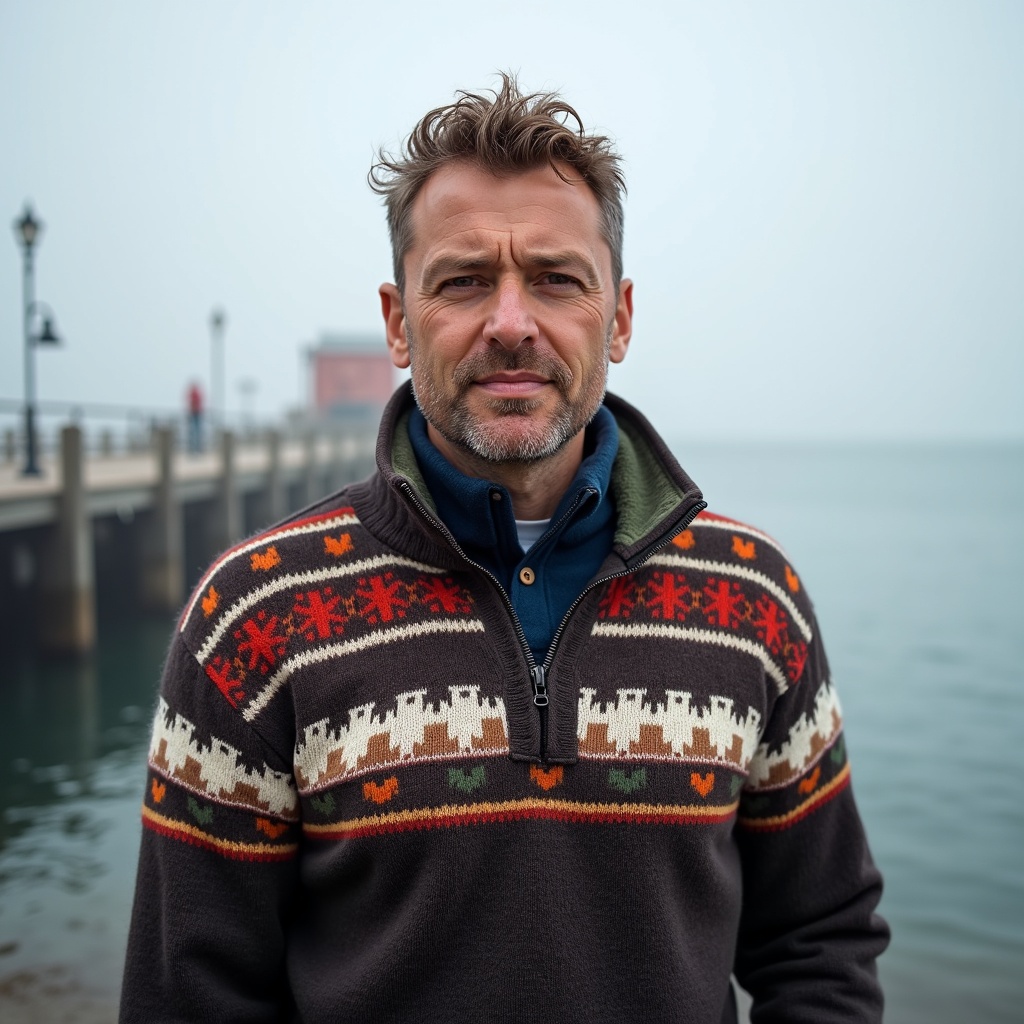} & 
    \includegraphics[width=0.195\linewidth]{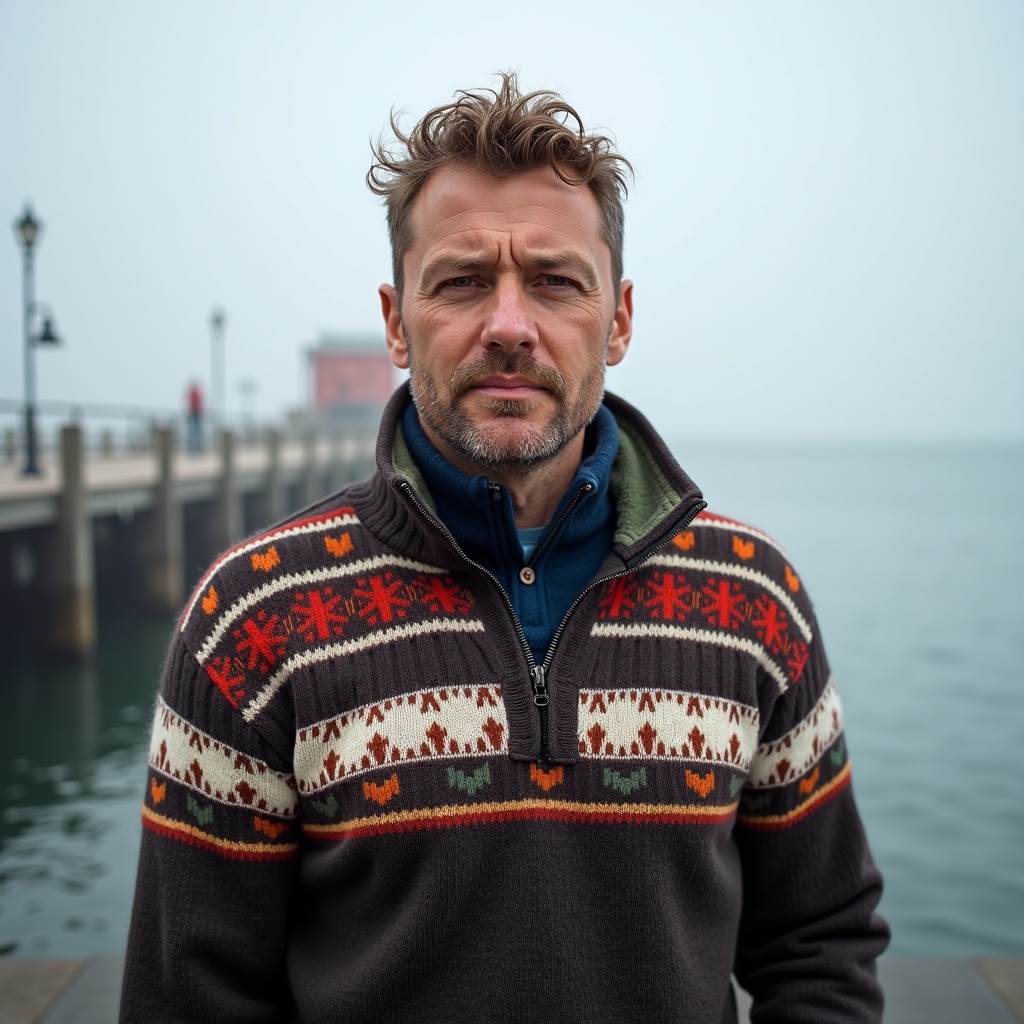} & 
    \includegraphics[width=0.195\linewidth]{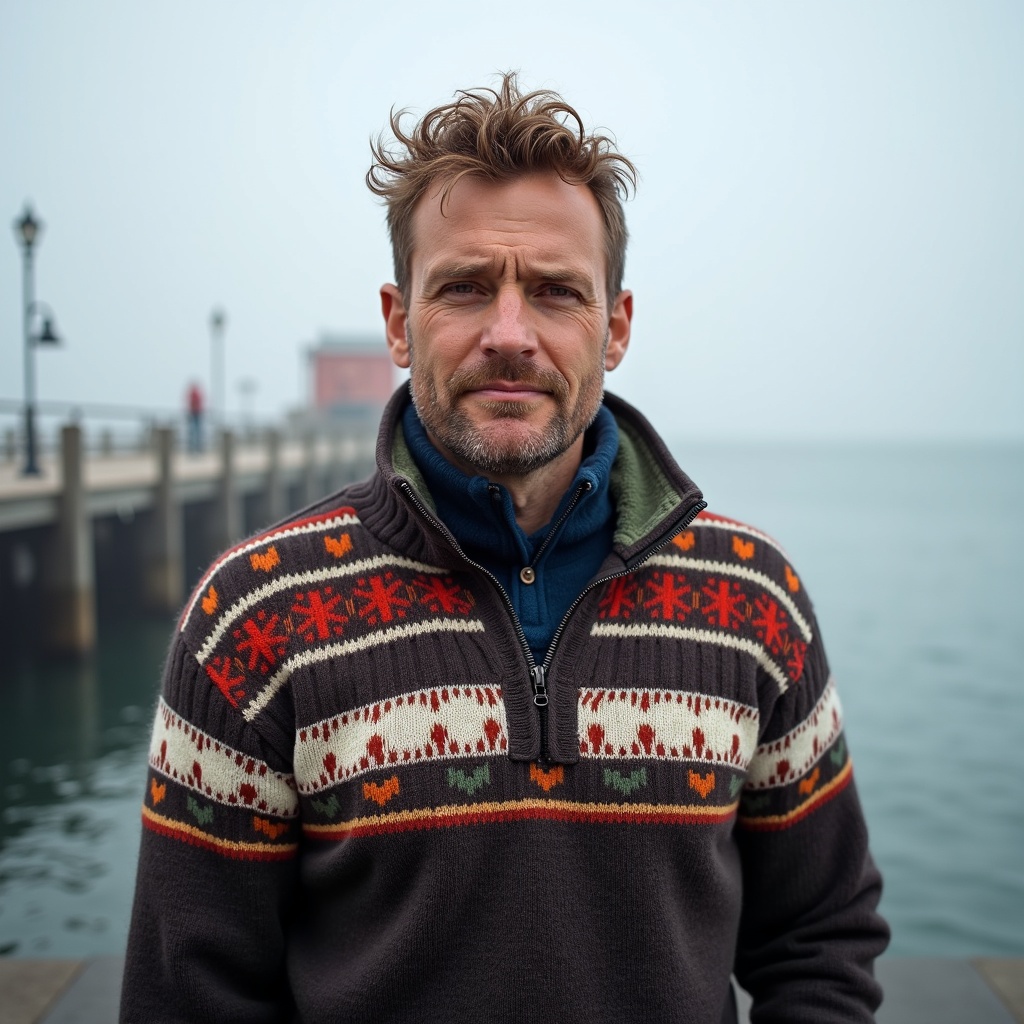} & 
    \includegraphics[width=0.195\linewidth]{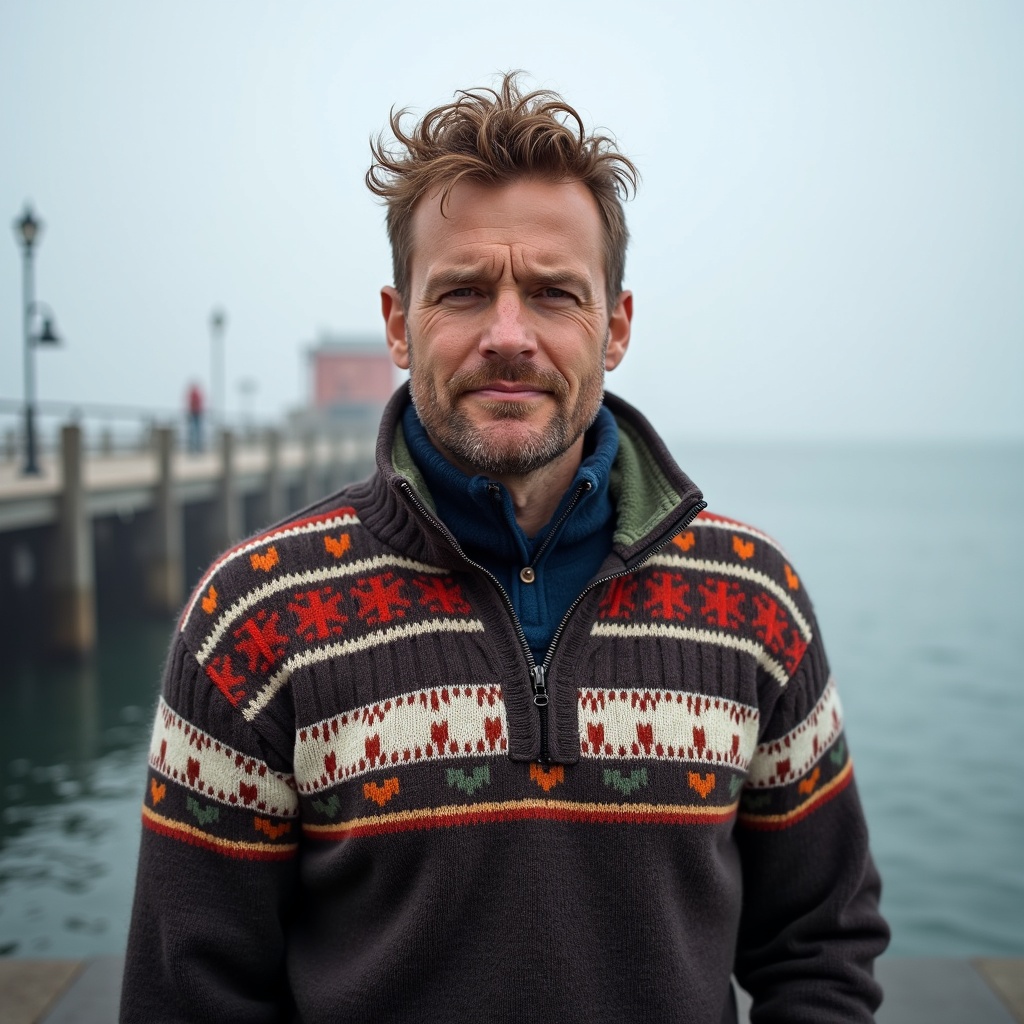} \\
    \raisebox{18pt}{\rotatebox[origin=t]{90}{\small{AdaOr (Ours)}}} & { } &
    \includegraphics[width=0.195\linewidth]{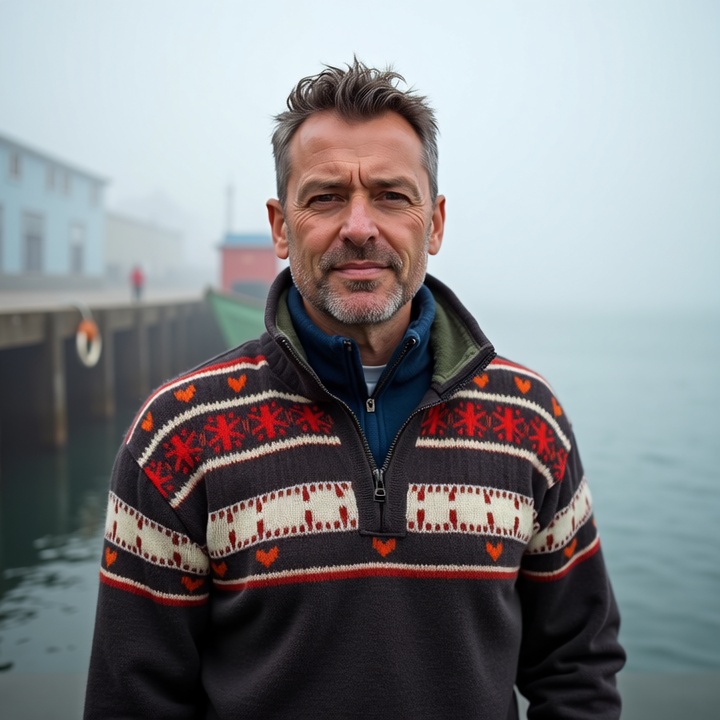} & 
    \includegraphics[width=0.195\linewidth]{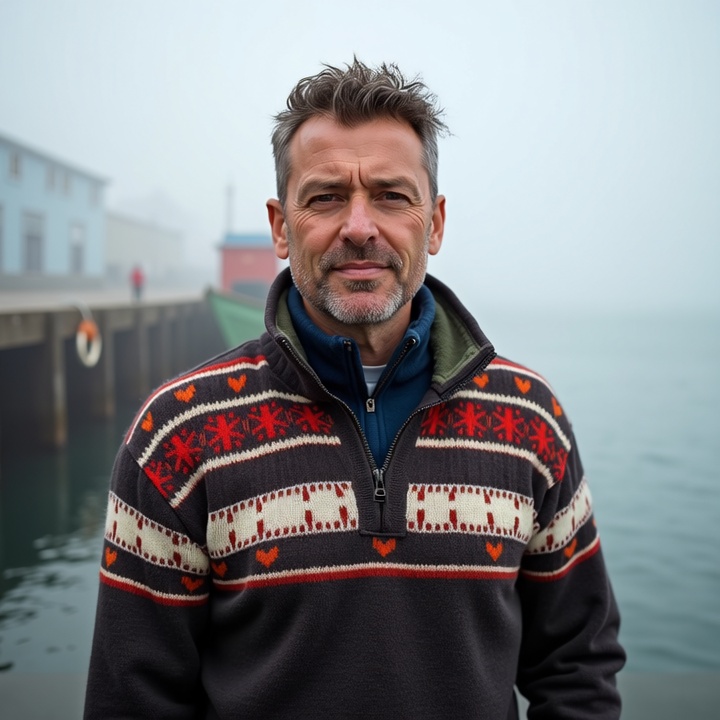} &
    \includegraphics[width=0.195\linewidth]{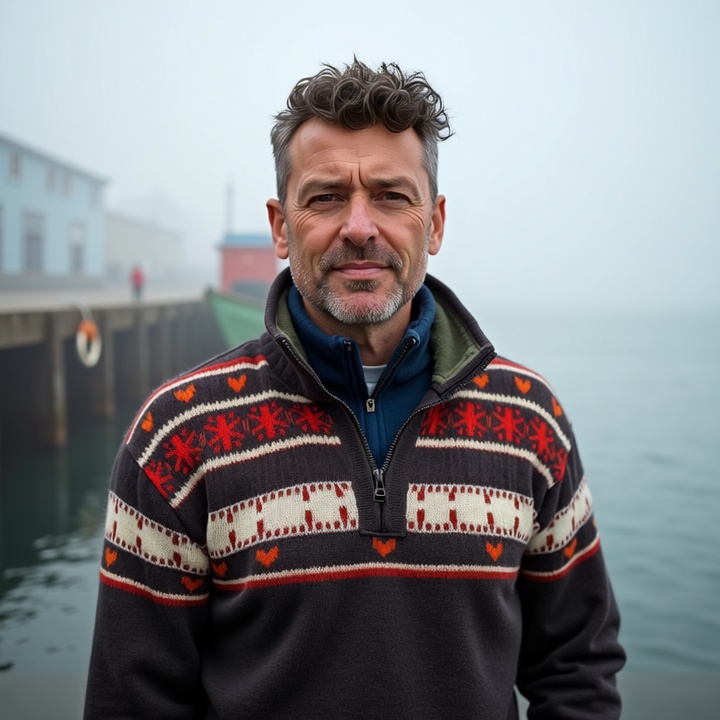} &
    \includegraphics[width=0.195\linewidth]{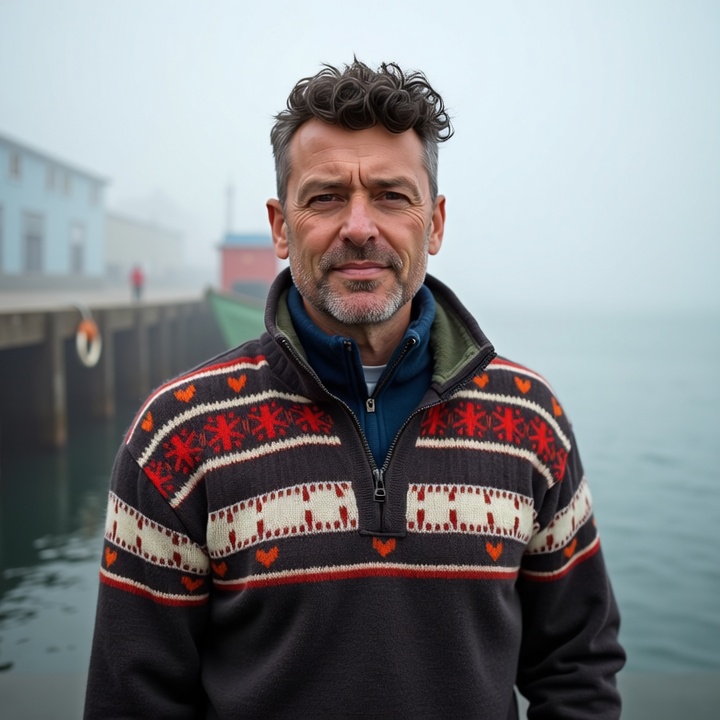} &
    \includegraphics[width=0.195\linewidth]{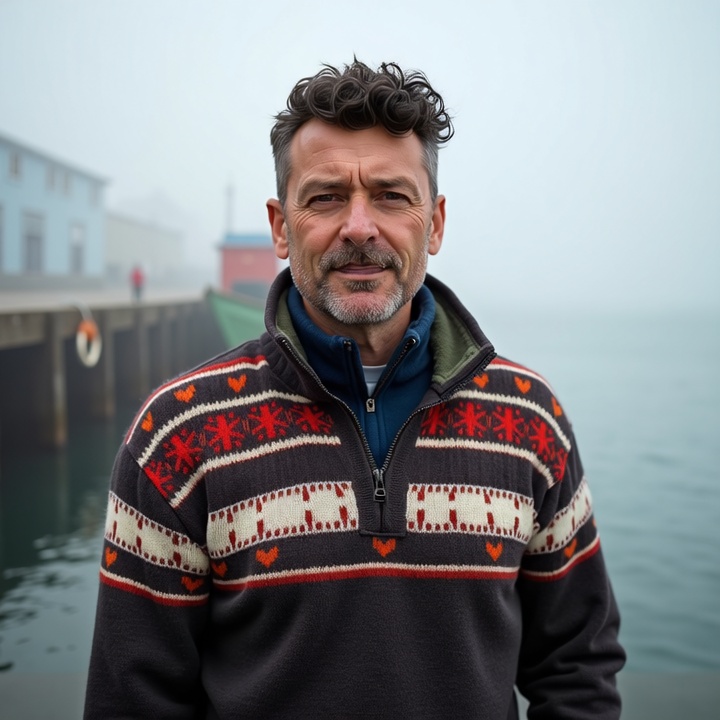} \\
    && \multicolumn{5}{l}{\small{Edit Intensity} $\xrightarrow{\hspace{175pt}}$} \\

    \end{tabular}
    \vspace{-10pt}
    \caption{\textbf{Qualitative comparison.}
We compare our method (bottom) against Concept Sliders (top) and SAEdit (middle).
Concept Sliders modifies the man's identity, while SAEdit introduces only weak curl patterns.
AdaOr (Ours) produces a smooth transition that effectively applies the edit while preserving the input content and the subject's identity.}
\vspace{-8pt}
    \label{fig:comparison_saedit}
\end{figure}

%% file: tables/baseline-comparison.tex
\begin{table}
\small
\setlength{\tabcolsep}{2pt}
\caption{\textbf{Quantitative evaluation (PIE-Bench).}
The top block compares our method to prior work, while the bottom block reports ablations of guidance formulations and scheduling strategies. We evaluate smoothness, text alignment consistency, perceptual trajectory consistency, and linearity.}
\vspace{-8pt}
    \centering
    \begin{tabular}{l  cc   cc }
    \toprule 
        \multirow{2}{*}{\textbf{Method}}
        & \multirow{2}{*}{$\boldsymbol{\delta_\textbf{smooth}}(\downarrow)$}
        & \textbf{Norm}. 
        & \textbf{DreamSim}
        & \multirow{2}{*}{\textbf{Linearity} $(\downarrow)$}
    \\ 
          
        & 
        & \textbf{CLIP-Dir} $(\uparrow) $
        & \textbf{Align} $(\uparrow)$
        & 
    \\ 
    \midrule 
     
        FreeMorph            
        & 0.26              
        & 1.71  
        & 0.23
        & 0.10
    \\
    
        Kontinuous K.    
        & \textbf{0.12}    
        & 1.75 
        & \underline{0.32}
        & 0.08
    \\
    \midrule
    
    CFG             
        & 0.61        
        & 1.48
        & 0.27
        & 0.12
    \\
     CFG-\REC             
        & 0.27              
        & 1.65
        & 0.30
        & \textbf{0.05}
    \\
     Linear Scheduler            
        & \underline{0.14}
        & \textbf{1.99}
        & \textbf{0.36}
        & \underline{0.07}
    \\
    \midrule
    
        AdaOr (Ours)           
        & \textbf{0.12}
        & \underline{1.89}
        & \textbf{0.36}
        & \underline{0.07}
    \\
    \bottomrule
    \end{tabular}
    \vspace{-10pt}
    \label{tab:quatn_results}
\end{table}

%% file: tables/concept-comparisons.tex
\begin{table}
\small
\setlength{\tabcolsep}{2pt}
\caption{\textbf{Quantitative evaluation (human-focused).}
We compare our method to two other continuous editing methods, evaluating smoothness, text alignment consistency, perceptual trajectory consistency, and linearity.}
\vspace{-6pt}

    \centering
    \begin{tabular}{l  cc   cc }
    \toprule 
        \multirow{2}{*}{\textbf{Method}}
        & \multirow{2}{*}{$\boldsymbol{\delta_\textbf{smooth}}(\downarrow)$}
        & \textbf{Norm.} 
        & \textbf{DreamSim}
        & \multirow{2}{*}{\textbf{Linearity} $(\downarrow)$}
    \\ 
          
        & 
        & \textbf{CLIP-Dir} $(\uparrow) $
        & \textbf{Align} $(\uparrow)$
        & 
    \\ 
    \midrule 
    
        ConceptSliders
        & 0.26    
        & 1.82 
        & 0.35
        & 0.15
    \\
    
    SAEdit
        & 0.28    
        & 2.07 
        & 0.36
        & 0.11
    \\
    \midrule
    
        AdaOr (Ours)         
        & \textbf{0.24}
        & \textbf{2.21}  
        & \textbf{0.37}
        & \textbf{0.10}
    \\
    \bottomrule
    \end{tabular}
    \vspace{-10pt}
    \label{tab:quatn_results_saedit}
\end{table}

%% file: figures/user_study/user_study.tex
\begin{figure}
  \includegraphics[width=\linewidth]{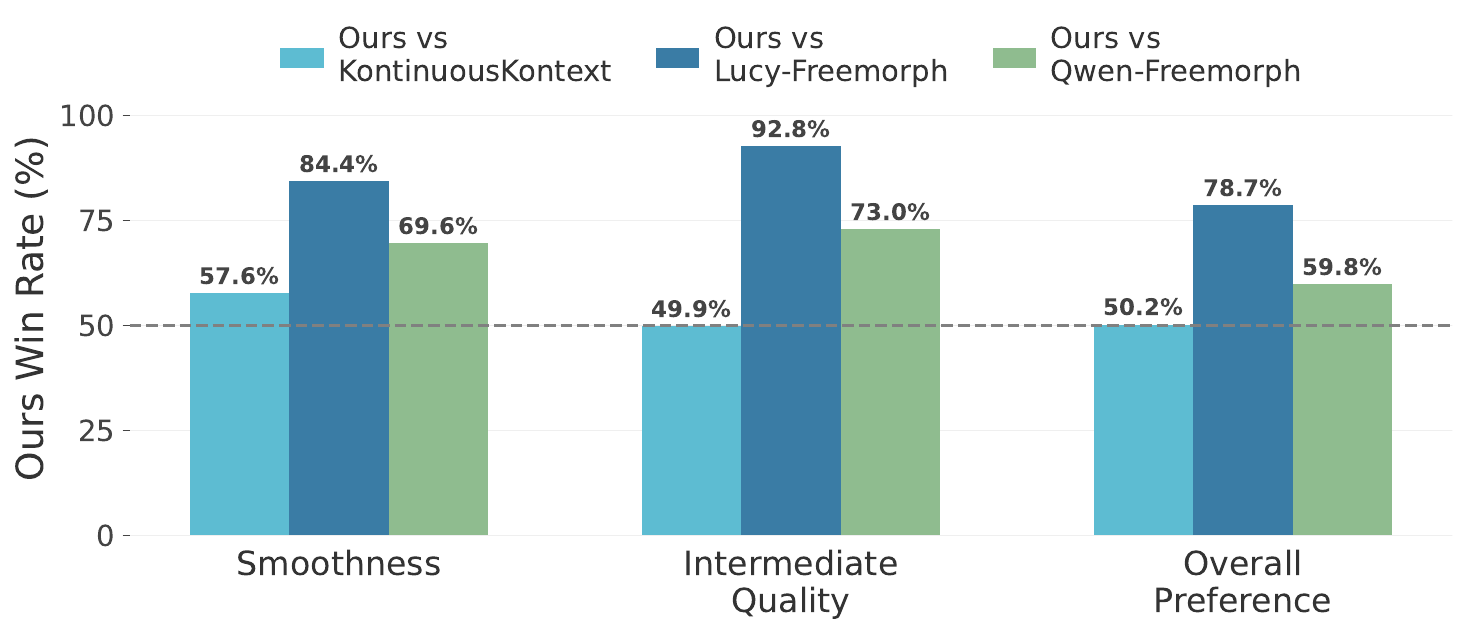}
  \vspace{-18pt}
  \caption{\textbf{User study results.}
We report the win rate of our method compared to three baselines: Kontinuous Kontext, Lucy-Freemorph, and Qwen-Freemorph.
Human evaluators assessed three aspects: smoothness of the transition, intermediate quality, and overall preference.}
\vspace{-6pt}
  \label{fig:userstudy}
\end{figure}

%% file: figures/10_ablation_full/ablation_full.tex
\begin{figure}
    \centering
    \setlength{\tabcolsep}{0pt}
    \begin{tabular}{cc c ccccc}
    \multicolumn{8}{c}{\textit{``\input{figures/10_ablation_full/instruction.txt}''}} \\

    \raisebox{21pt}{\rotatebox[origin=t]{90}{\small{CFG}}} & { } &
    \includegraphics[width=0.15\linewidth, trim=120 10 80 40, clip]{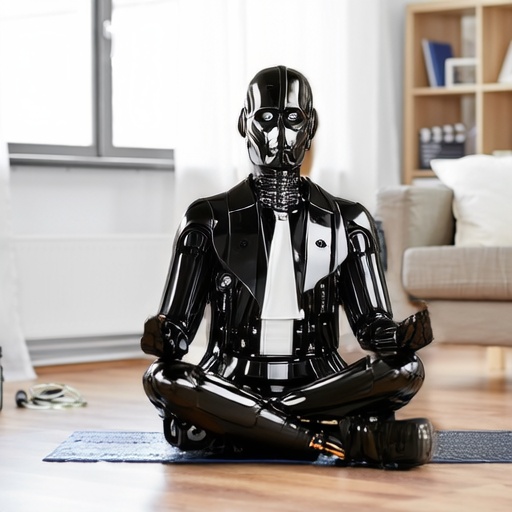} & 
    \includegraphics[width=0.15\linewidth, trim=120 10 80 40, clip]{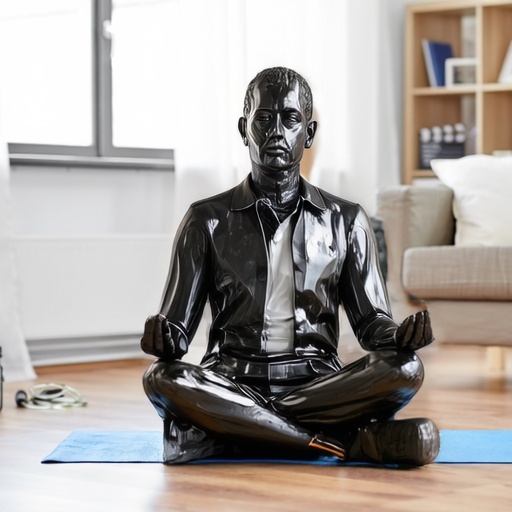} & 
    \includegraphics[width=0.15\linewidth, trim=120 10 80 40, clip]{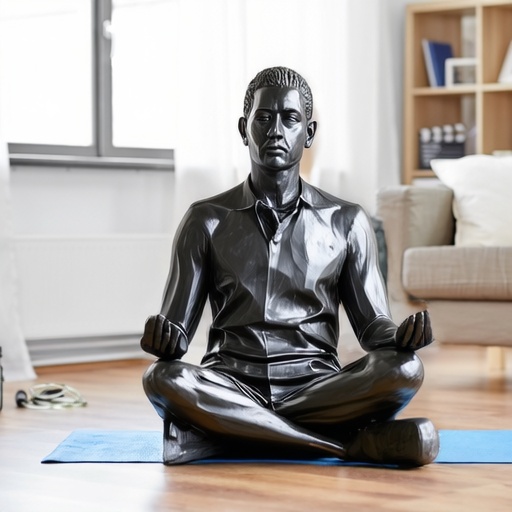} &
    \includegraphics[width=0.15\linewidth, trim=120 10 80 40, clip]{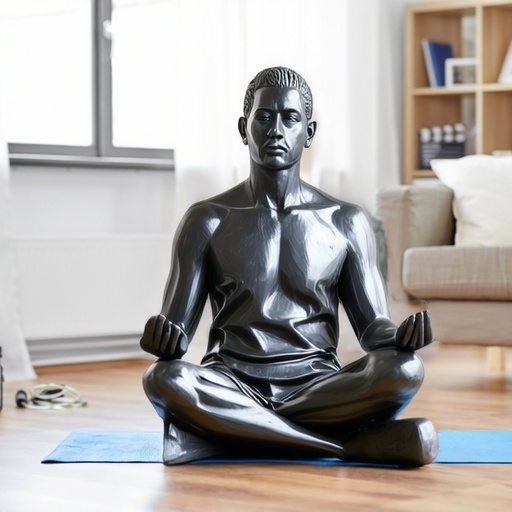} &
    \includegraphics[width=0.15\linewidth, trim=120 10 80 40, clip]{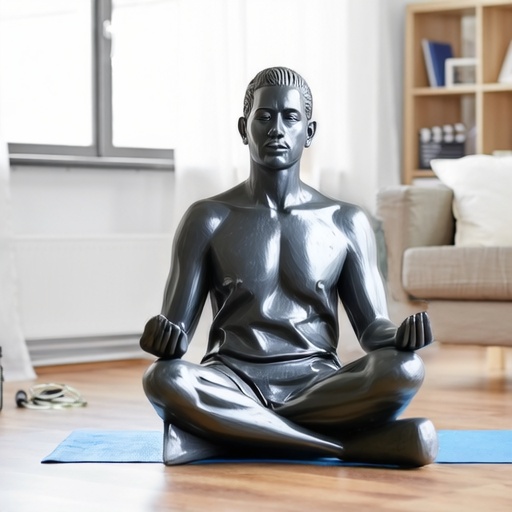} &
    \includegraphics[width=0.15\linewidth, trim=120 10 80 40, clip]{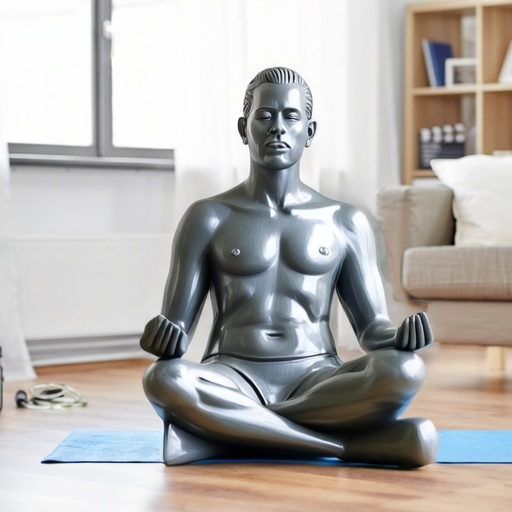} \\
    \raisebox{21pt}{\rotatebox[origin=t]{90}{\small{CFG-\REC}}} & { } &
    \includegraphics[width=0.15\linewidth, trim=120 10 80 40, clip]{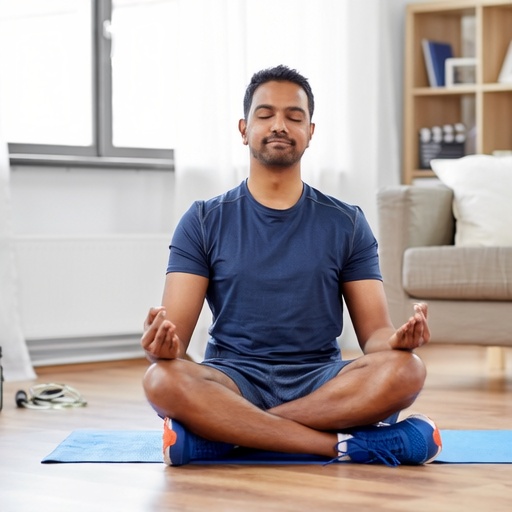} & 
    \includegraphics[width=0.15\linewidth, trim=120 10 80 40, clip]{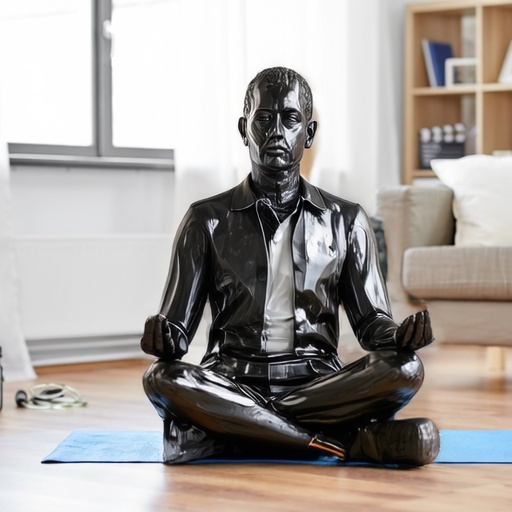} & 
    \includegraphics[width=0.15\linewidth, trim=120 10 80 40, clip]{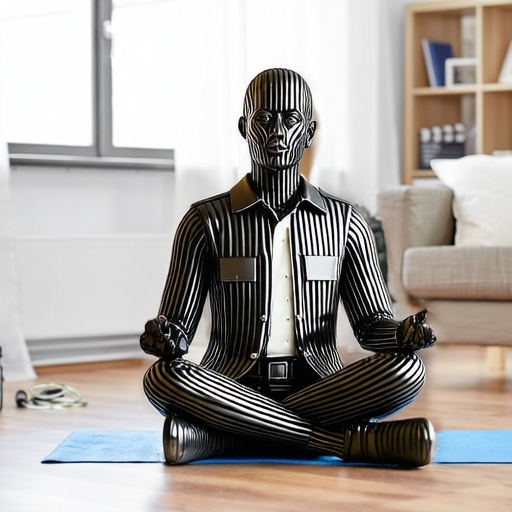} &
    \includegraphics[width=0.15\linewidth, trim=120 10 80 40, clip]{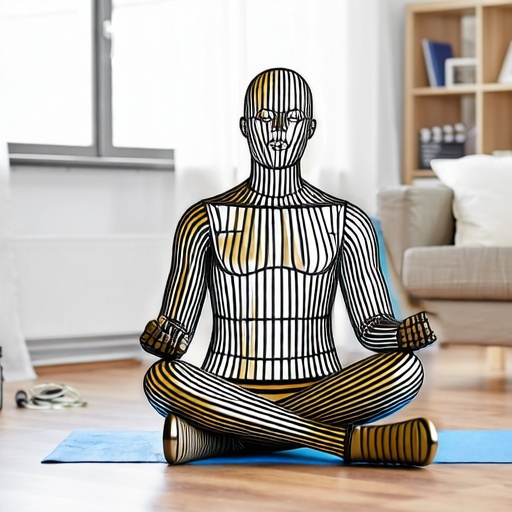} &
    \includegraphics[width=0.15\linewidth, trim=120 10 80 40, clip]{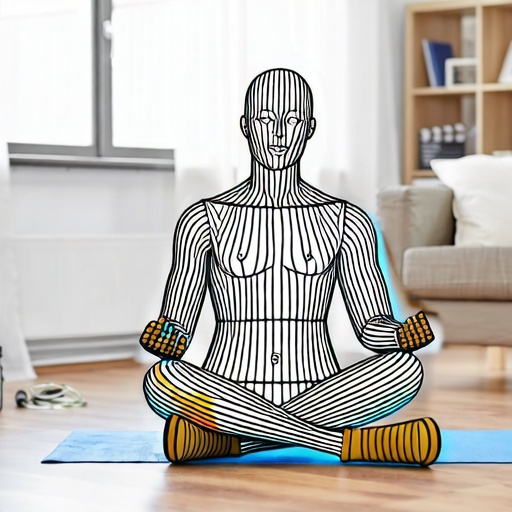} &
    \includegraphics[width=0.15\linewidth, trim=120 10 80 40, clip]{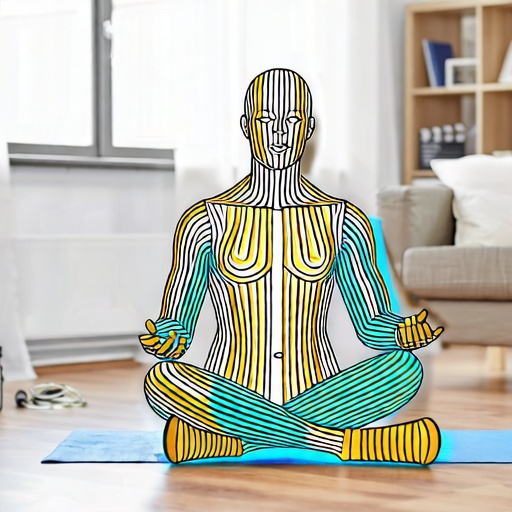} \\
    \raisebox{21pt}{\rotatebox[origin=t]{90}{\small{Lin. Scheduler}}} & { } &
    \includegraphics[width=0.15\linewidth, trim=120 10 80 40, clip]{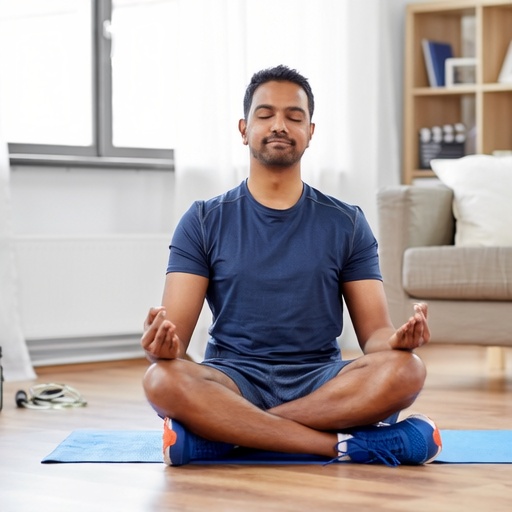} & 
    \includegraphics[width=0.15\linewidth, trim=120 10 80 40, clip]{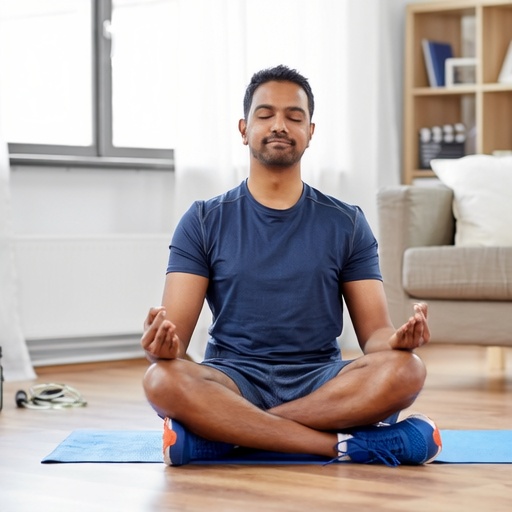} &
    \includegraphics[width=0.15\linewidth, trim=120 10 80 40, clip]{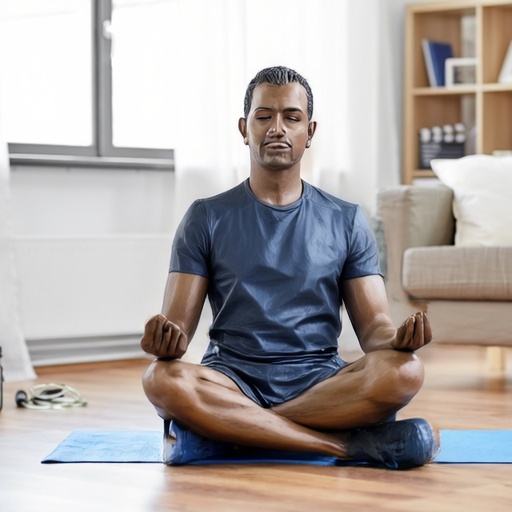} &
    \includegraphics[width=0.15\linewidth, trim=120 10 80 40, clip]{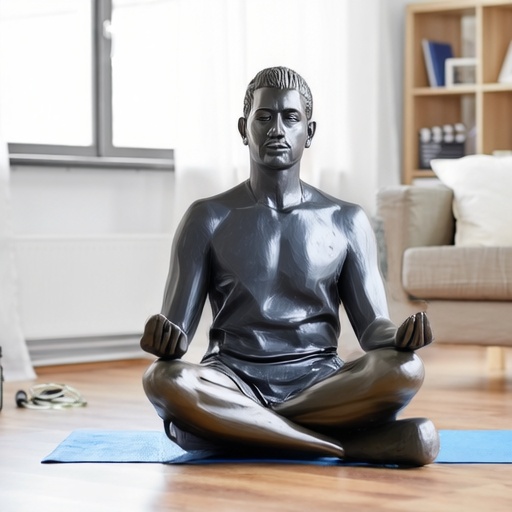} &
    \includegraphics[width=0.15\linewidth, trim=120 10 80 40, clip]{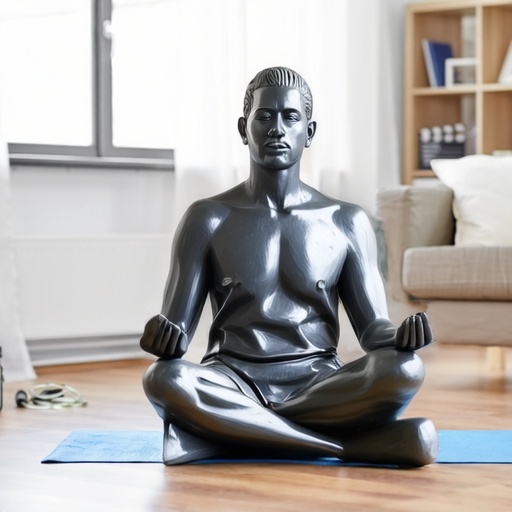} & 
    \includegraphics[width=0.15\linewidth, trim=120 10 80 40, clip]{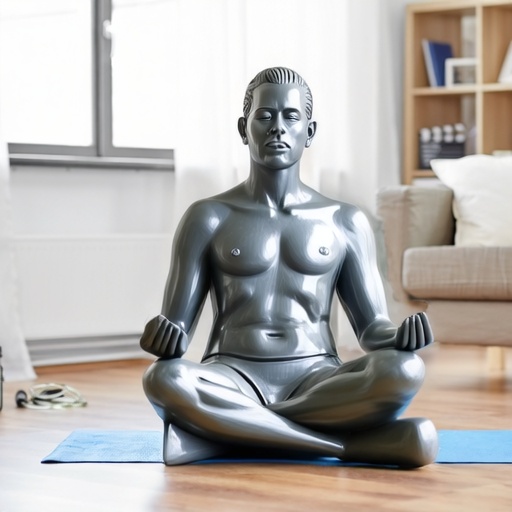} \\
    \raisebox{21pt}{\rotatebox[origin=t]{90}{\small{AdaOr (Ours)}}} & { } &
    \includegraphics[width=0.15\linewidth, trim=120 10 80 40, clip]{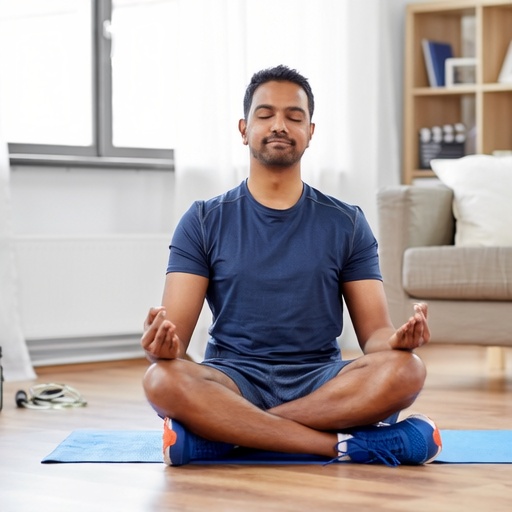} & 
    \includegraphics[width=0.15\linewidth, trim=120 10 80 40, clip]{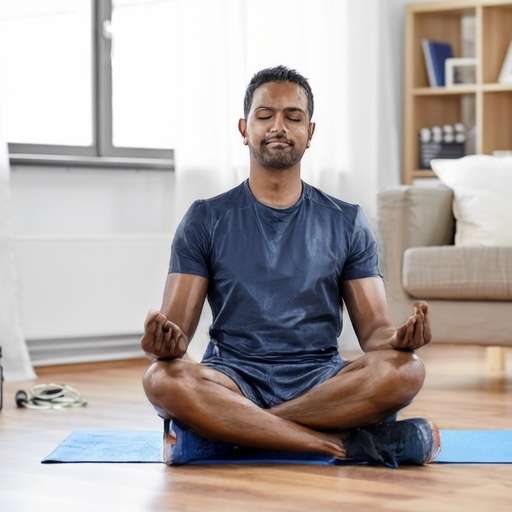} &
    \includegraphics[width=0.15\linewidth, trim=120 10 80 40, clip]{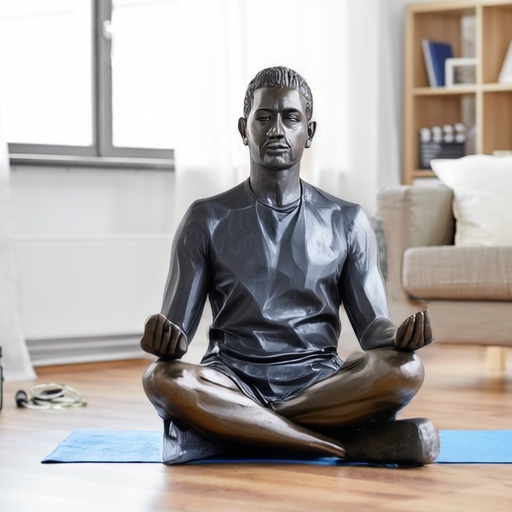} &
    \includegraphics[width=0.15\linewidth, trim=120 10 80 40, clip]{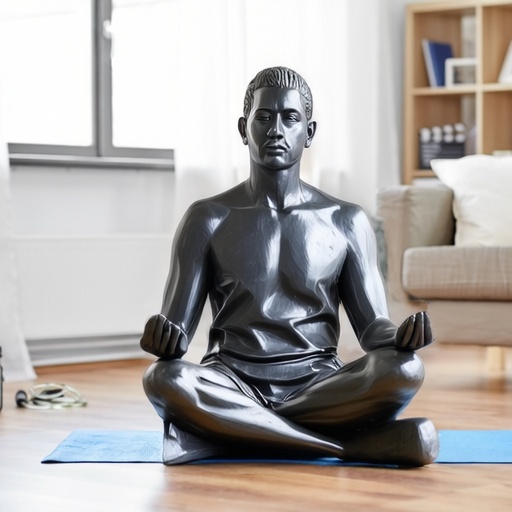} &
    \includegraphics[width=0.15\linewidth, trim=120 10 80 40, clip]{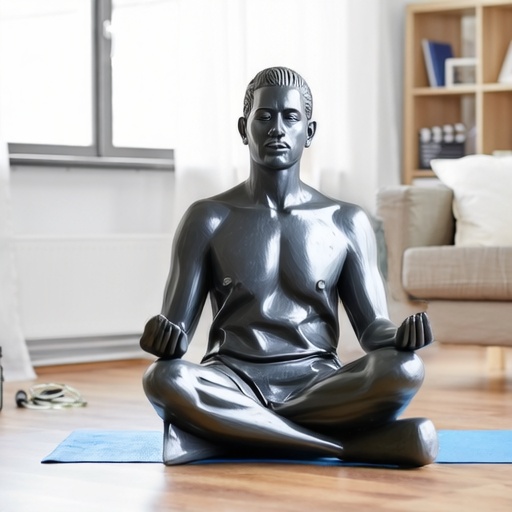} & 
    \includegraphics[width=0.15\linewidth, trim=120 10 80 40, clip]{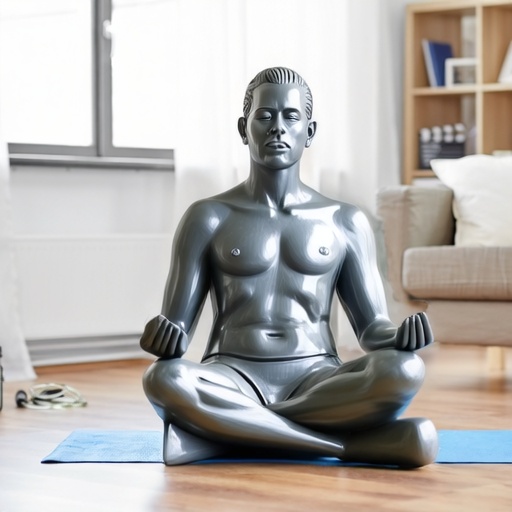} \\
    && 0.0 & 0.2 & 0.4 & 0.6 & 0.8 & 1.0
    \\

    \end{tabular}
    \vspace{-10pt}
    \caption{\textbf{Qualitative results of the ablation study.}
    Note that the input image is displayed in the first column of rows 2, 3, and 4.
    Standard CFG (top) yields arbitrary low-scale content.
    CFG-\REC{} (row 2) diverges at high scales, while the linear scheduler (row 3) exhibits inconsistent change magnitudes.
    Our AdaOr (bottom) ensures a smooth, consistent transition.}
    \vspace{-10pt}
    \label{fig:ablation_full}
\end{figure}

%% file: figures/10_ablation_full/instruction.txt
Transform the man into a plastic sculpture.

%% file: figures/11_limitations/limitations.tex
\begin{figure}
    \centering
    \setlength{\tabcolsep}{0pt}
    \begin{tabular}{ccccc}
    
    \multicolumn{5}{c}{\textit{\small{``Change the cartoon boy into a cartoon girl and make the car a taxi.''}}} \\
    \includegraphics[width=0.19\linewidth]{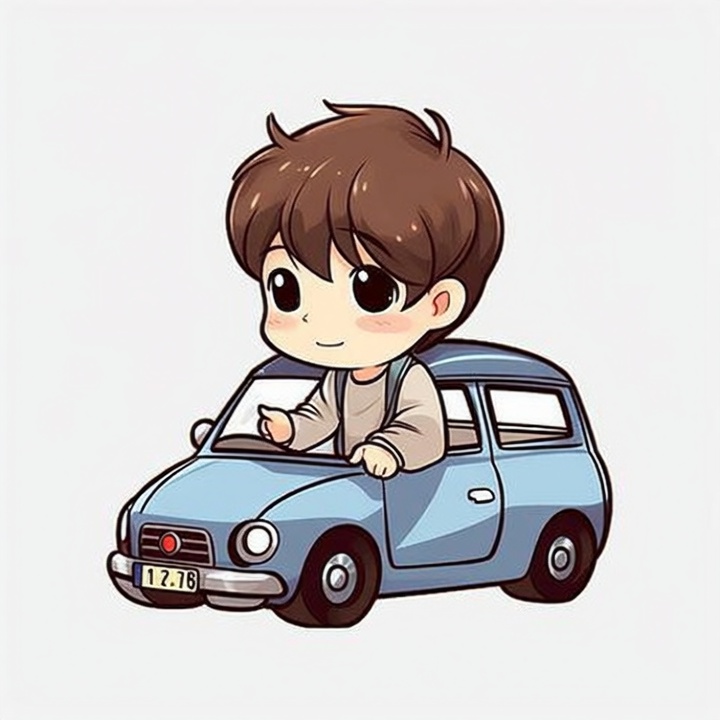} & 
    \includegraphics[width=0.19\linewidth]{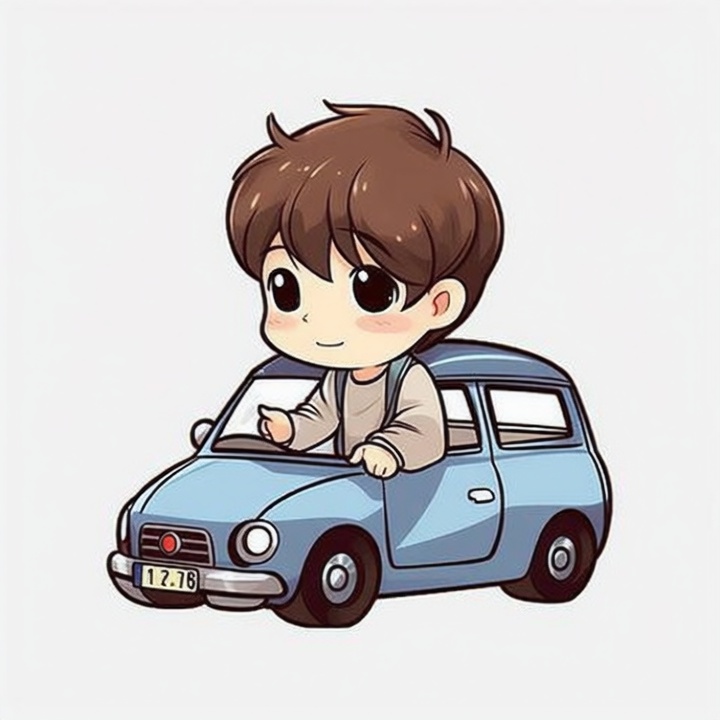} &
    \includegraphics[width=0.19\linewidth]{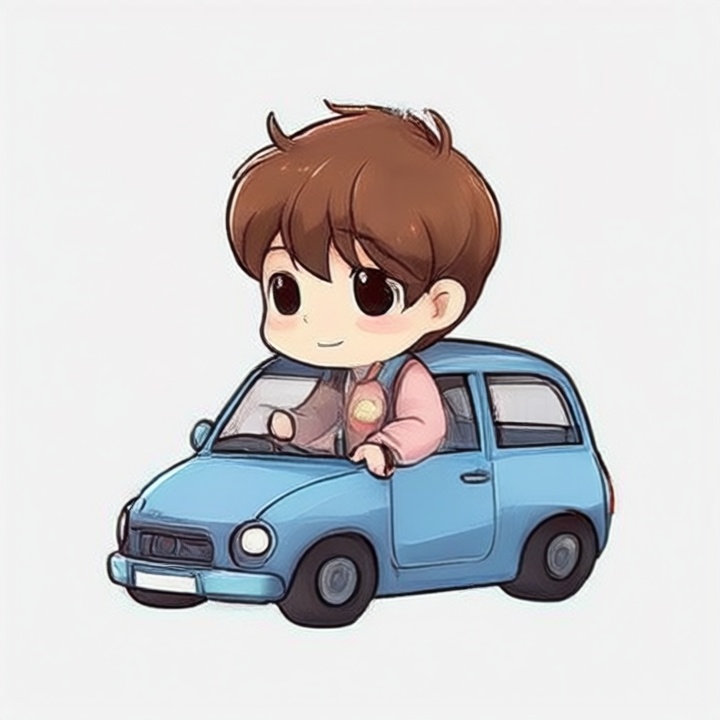} &
    \includegraphics[width=0.19\linewidth]{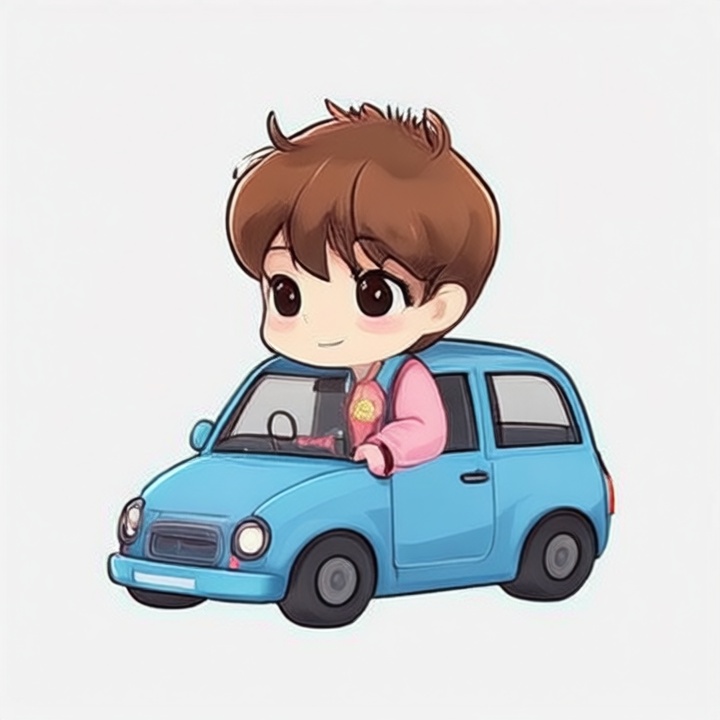} &
    \includegraphics[width=0.19\linewidth]{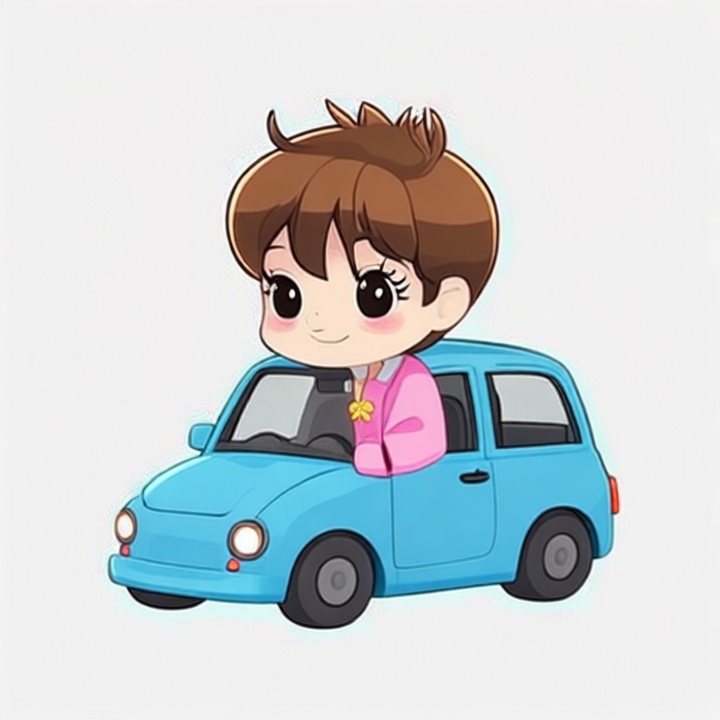}  \\
    \multicolumn{5}{c}{\textit{\small{``Change the dog to be lying on its side and have it look at the camera.''}}} \\
    \includegraphics[width=0.19\linewidth]{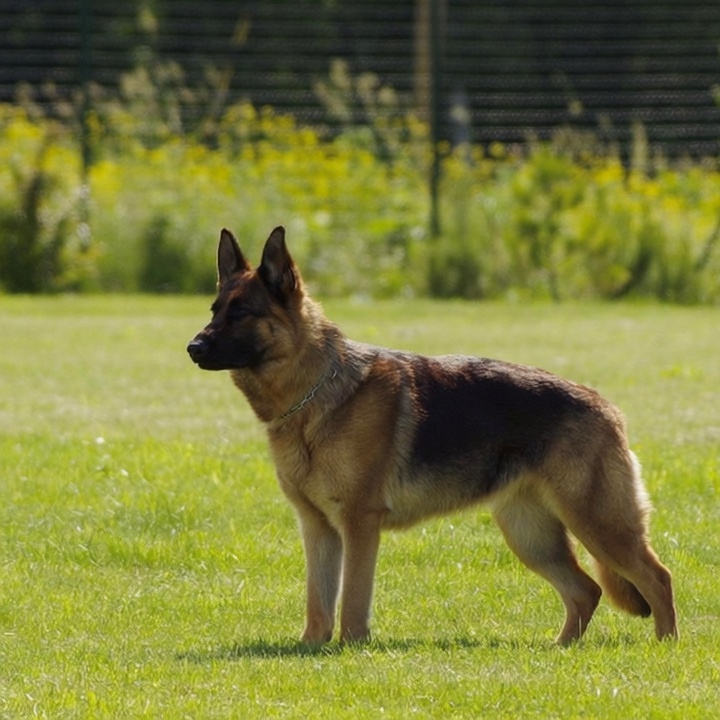} & 
    \includegraphics[width=0.19\linewidth]{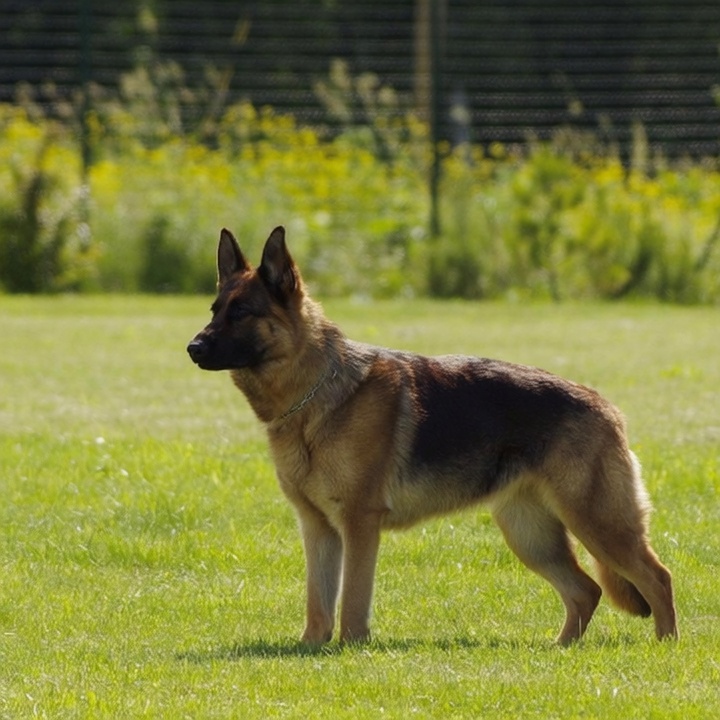} &
    \includegraphics[width=0.19\linewidth]{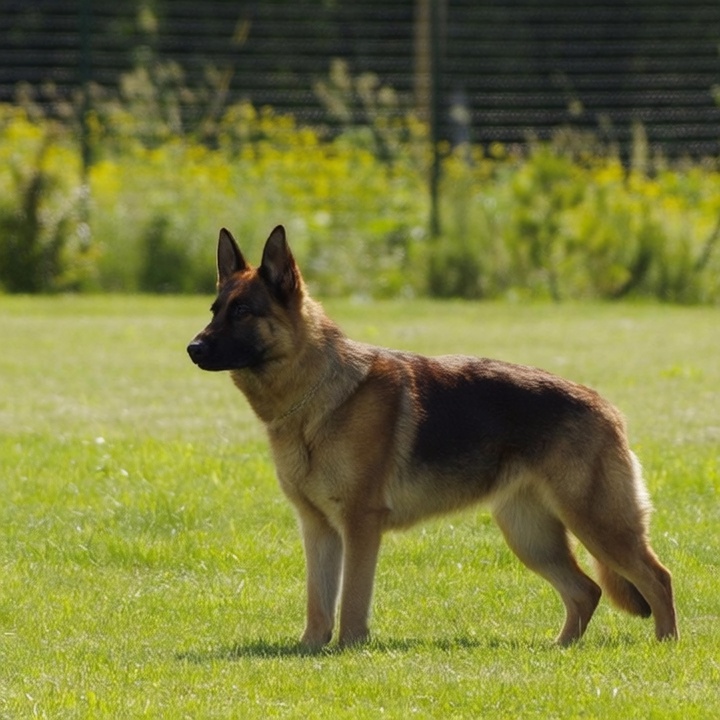} &
    \includegraphics[width=0.19\linewidth]{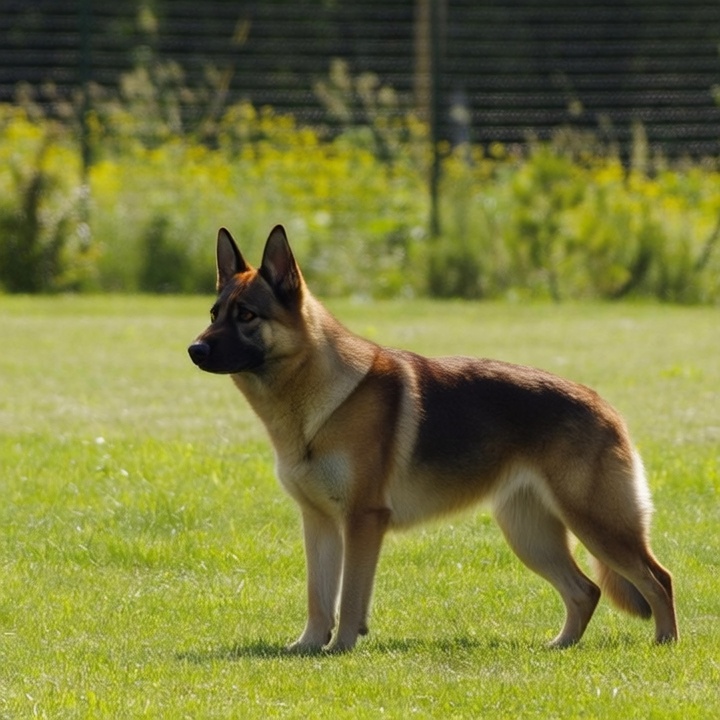} &
    \includegraphics[width=0.19\linewidth]{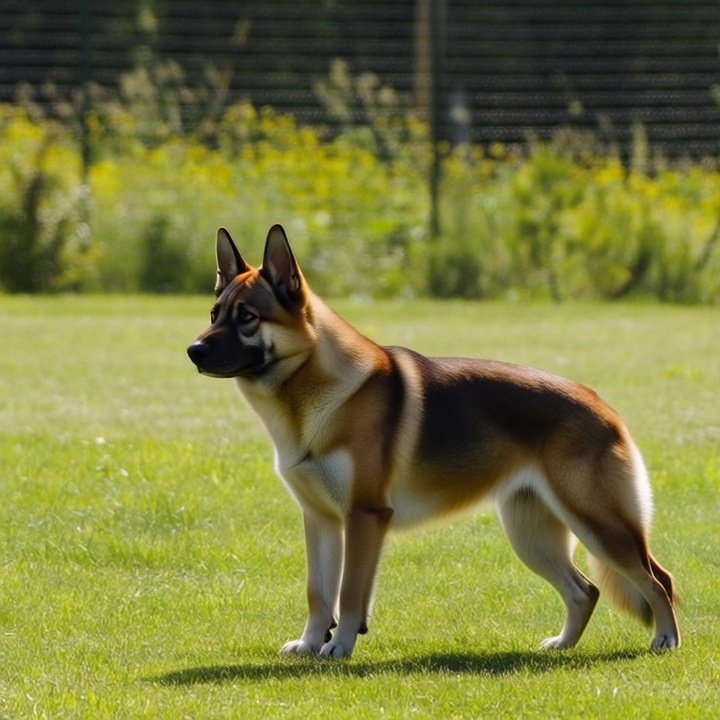}  \\
    \multicolumn{5}{l}{\small{Edit Intensity} $\xrightarrow{\hspace{180pt}}$}

    \end{tabular}
    \vspace{-8pt}
\caption{\textbf{Limitations.}
We illustrate cases where the backbone model fails to execute the target edit.
In the top row, the car fails to transform into a ``taxi''.
In the bottom row, the dog fails to adopt the ``lying down'' pose.
Notably, even in these failure cases, our method generates a smooth and continuous sequence, indicating that the failure stems from the backbone's limited representational scope rather than a failure of our continuity mechanism.}
\vspace{-10pt}
    \label{fig:limitations}
\end{figure}

%% file: 5_conclusion.tex
\section{Conclusions}

We presented Adaptive-Origin Guidance (AdaOr), a method for generating smooth, continuous editing sequences from diffusion-based editing models. Our approach is grounded in the observation that at low guidance scales, the generation with CFG is dominated by the unconditional prediction. In the context of editing models, this unconditional prediction corresponds to arbitrary edits rather than an identity edit. By introducing a learnable identity instruction (\REC{}), we establish a semantically valid guidance origin, allowing for linear interpolation between the input image and the target edit.

Our method does not require a specialized dataset for continuous edits, nor does it rely on per-edit-type procedures. We demonstrate that explicitly teaching the model the identity instruction facilitates arithmetic in the prediction space, enabling the execution of complex downstream tasks (e.g., continuous editing). 
We believe this approach holds further potential for additional generative applications. 

For future work, we believe our method can serve as a data generation engine. The continuous sequences produced by our approach could be used to train next-generation editing models that incorporate edit strength directly as a conditioning parameter, distilling inference-time guidance control into native architectural capability.

%% file: more_results/additional_results1.tex
\newcommand{\editrow}[1]{
    \multicolumn{8}{c}{``\textit{\input{figures/2_results-images/images/pie-bench/#1/instruction.txt}}''} \\
    \includegraphics[width=0.14\linewidth]{figures/2_results-images/images/pie-bench/#1/0.0.jpg} & { } &
    \includegraphics[width=0.14\linewidth]{figures/2_results-images/images/pie-bench/#1/0.0.jpg} &
    \includegraphics[width=0.14\linewidth]{figures/2_results-images/images/pie-bench/#1/0.2.jpg} &
    \includegraphics[width=0.14\linewidth]{figures/2_results-images/images/pie-bench/#1/0.4.jpg} &
    \includegraphics[width=0.14\linewidth]{figures/2_results-images/images/pie-bench/#1/0.6.jpg} &
    \includegraphics[width=0.14\linewidth]{figures/2_results-images/images/pie-bench/#1/0.8.jpg} &
    \includegraphics[width=0.14\linewidth]{figures/2_results-images/images/pie-bench/#1/1.0.jpg} \\
}

\newcommand{\editrowmore}[1]{
    \includegraphics[width=0.11\linewidth]{more_results/#1/alpha/0.000.jpg} & { } &
    \includegraphics[width=0.11\linewidth]{more_results/#1/alpha/0.000.jpg} &
    \includegraphics[width=0.11\linewidth]{more_results/#1/alpha/0.143.jpg} &
    \includegraphics[width=0.11\linewidth]{more_results/#1/alpha/0.286.jpg} &
    \includegraphics[width=0.11\linewidth]{more_results/#1/alpha/0.429.jpg} &
    \includegraphics[width=0.11\linewidth]{more_results/#1/alpha/0.571.jpg} &
    \includegraphics[width=0.11\linewidth]{more_results/#1/alpha/0.714.jpg} &
    \includegraphics[width=0.11\linewidth]{more_results/#1/alpha/0.857.jpg} &
    \includegraphics[width=0.11\linewidth]{more_results/#1/alpha/1.000.jpg} \\
}

\begin{figure*}
    \centering
    \setlength{\tabcolsep}{0pt}
    \begin{tabular}{cc cccccccc}
        \multicolumn{10}{c}{\textit{``add snow covering the ground making it look like winter and add the northern lights.''}} \\
        \editrowmore{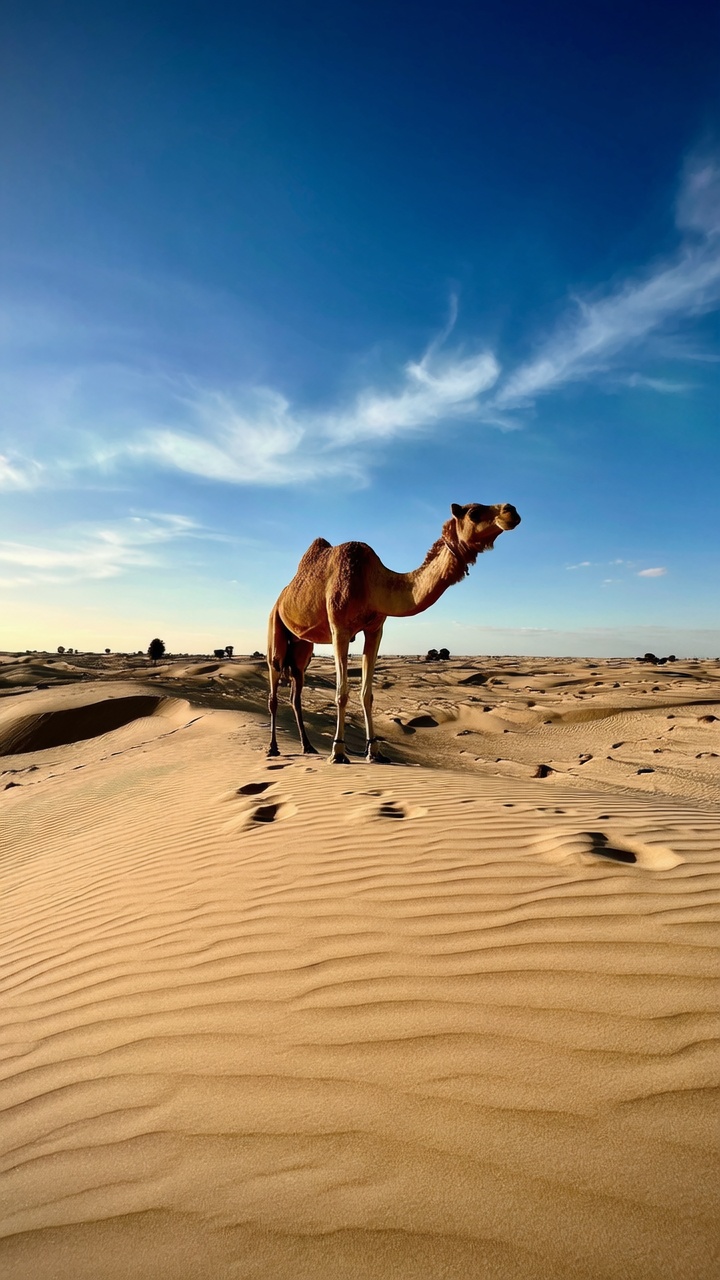}
        \multicolumn{10}{c}{\textit{``Transform the teddy bear into a robotic bear with visible mechanical parts, metallic surfaces, and subtle joints.''}} \\
        \editrowmore{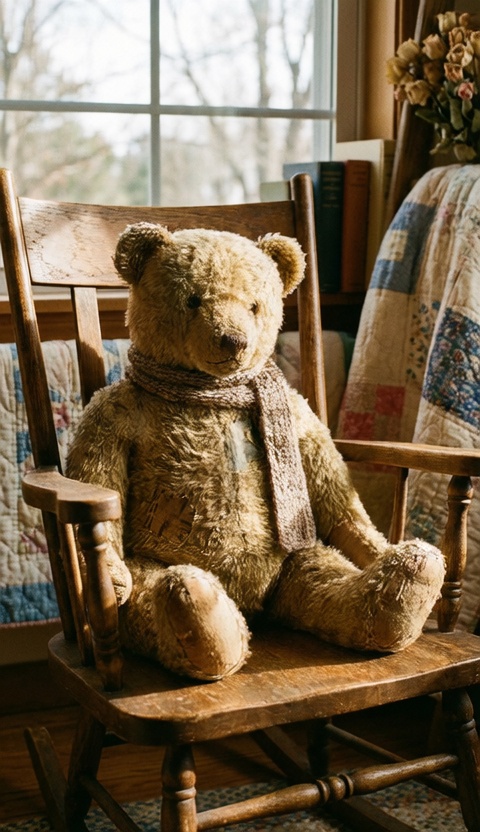}
    \end{tabular}
    \label{fig:additional_results_2}
\end{figure*}

\newcommand{\editrowmorehor}[1]{
    \includegraphics[width=0.11\linewidth]{more_results/#1/alpha/0.000.jpg} & { } &
    \includegraphics[width=0.11\linewidth]{more_results/#1/alpha/0.000.jpg} &
    \includegraphics[width=0.11\linewidth]{more_results/#1/alpha/0.143.jpg} &
    \includegraphics[width=0.11\linewidth]{more_results/#1/alpha/0.286.jpg} &
    \includegraphics[width=0.11\linewidth]{more_results/#1/alpha/0.429.jpg} &
    \includegraphics[width=0.11\linewidth]{more_results/#1/alpha/0.571.jpg} &
    \includegraphics[width=0.11\linewidth]{more_results/#1/alpha/0.714.jpg} &
    \includegraphics[width=0.11\linewidth]{more_results/#1/alpha/0.857.jpg} &
    \includegraphics[width=0.11\linewidth]{more_results/#1/alpha/1.000.jpg} \\
}

\begin{figure*}
    \centering
    \setlength{\tabcolsep}{0pt}
    \begin{tabular}{cc cccccccc}
        \multicolumn{10}{c}{\textit{``the season is autumn the floor is covered with orange leaves.''}} \\
        \editrowmorehor{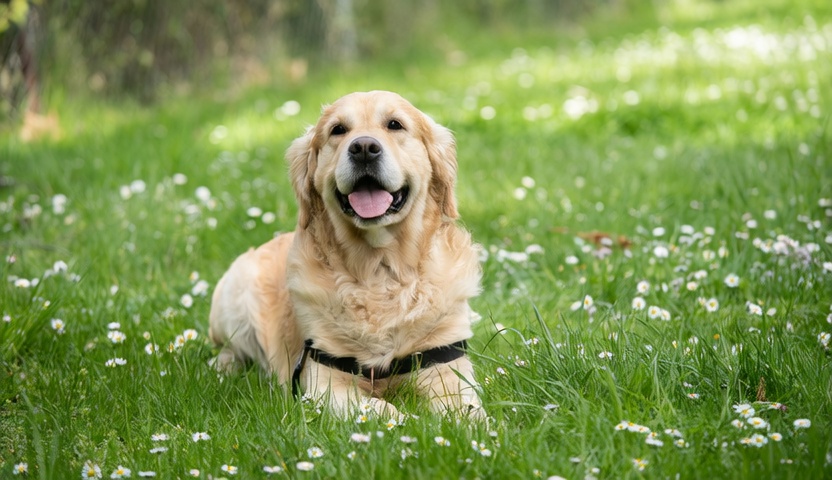}
        \multicolumn{10}{c}{\textit{``Transform the horse into a carved wooden horse.''}} \\
        \editrowmorehor{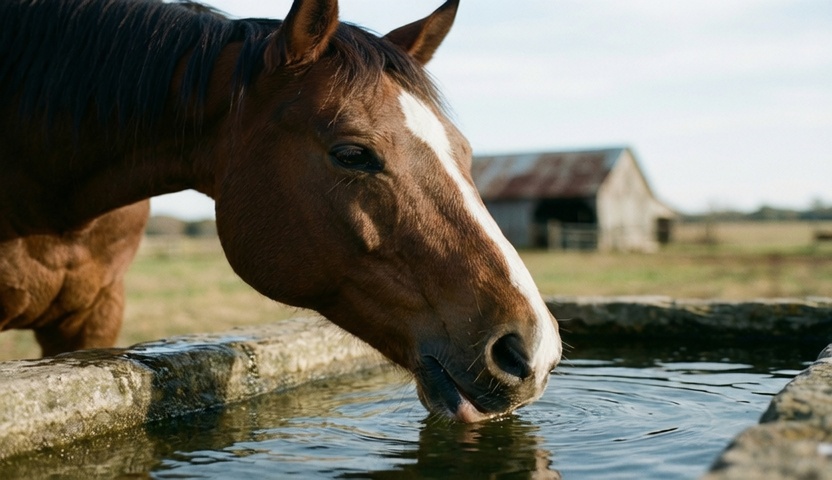}
        Input && 0.000 & 0.143 & 0.286 & 0.429 & 0.571 & 0.714 & 0.857 & 1.000
    \end{tabular}
    \label{fig:additional_results_3}
\end{figure*}

\begin{figure*}
    \centering
    \setlength{\tabcolsep}{0pt}
    \begin{tabular}{cc cccccc}
        
    \multicolumn{8}{c}{``\textit{\input{figures/2_results-images/images/pie-bench/69/instruction.txt}}''} \\
    \includegraphics[width=0.14\linewidth]{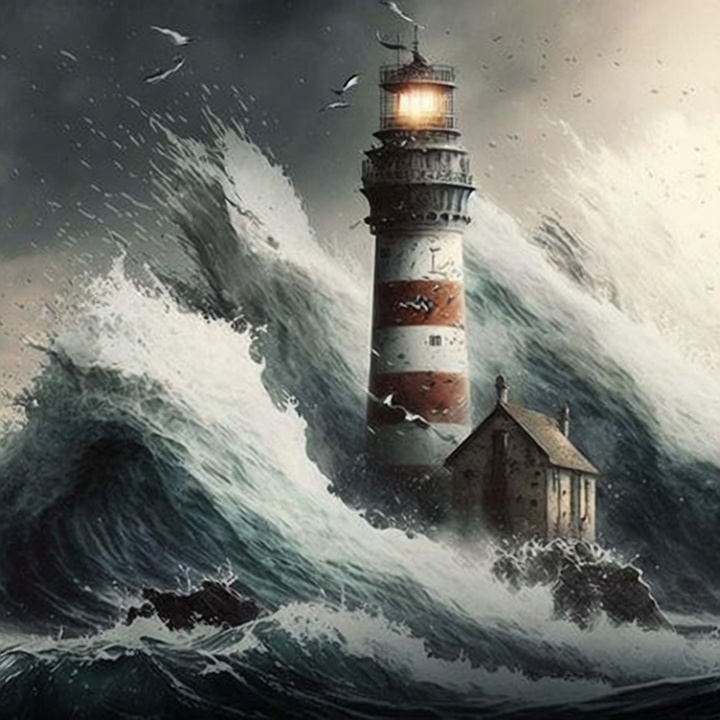} & { } &
    \includegraphics[width=0.14\linewidth]{figures/2_results-images/images/pie-bench/69/0.0.jpg} &
    \includegraphics[width=0.14\linewidth]{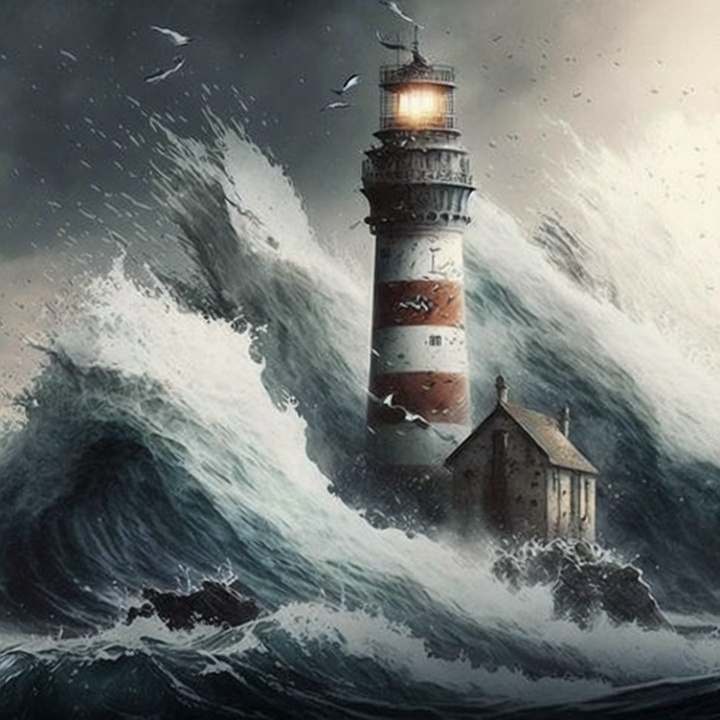} &
    \includegraphics[width=0.14\linewidth]{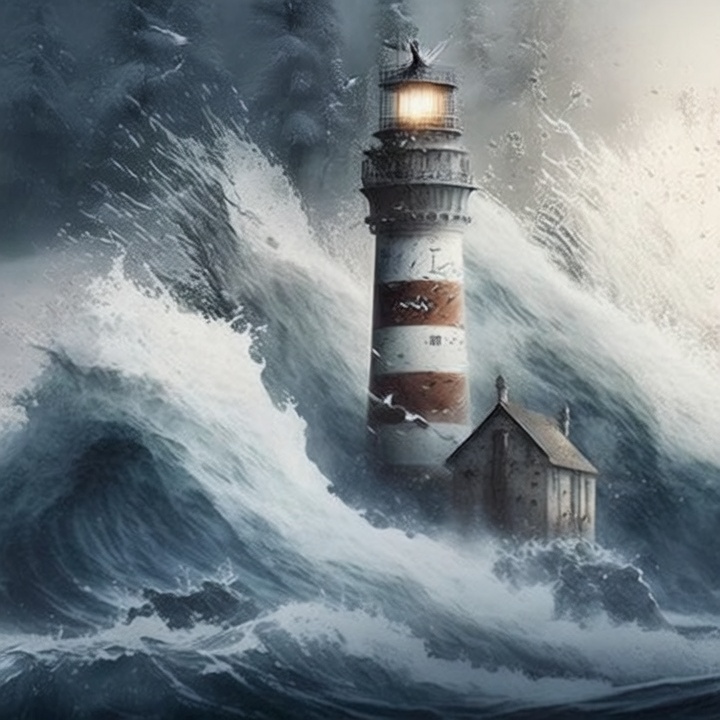} &
    \includegraphics[width=0.14\linewidth]{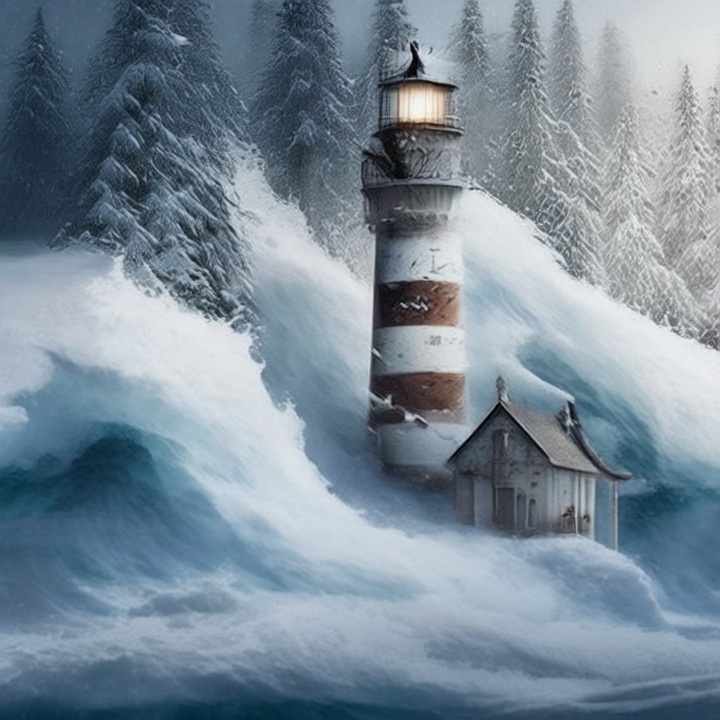} &
    \includegraphics[width=0.14\linewidth]{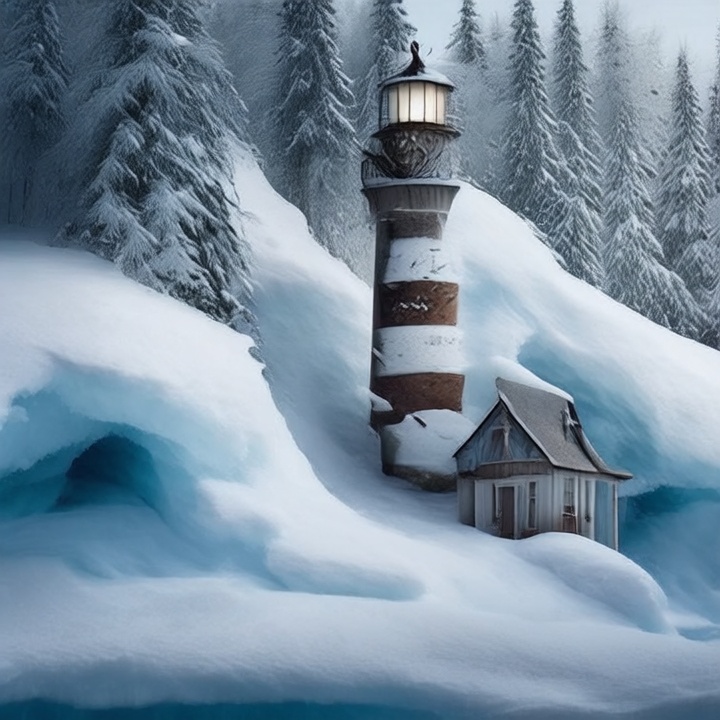} &
    \includegraphics[width=0.14\linewidth]{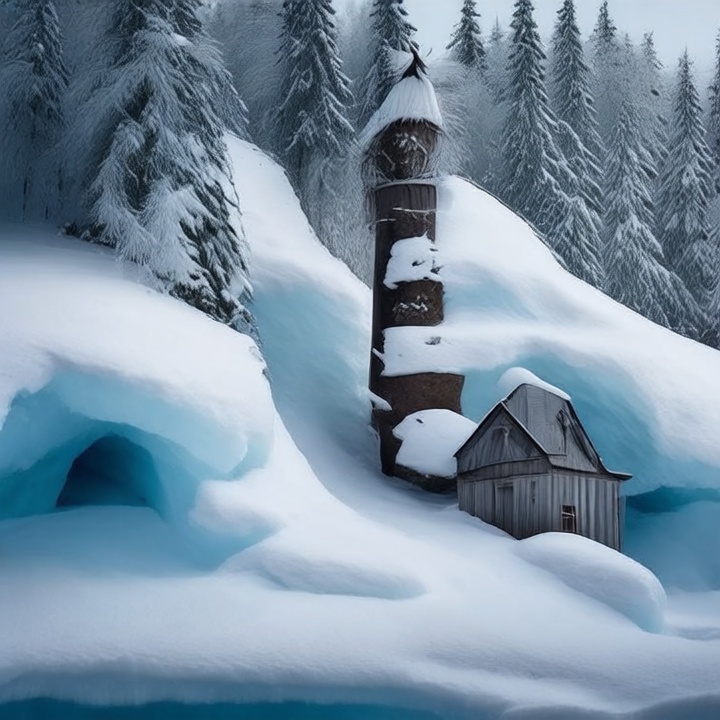} \\

    \multicolumn{8}{c}{``\textit{\input{figures/2_results-images/images/pie-bench/74/instruction.txt}}''} \\
    \includegraphics[width=0.14\linewidth]{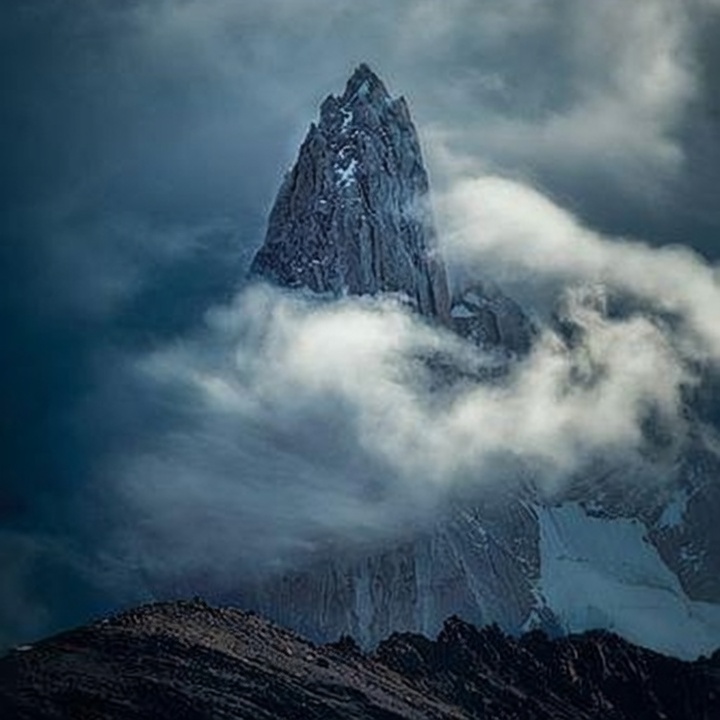} & { } &
    \includegraphics[width=0.14\linewidth]{figures/2_results-images/images/pie-bench/74/0.0.jpg} &
    \includegraphics[width=0.14\linewidth]{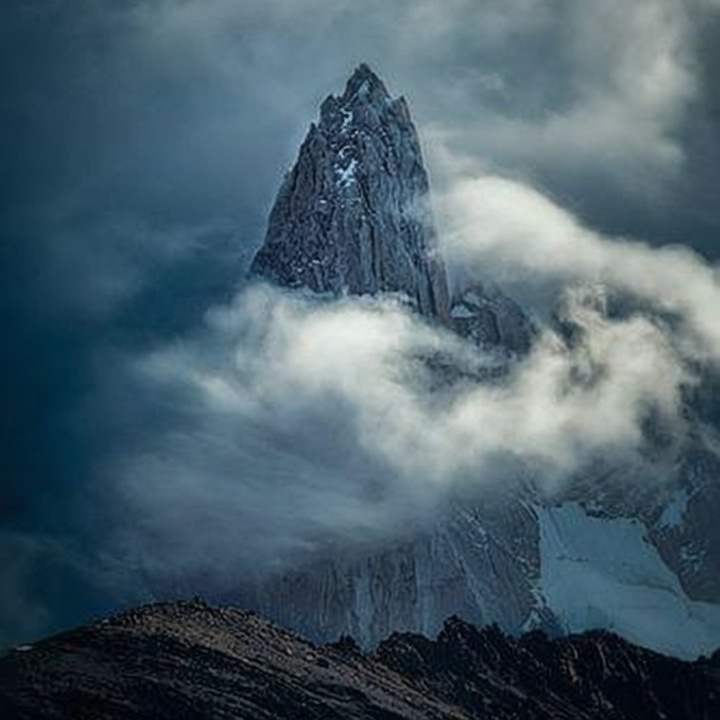} &
    \includegraphics[width=0.14\linewidth]{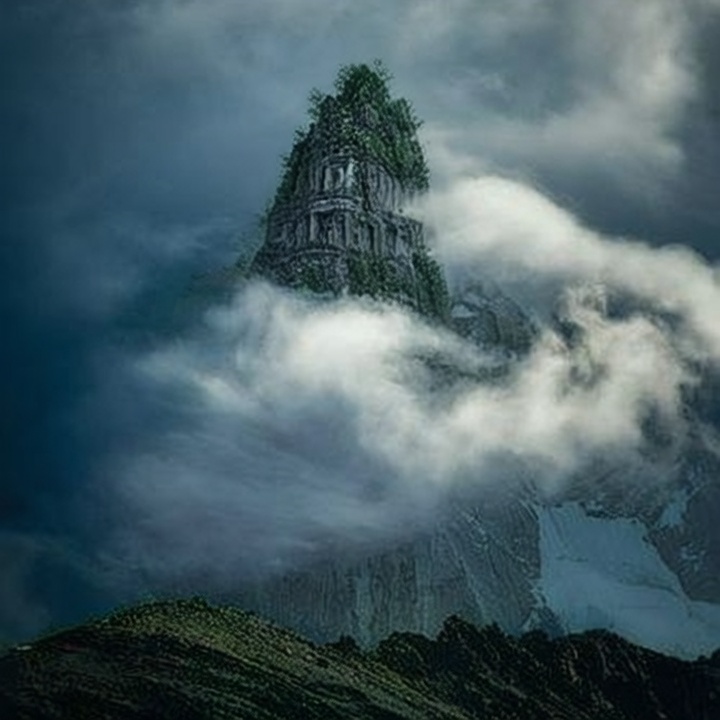} &
    \includegraphics[width=0.14\linewidth]{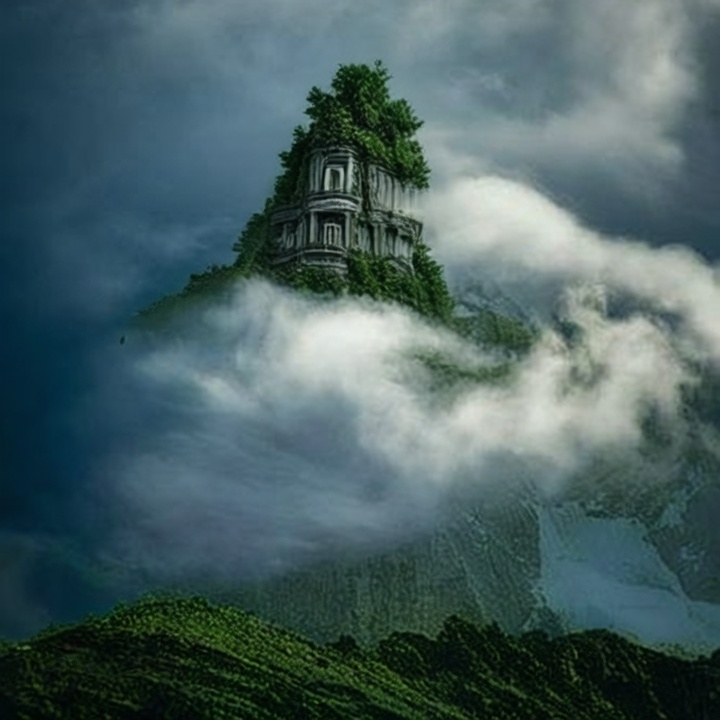} &
    \includegraphics[width=0.14\linewidth]{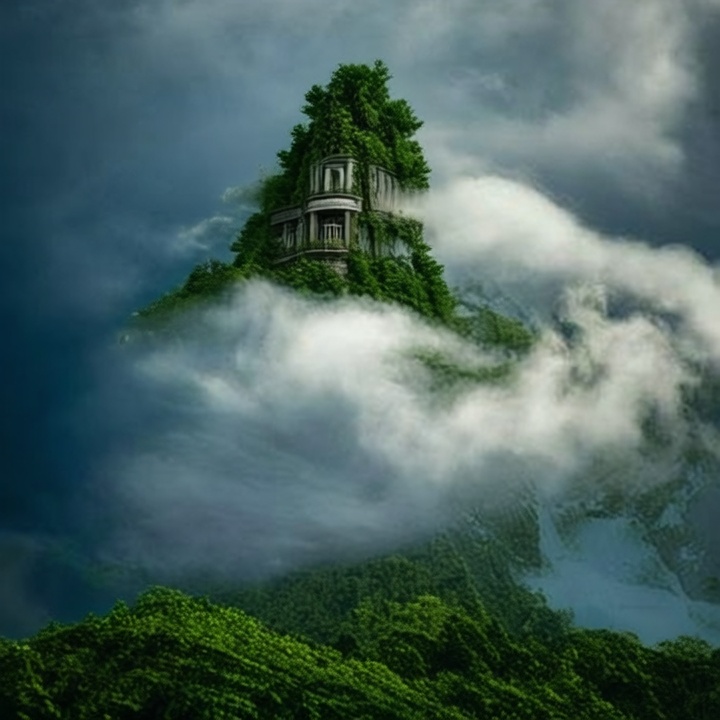} &
    \includegraphics[width=0.14\linewidth]{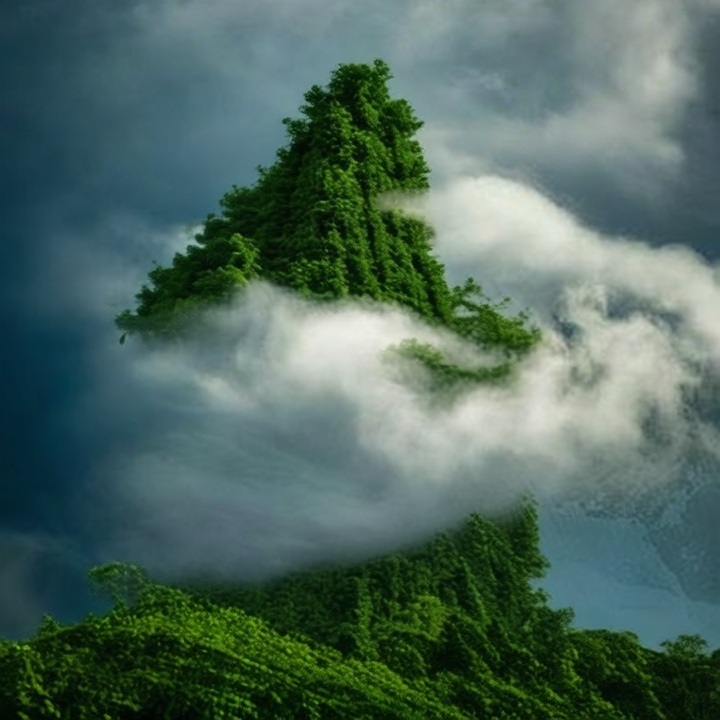} \\

    \multicolumn{8}{c}{``\textit{\input{figures/2_results-images/images/pie-bench/37/instruction.txt}}''} \\
    \includegraphics[width=0.14\linewidth]{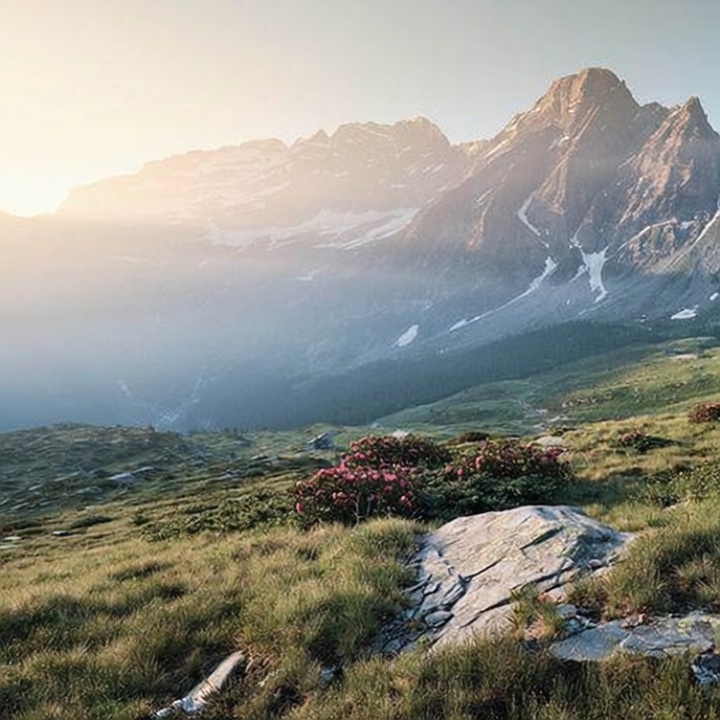} & { } &
    \includegraphics[width=0.14\linewidth]{figures/2_results-images/images/pie-bench/37/0.0.jpg} &
    \includegraphics[width=0.14\linewidth]{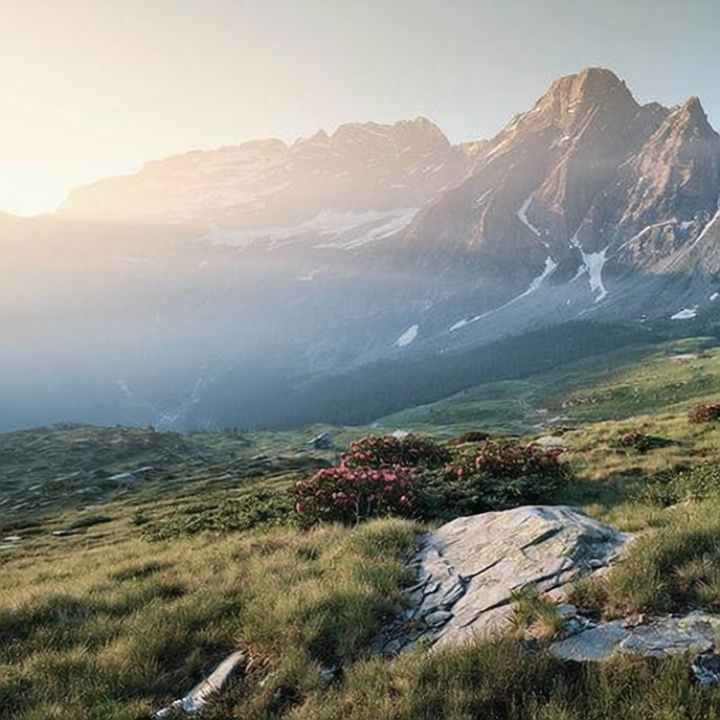} &
    \includegraphics[width=0.14\linewidth]{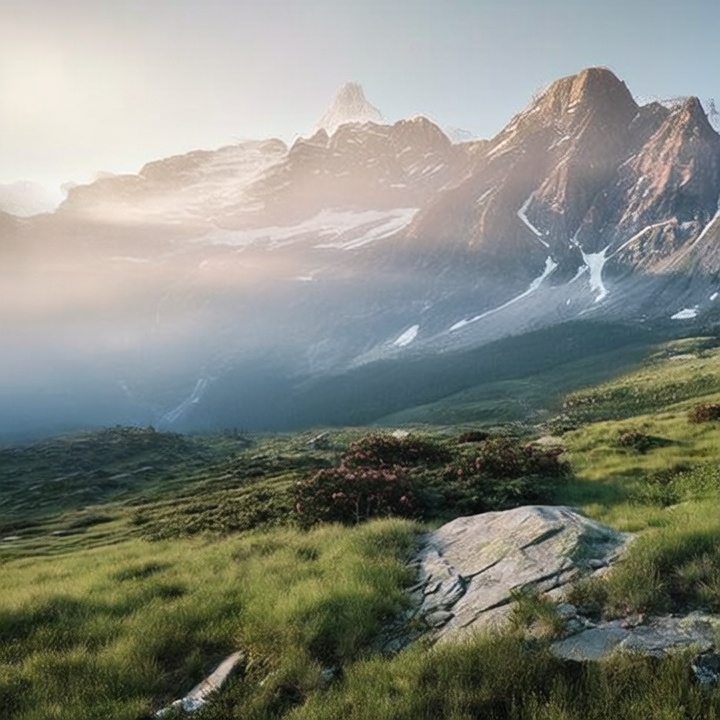} &
    \includegraphics[width=0.14\linewidth]{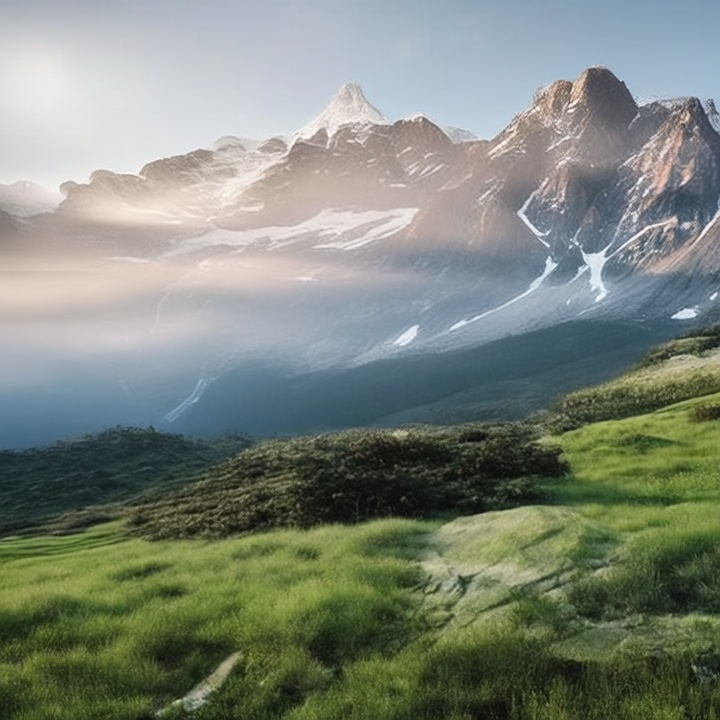} &
    \includegraphics[width=0.14\linewidth]{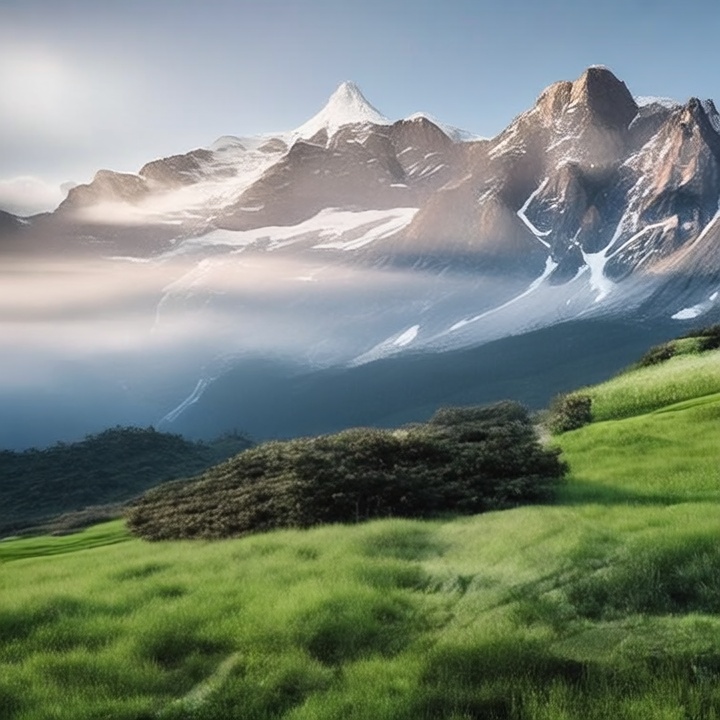} &
    \includegraphics[width=0.14\linewidth]{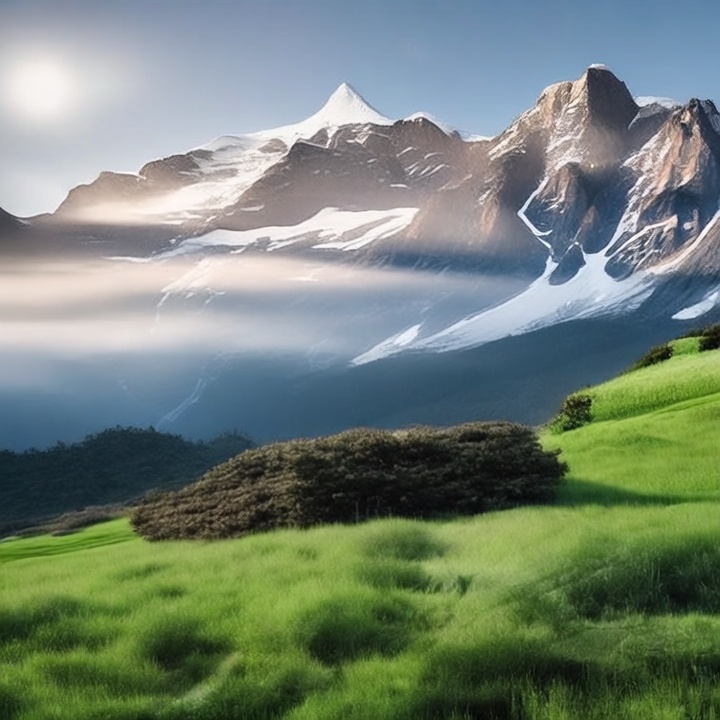} \\

        Input && 0.0 & 0.2 & 0.4 & 0.6 & 0.8 & 1.0
    \end{tabular}
    \label{fig:additional_results_1}
\end{figure*}

%% file: figures/2_results-images/images/pie-bench/69/instruction.txt
Transform the sea and house into a snowy forest scene.

%% file: figures/2_results-images/images/pie-bench/74/instruction.txt
Transform the mountain into a building and cover it in leaves.

%% file: figures/2_results-images/images/pie-bench/37/instruction.txt
Add snow to the mountain landscape and change the grass to green.

%% file: figures/3_results-videos/results-videos.tex
\begin{figure*}
    \centering
    \setlength{\tabcolsep}{1pt}
    \begin{tabular}{c cccccc}
    \multicolumn{7}{c}{\textit{``Add full thick long hair to the man.''}} \\
    {\multirow{3}{*}{{\rotatebox[origin=c]{90}{\hspace{-28pt}Video Frames}}}} &
    \includegraphics[width=0.16\linewidth]{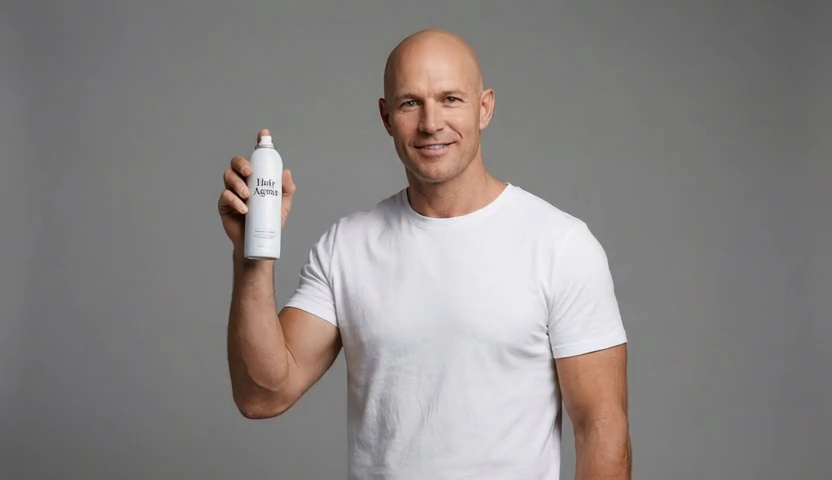} & 
    \includegraphics[width=0.16\linewidth]{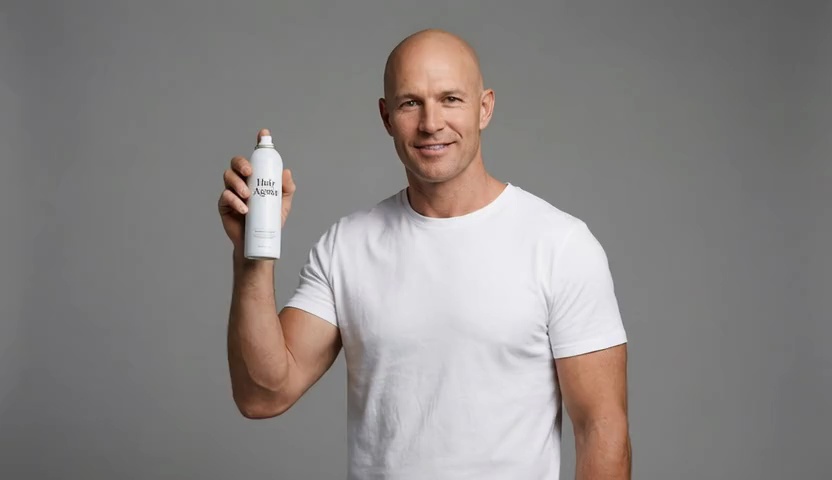} & 
    \includegraphics[width=0.16\linewidth]{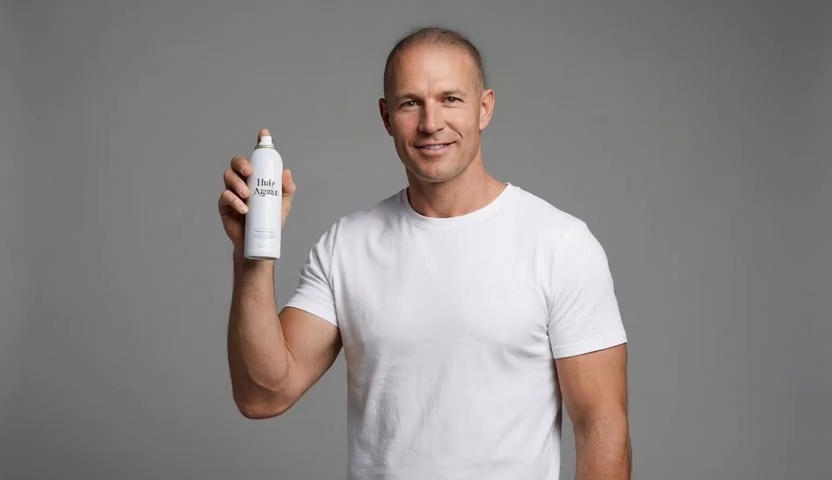} & 
    \includegraphics[width=0.16\linewidth]{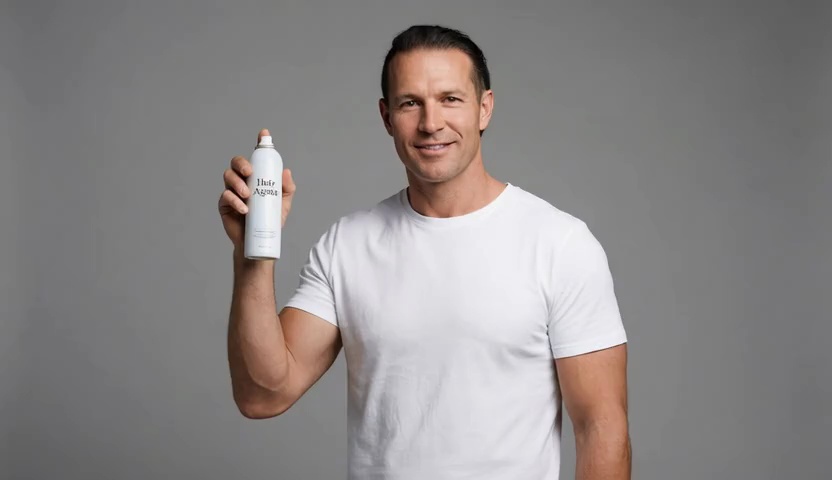} & 
    \includegraphics[width=0.16\linewidth]{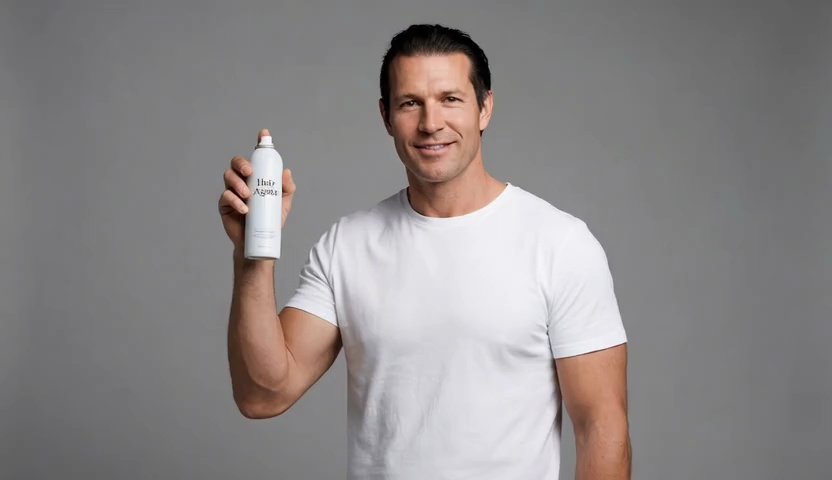} & 
    \includegraphics[width=0.16\linewidth]{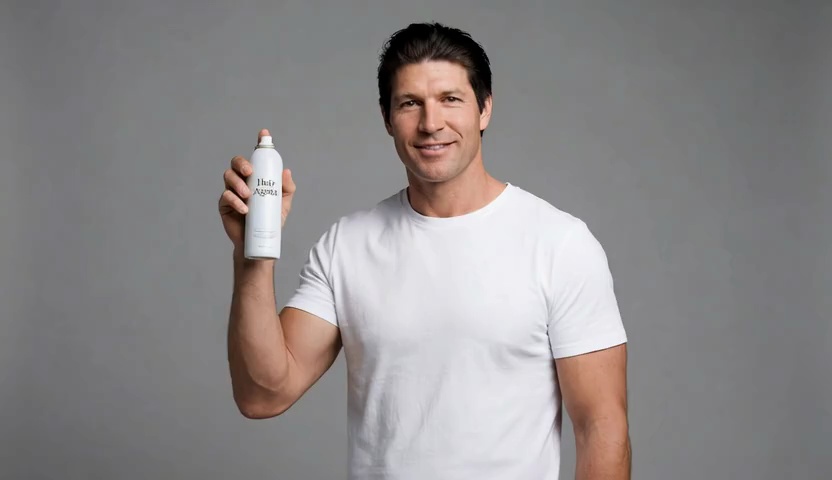} \\[-3.5pt]
    &
    \includegraphics[width=0.16\linewidth]{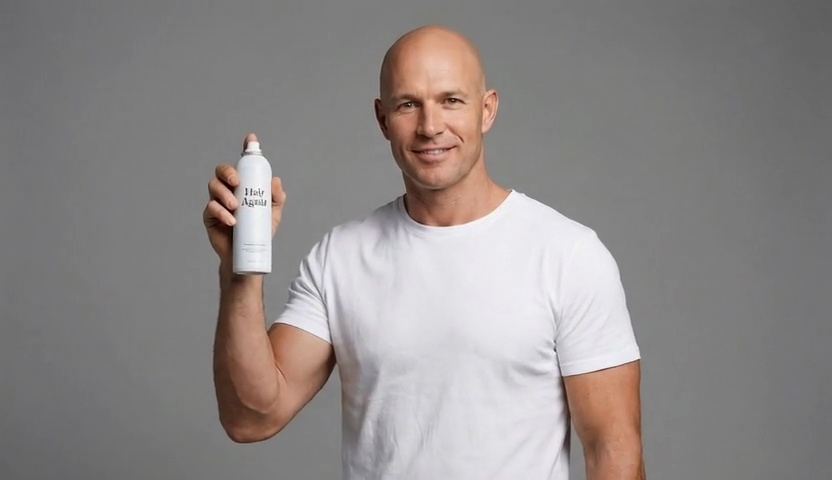} & 
    \includegraphics[width=0.16\linewidth]{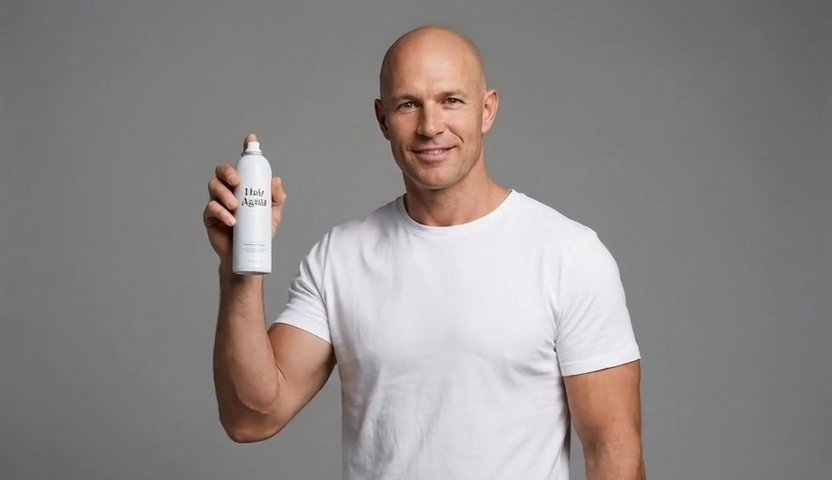} & 
    \includegraphics[width=0.16\linewidth]{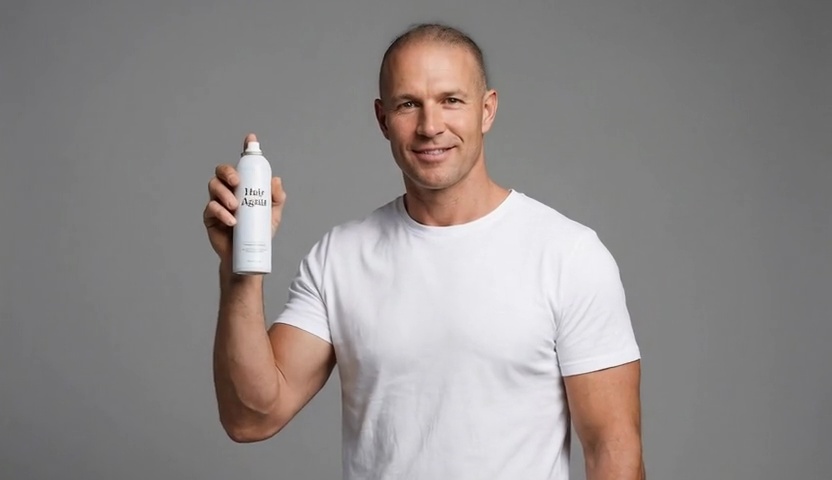} & 
    \includegraphics[width=0.16\linewidth]{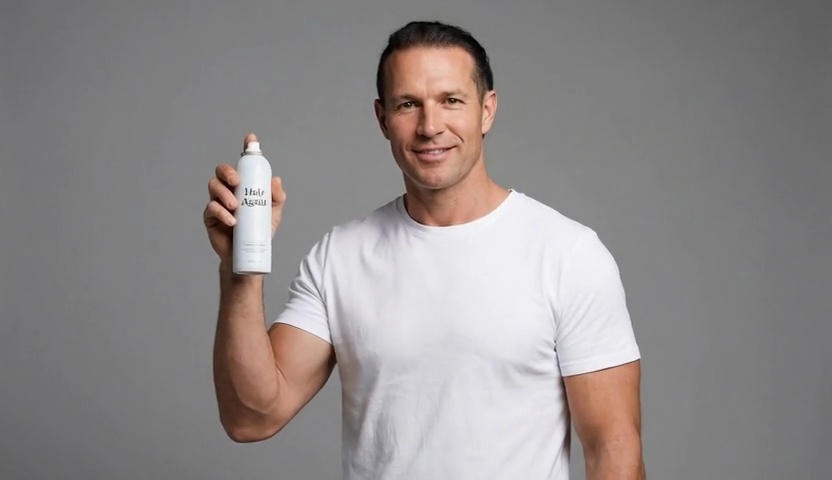} & 
    \includegraphics[width=0.16\linewidth]{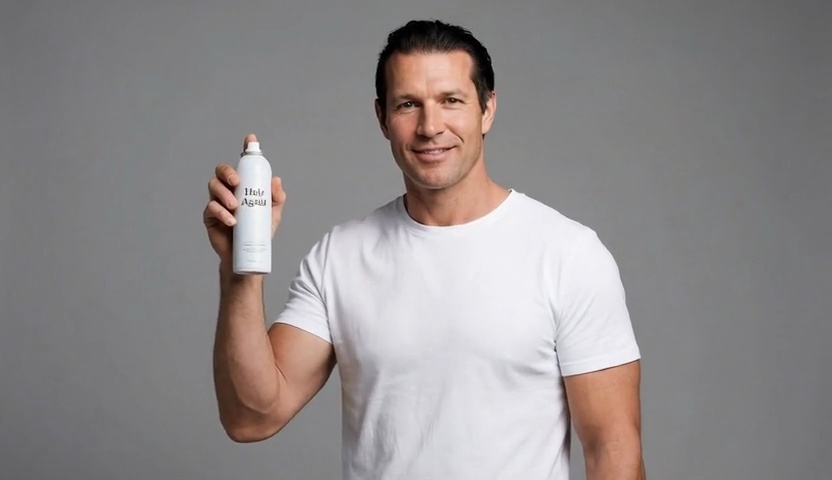} & 
    \includegraphics[width=0.16\linewidth]{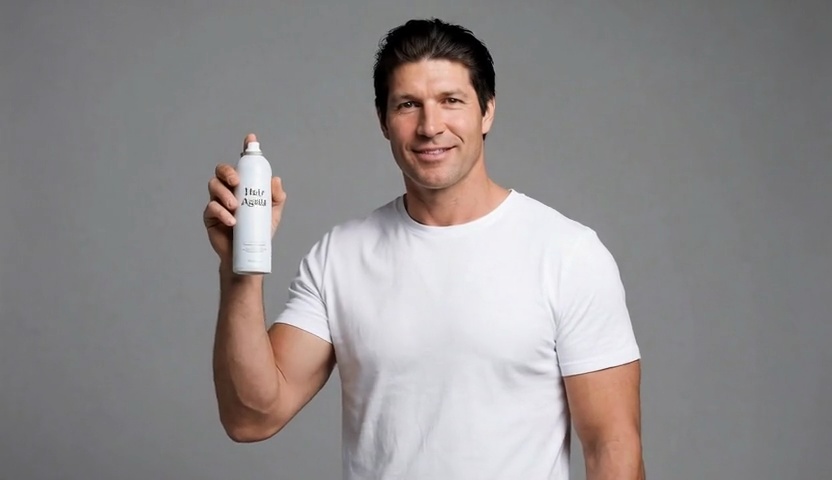} \\[-3.5pt]
    &
    \includegraphics[width=0.16\linewidth]{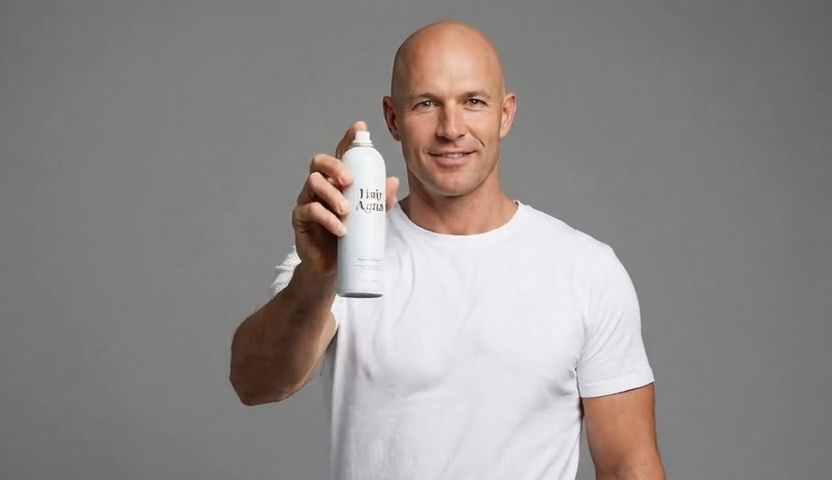} & 
    \includegraphics[width=0.16\linewidth]{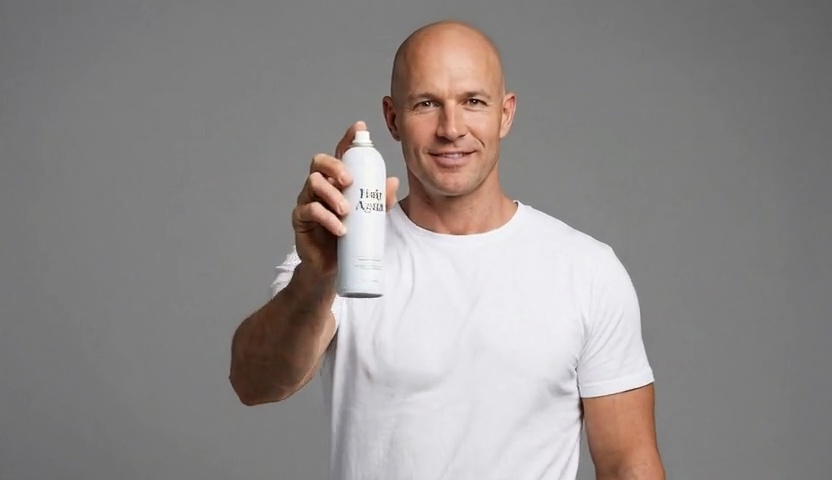} & 
    \includegraphics[width=0.16\linewidth]{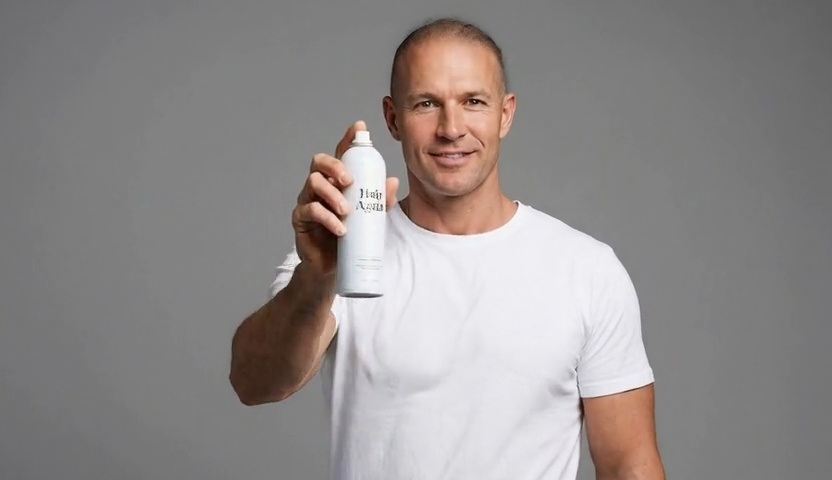} & 
    \includegraphics[width=0.16\linewidth]{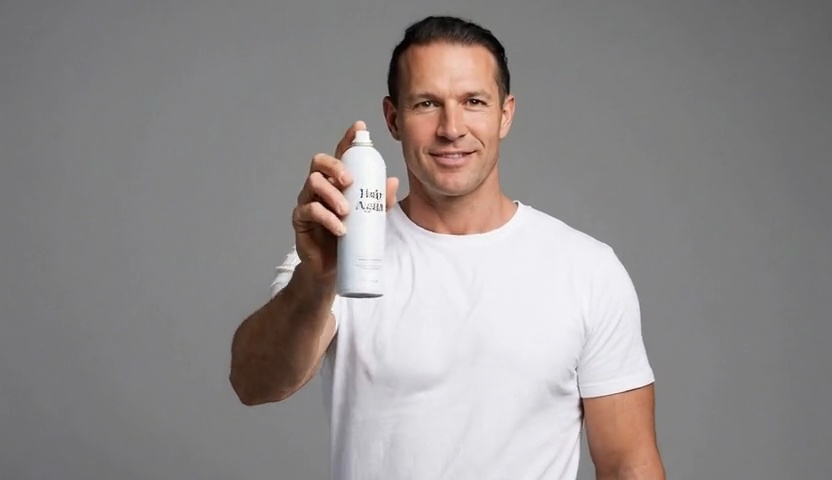} & 
    \includegraphics[width=0.16\linewidth]{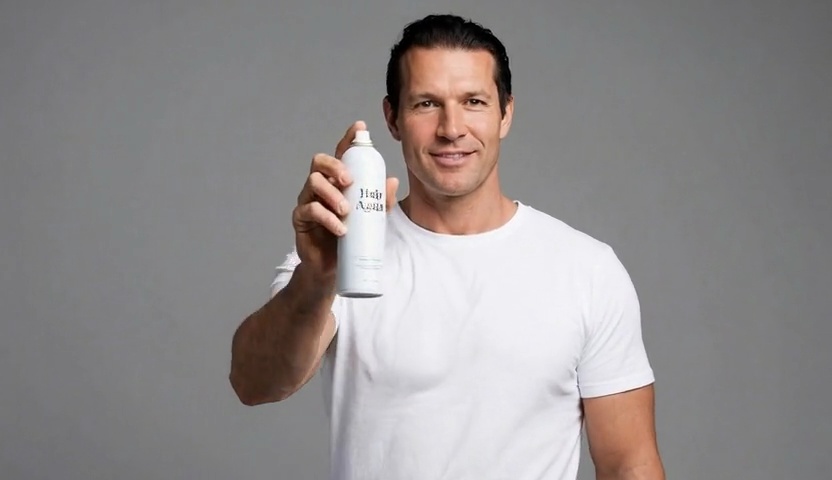} & 
    \includegraphics[width=0.16\linewidth]{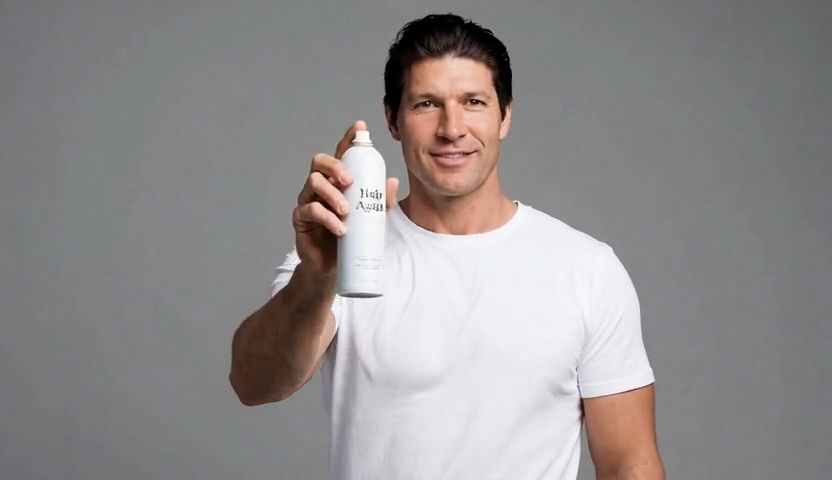} \\
    \multicolumn{7}{c}{\textit{``Replace the white bunny with a chocolate bunny made of brown chocolate
.''}} \\
    {\multirow{3}{*}{{\rotatebox[origin=c]{90}{\hspace{-28pt}Video Frames}}}} &
    \includegraphics[width=0.16\linewidth]{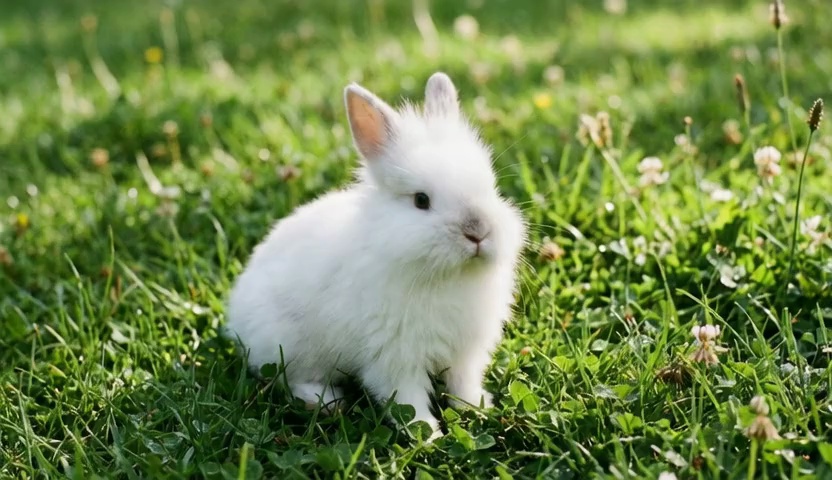} & 
    \includegraphics[width=0.16\linewidth]{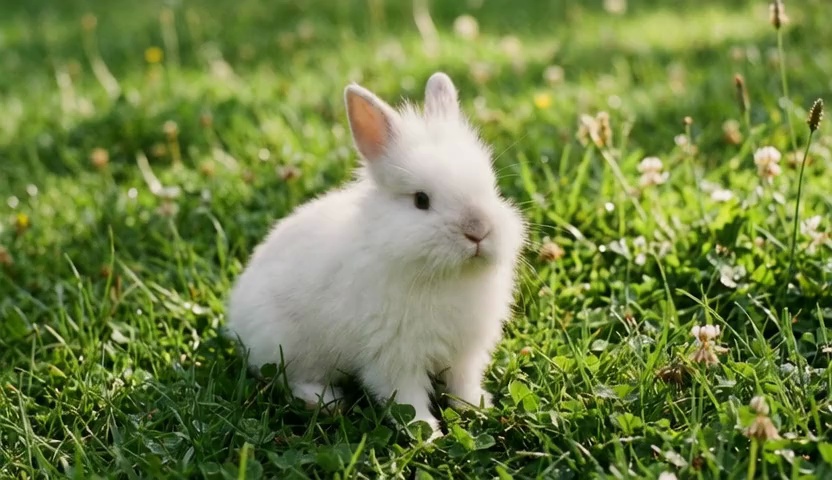} & 
    \includegraphics[width=0.16\linewidth]{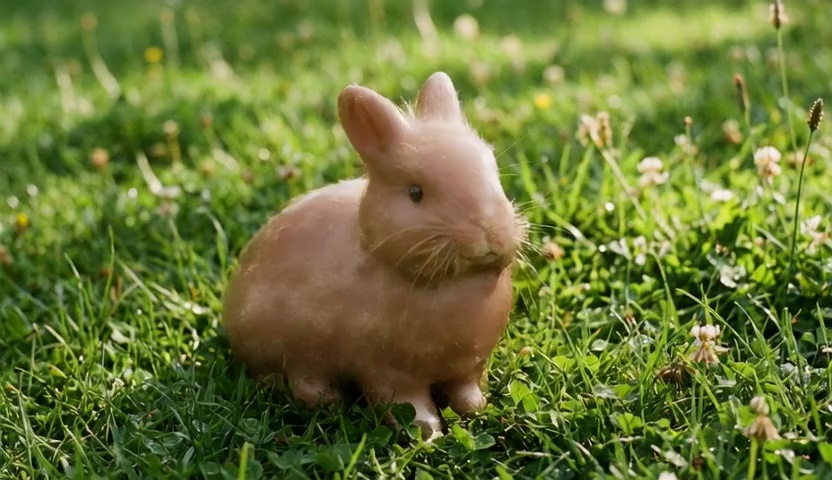} & 
    \includegraphics[width=0.16\linewidth]{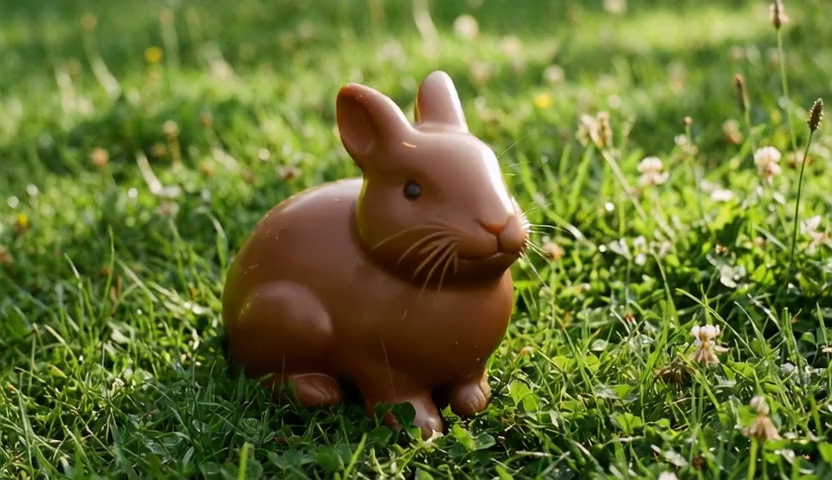} & 
    \includegraphics[width=0.16\linewidth]{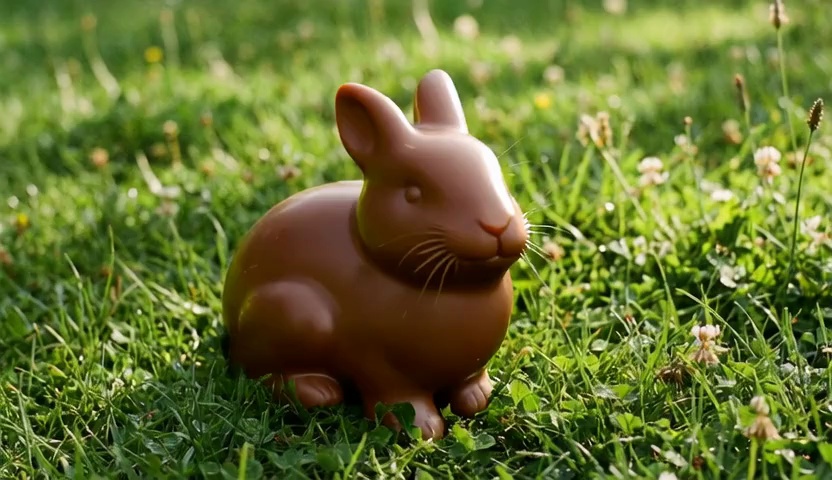} & 
    \includegraphics[width=0.16\linewidth]{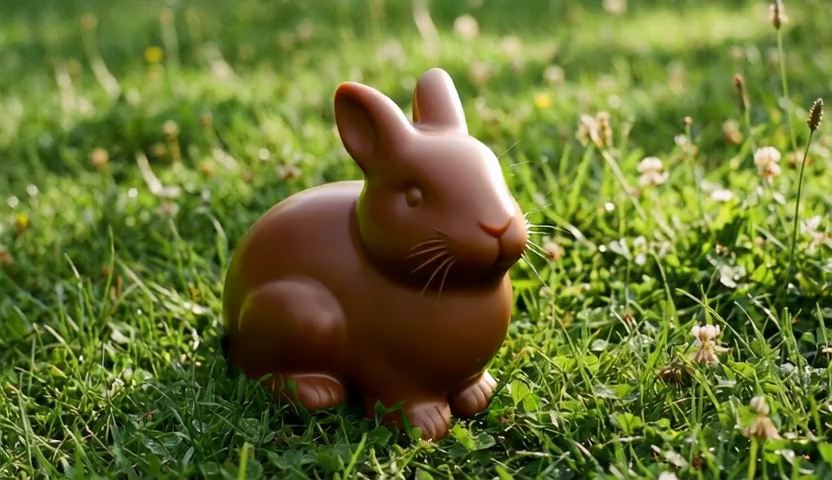} \\[-3.5pt]
    &
    \includegraphics[width=0.16\linewidth]{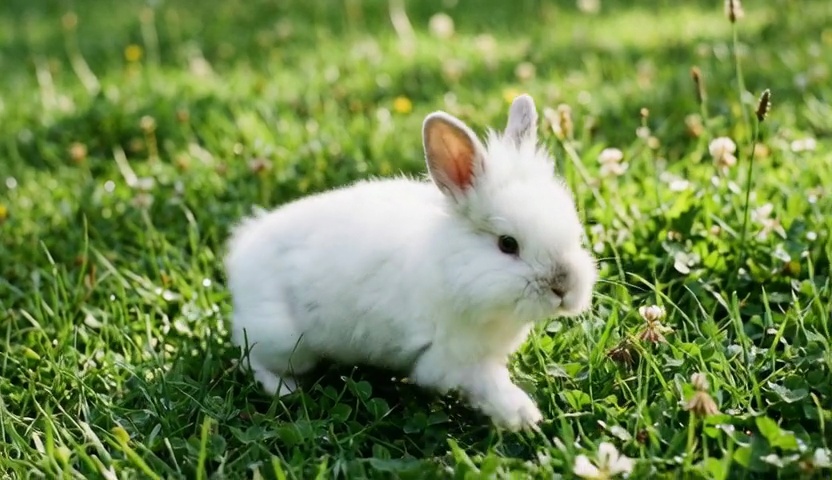} & 
    \includegraphics[width=0.16\linewidth]{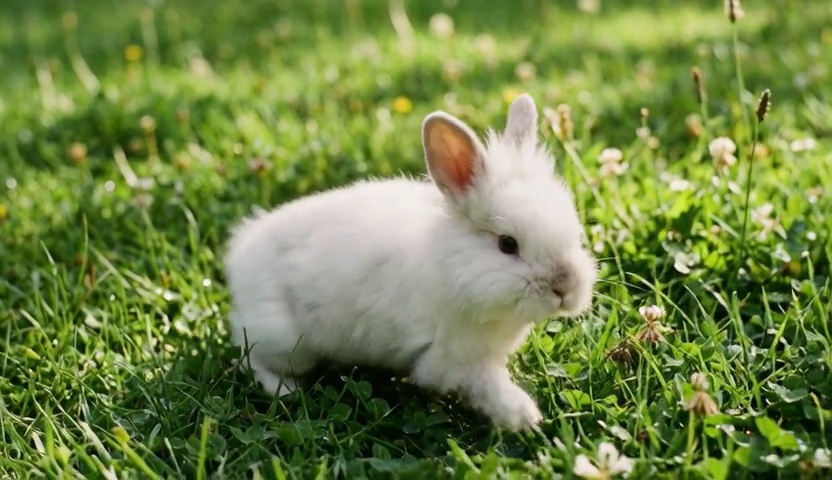} & 
    \includegraphics[width=0.16\linewidth]{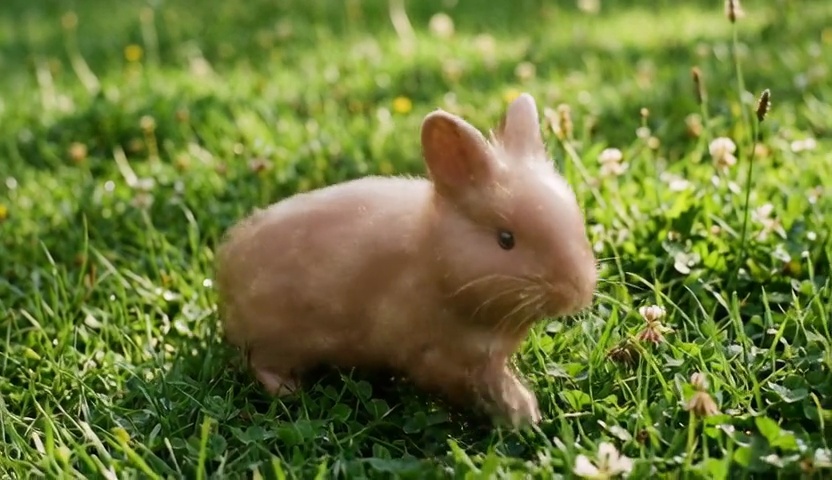} & 
    \includegraphics[width=0.16\linewidth]{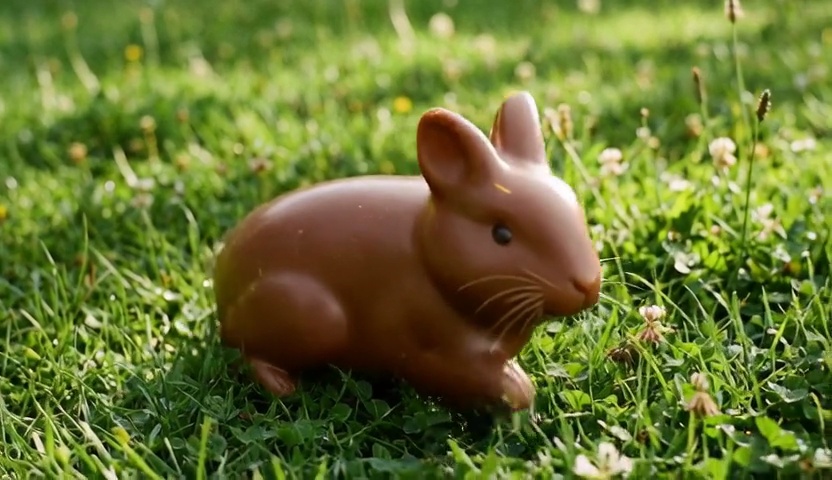} & 
    \includegraphics[width=0.16\linewidth]{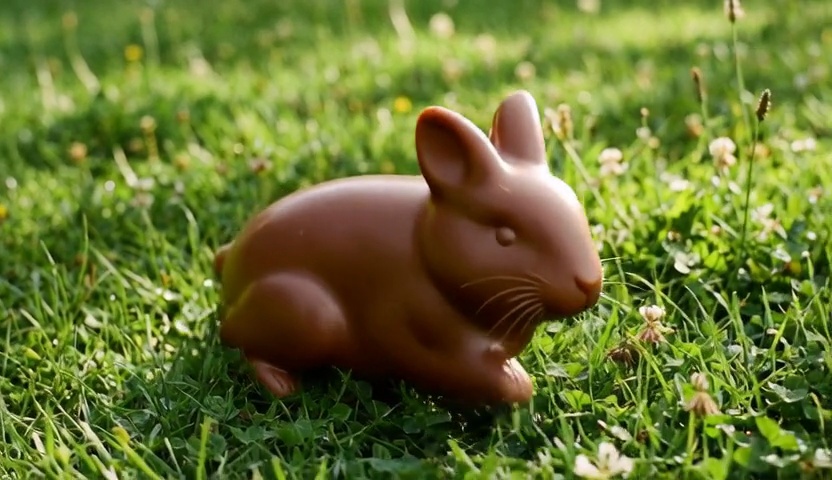} & 
    \includegraphics[width=0.16\linewidth]{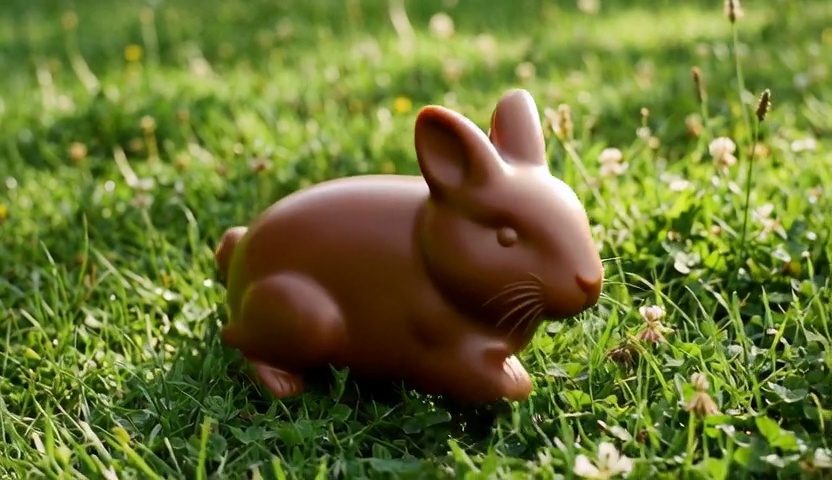} \\[-3.5pt]
    &
    \includegraphics[width=0.16\linewidth]{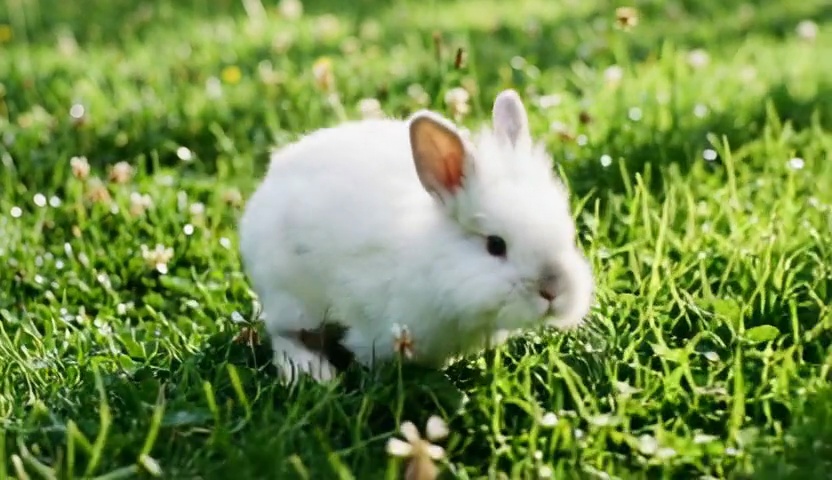} & 
    \includegraphics[width=0.16\linewidth]{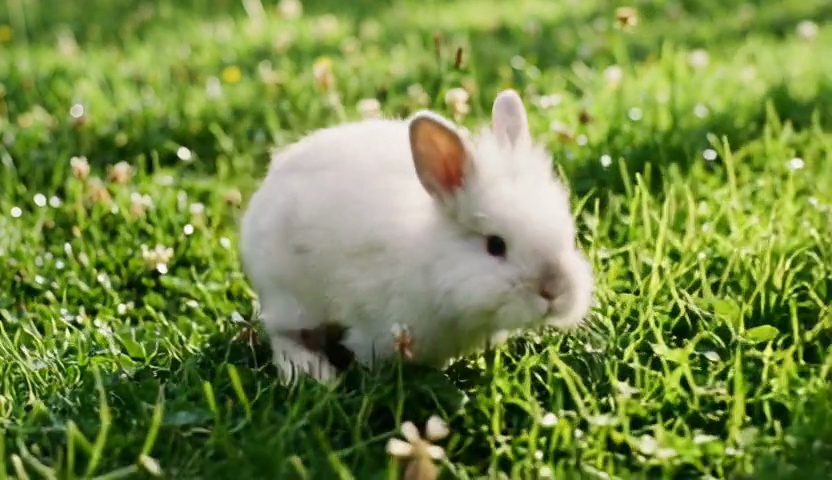} & 
    \includegraphics[width=0.16\linewidth]{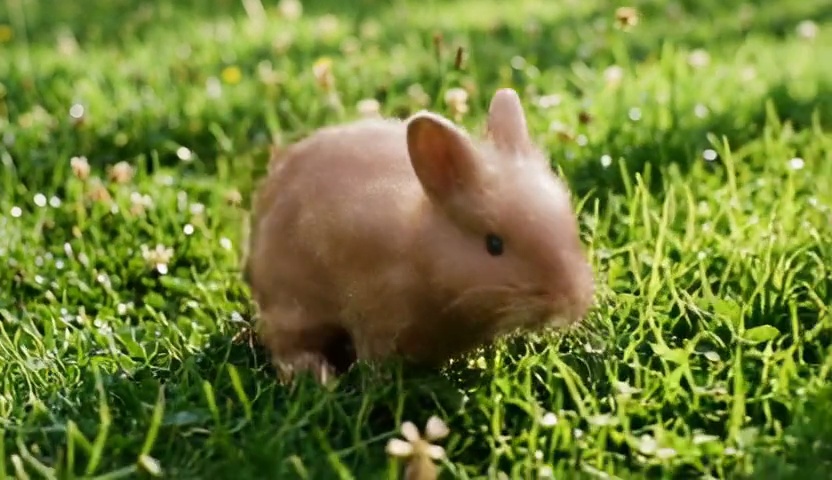} & 
    \includegraphics[width=0.16\linewidth]{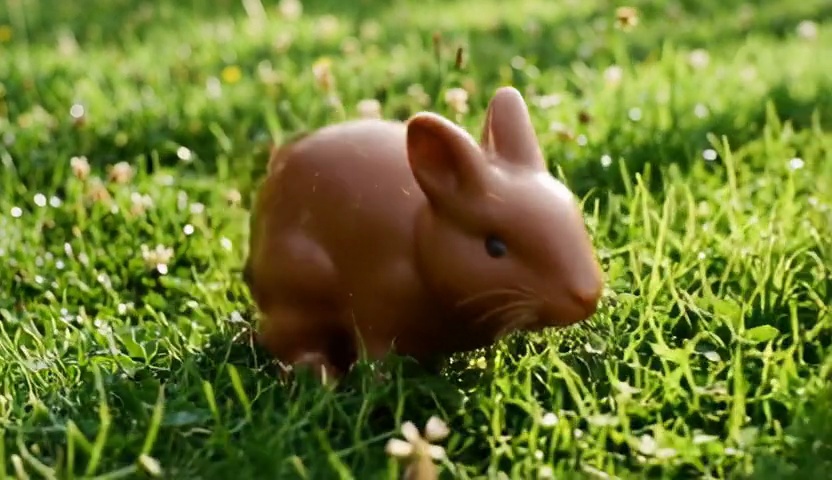} & 
    \includegraphics[width=0.16\linewidth]{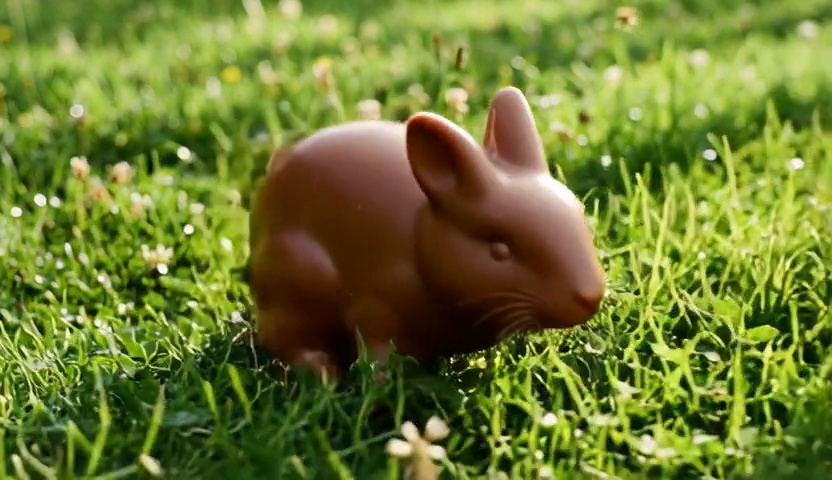} & 
    \includegraphics[width=0.16\linewidth]{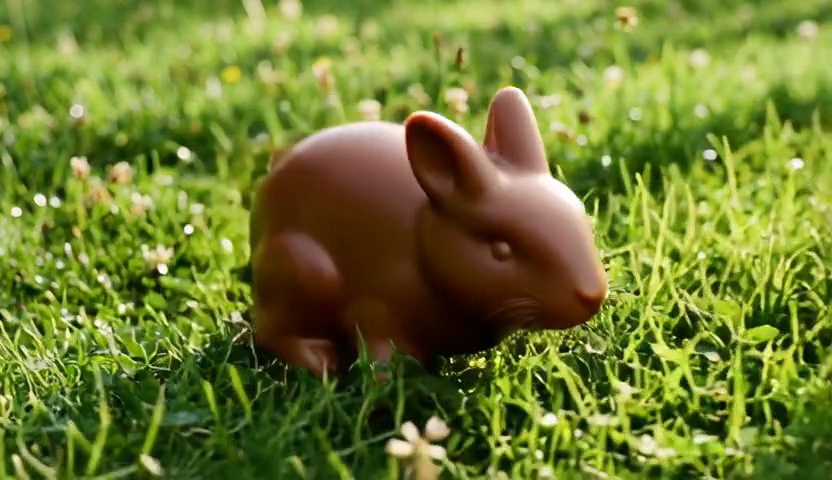} \\
    \multicolumn{7}{c}{\textit{``Replace the dog with a lion.''}} \\
    {\multirow{3}{*}{{\rotatebox[origin=c]{90}{\hspace{-28pt}Video Frames}}}} &
    \includegraphics[width=0.16\linewidth]{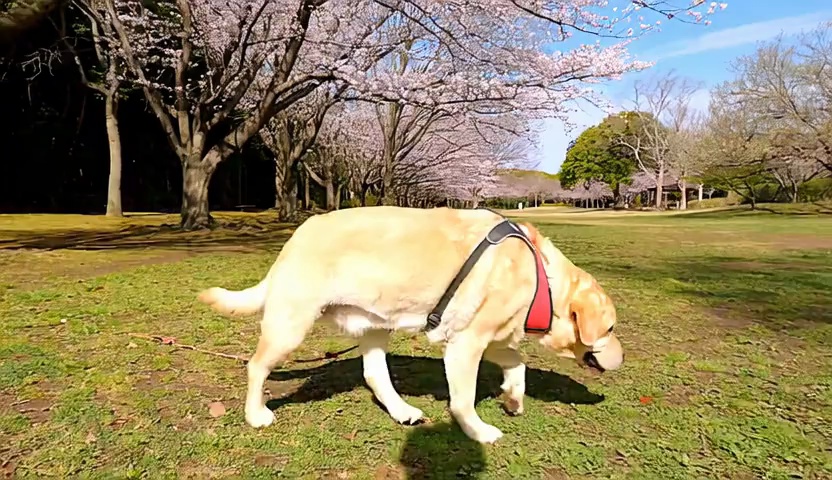} & 
    \includegraphics[width=0.16\linewidth]{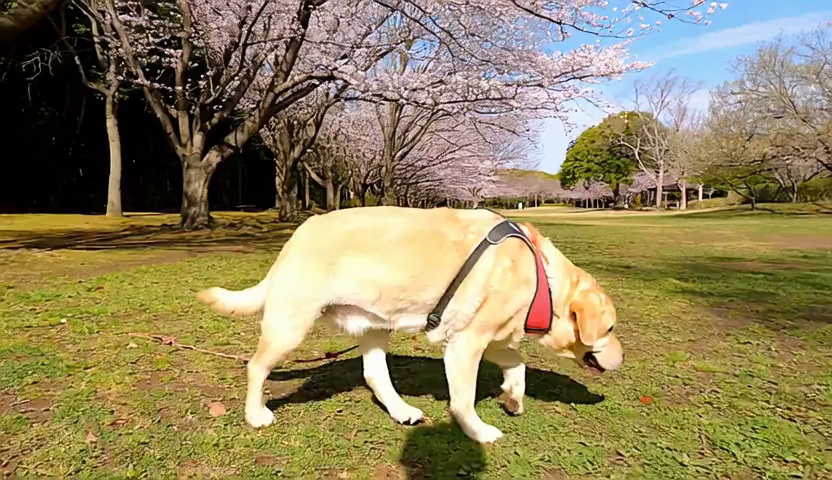} & 
    \includegraphics[width=0.16\linewidth]{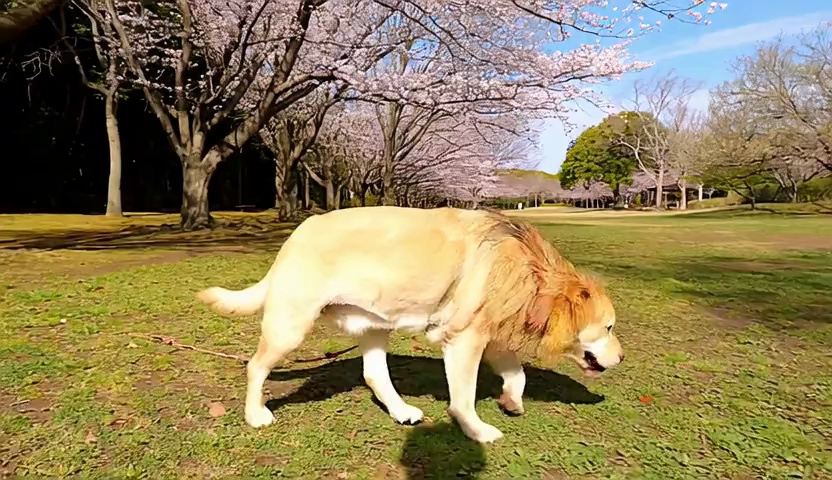} & 
    \includegraphics[width=0.16\linewidth]{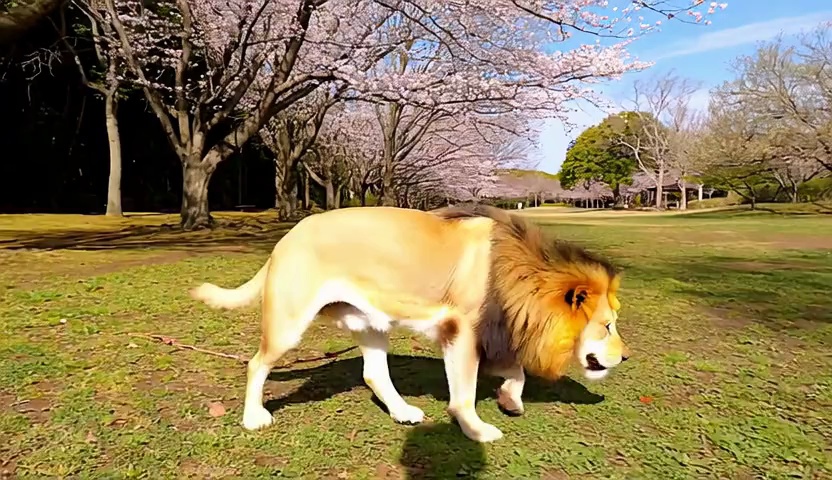} & 
    \includegraphics[width=0.16\linewidth]{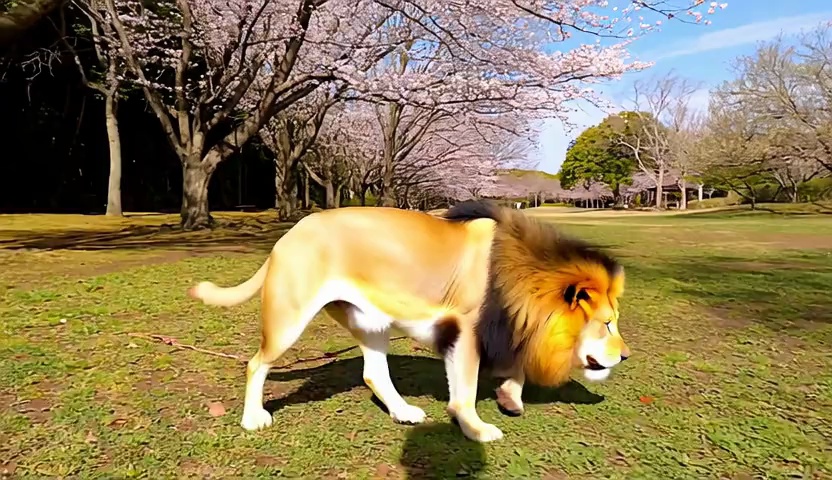} & 
    \includegraphics[width=0.16\linewidth]{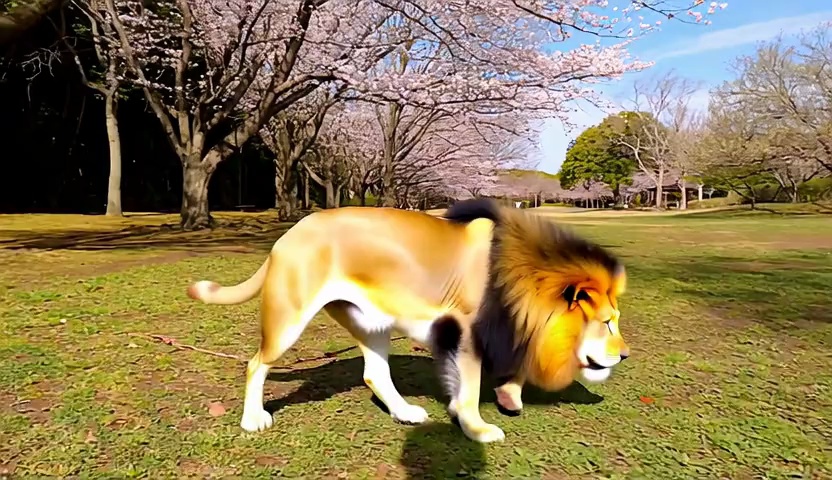} \\[-3.5pt]
    &
    \includegraphics[width=0.16\linewidth]{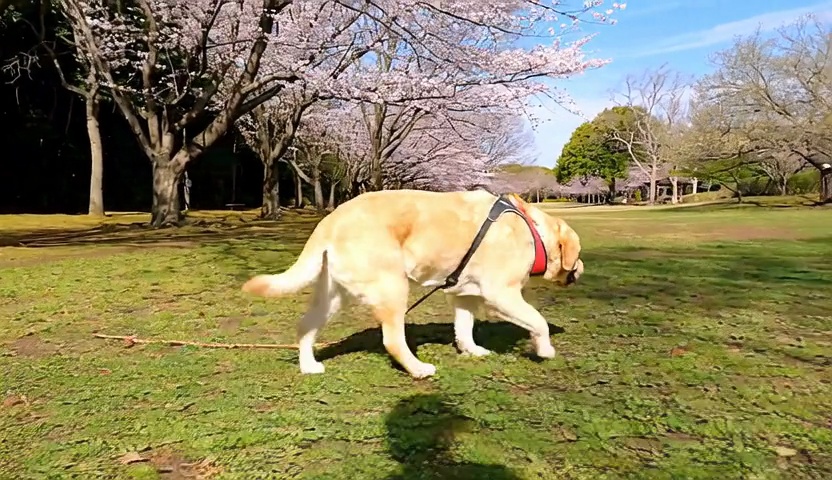} & 
    \includegraphics[width=0.16\linewidth]{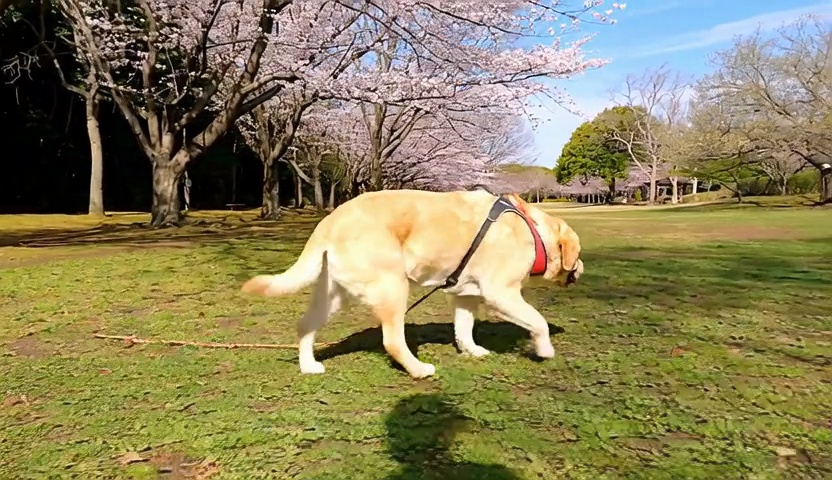} & 
    \includegraphics[width=0.16\linewidth]{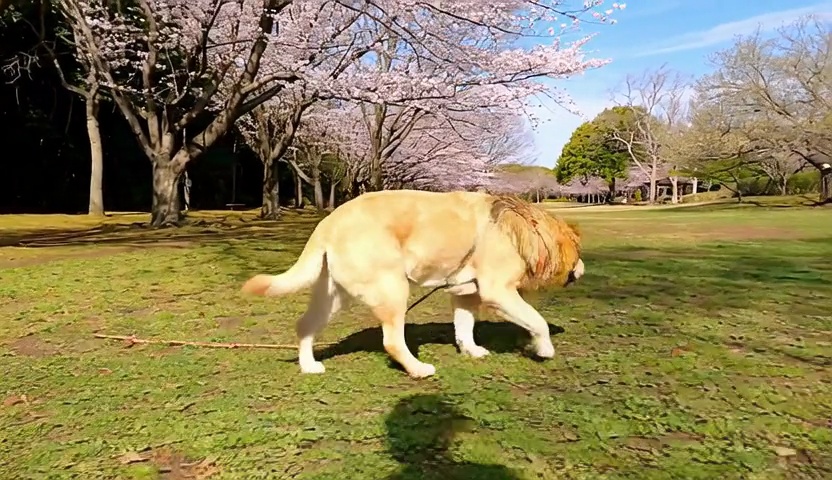} & 
    \includegraphics[width=0.16\linewidth]{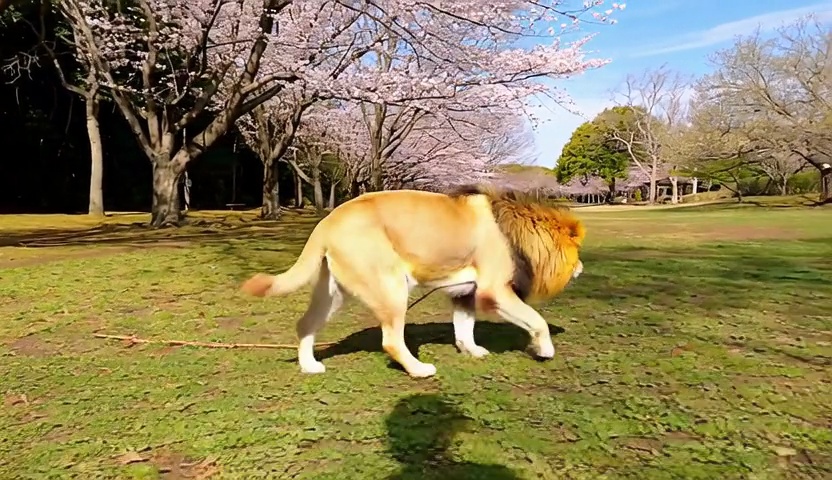} & 
    \includegraphics[width=0.16\linewidth]{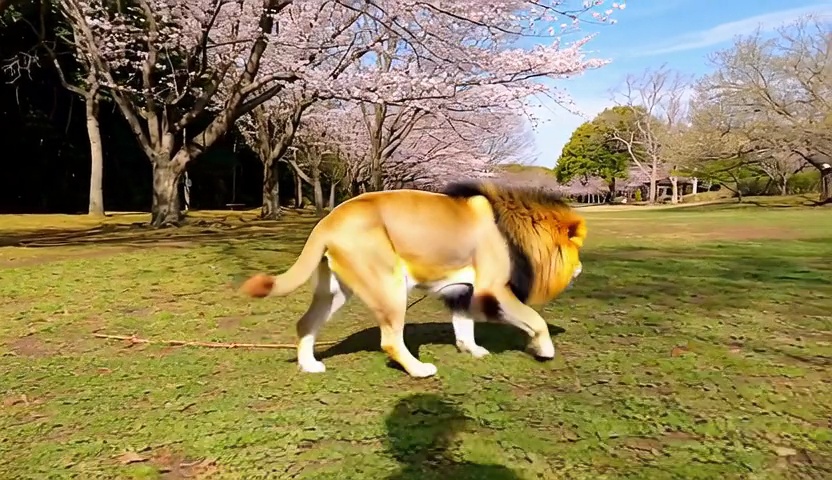} & 
    \includegraphics[width=0.16\linewidth]{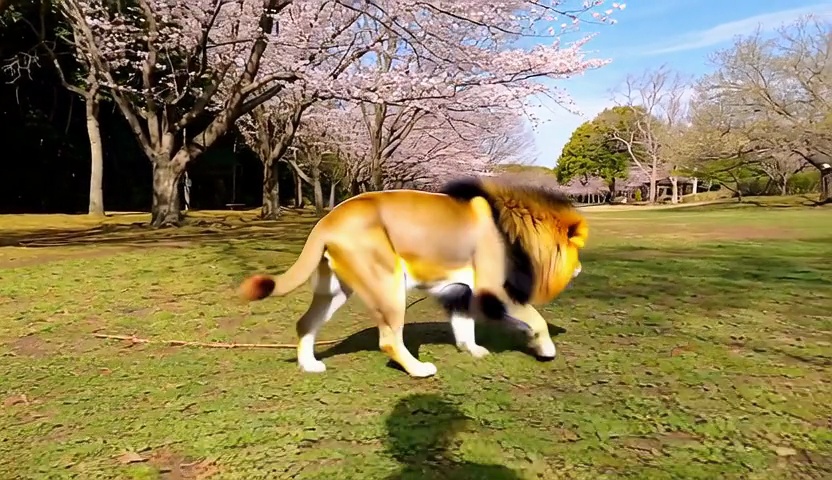} \\[-3.5pt]
    &
    \includegraphics[width=0.16\linewidth]{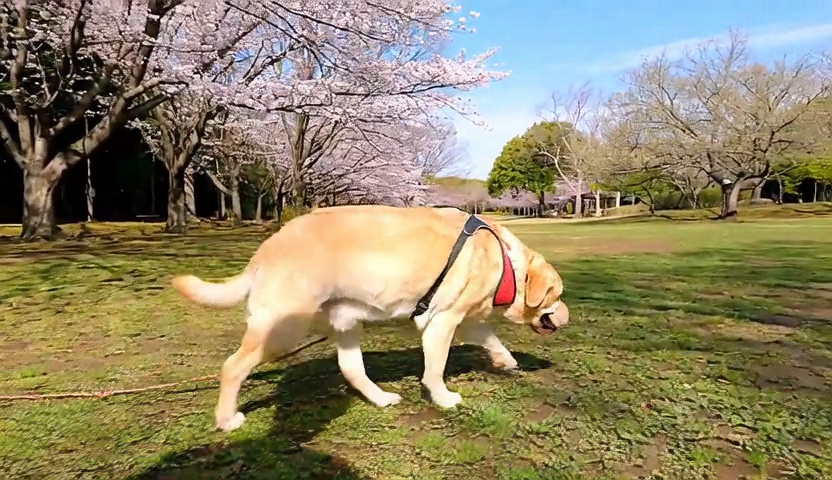} & 
    \includegraphics[width=0.16\linewidth]{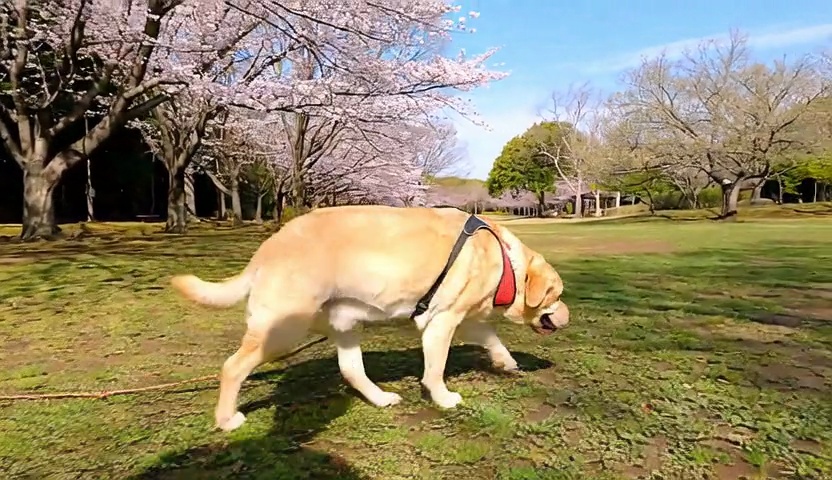} & 
    \includegraphics[width=0.16\linewidth]{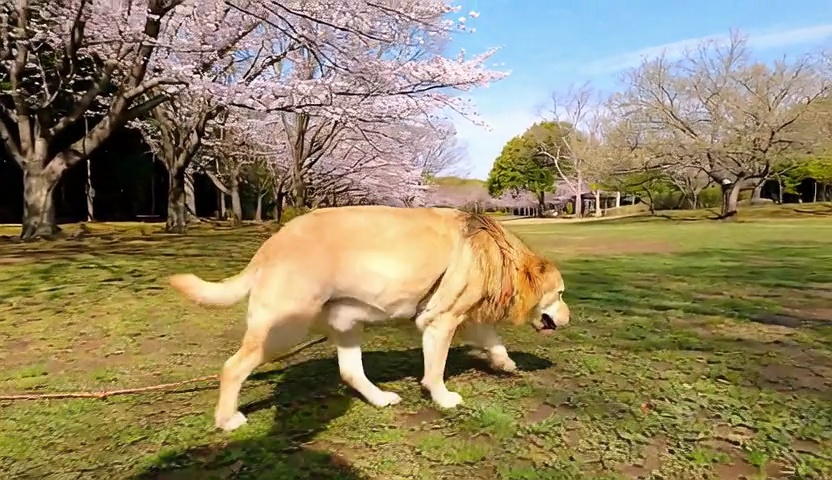} & 
    \includegraphics[width=0.16\linewidth]{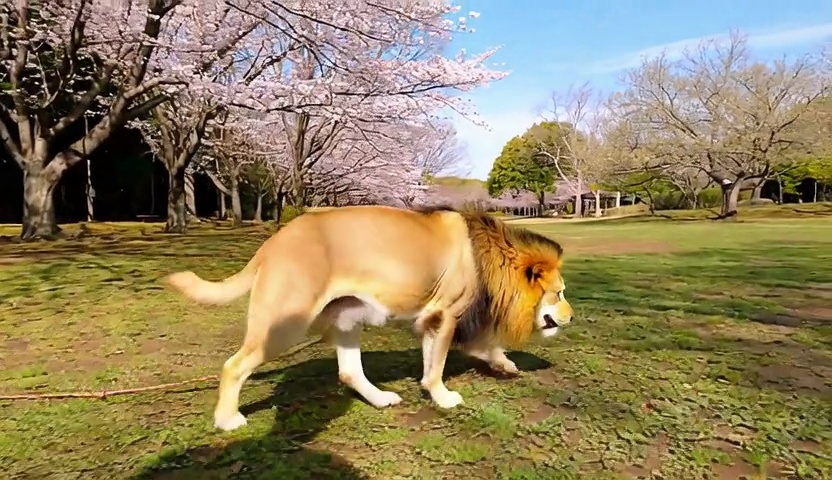} & 
    \includegraphics[width=0.16\linewidth]{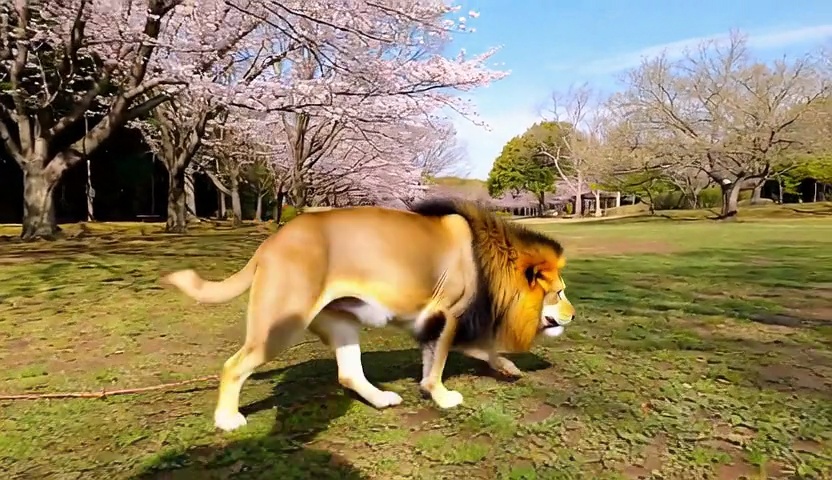} & 
    \includegraphics[width=0.16\linewidth]{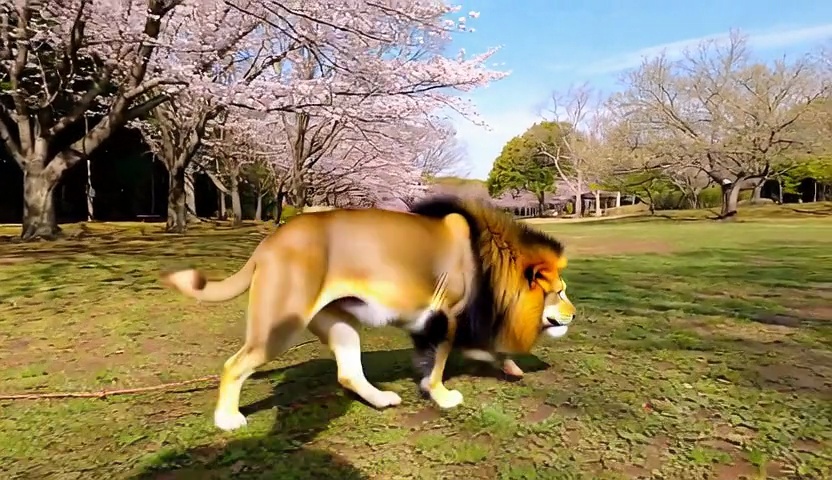} \\
    \multicolumn{7}{c}{\textit{``Change the yellow dress to a pink kimono.''}} \\
    {\multirow{3}{*}{{\rotatebox[origin=c]{90}{\hspace{-28pt}Video Frames}}}} &
    \includegraphics[width=0.16\linewidth]{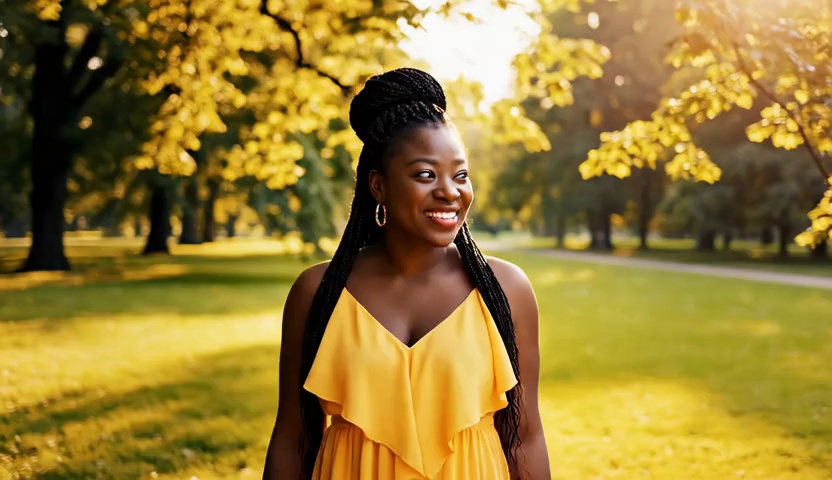} & 
    \includegraphics[width=0.16\linewidth]{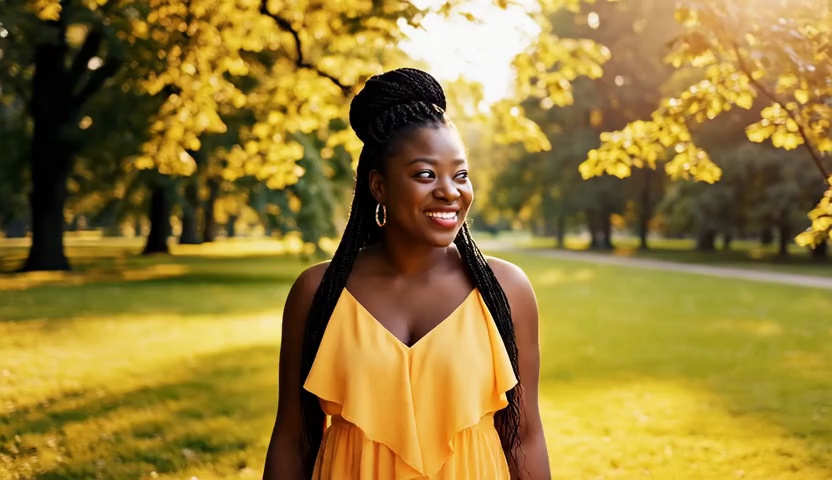} & 
    \includegraphics[width=0.16\linewidth]{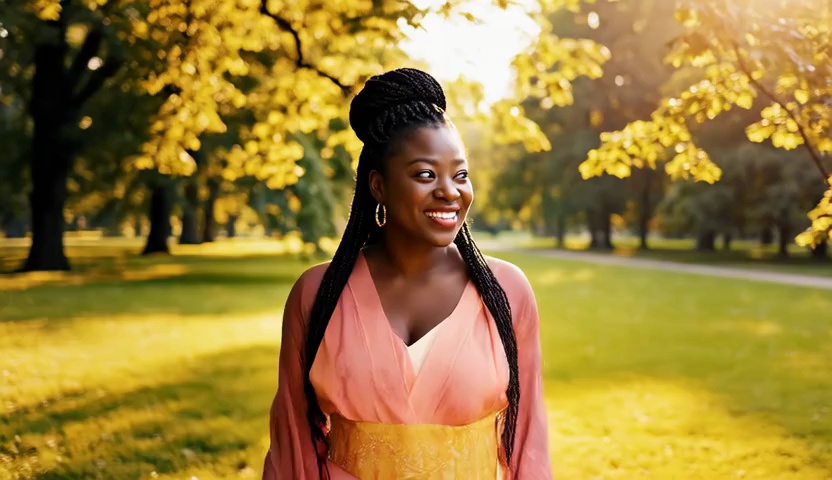} & 
    \includegraphics[width=0.16\linewidth]{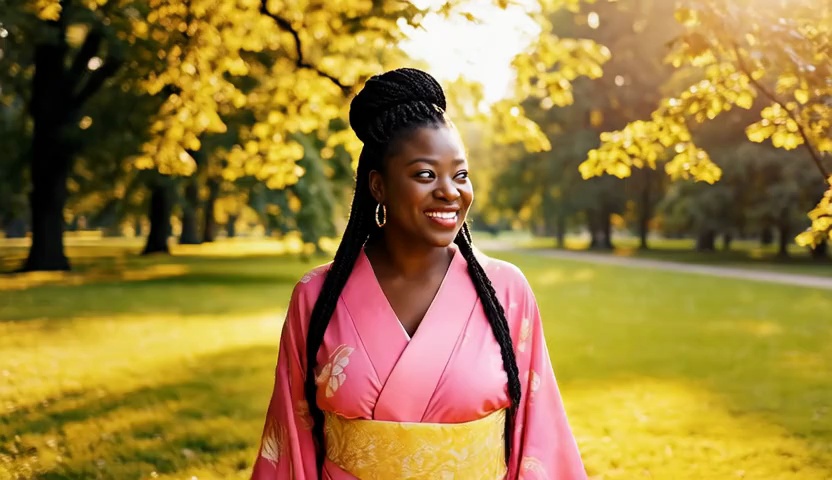} & 
    \includegraphics[width=0.16\linewidth]{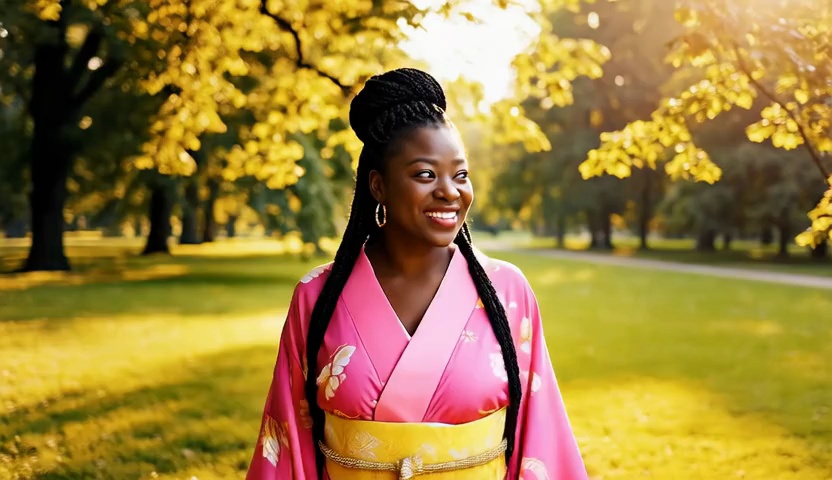} & 
    \includegraphics[width=0.16\linewidth]{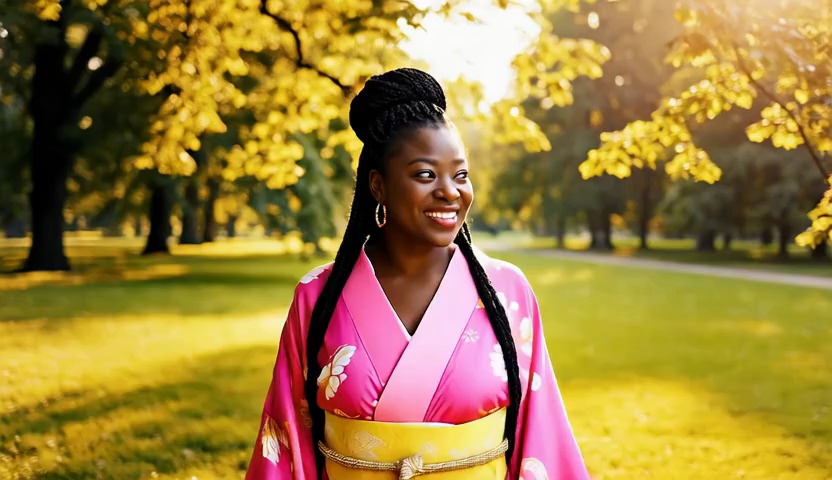} \\[-3.5pt]
    &
    \includegraphics[width=0.16\linewidth]{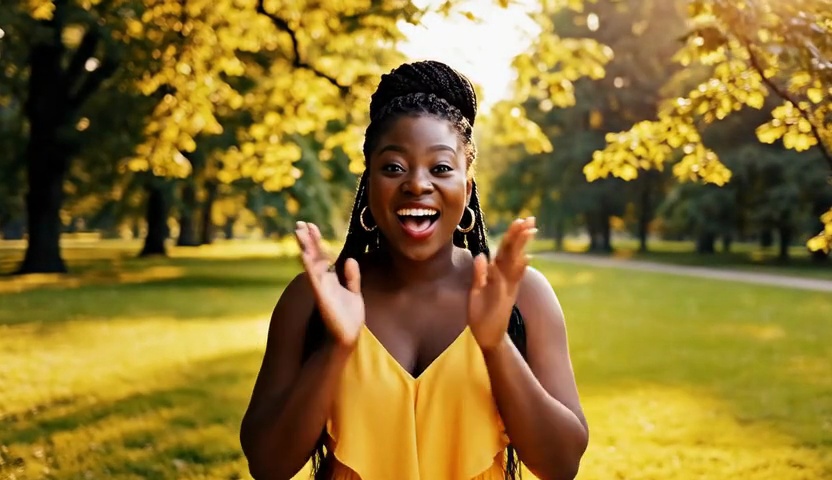} & 
    \includegraphics[width=0.16\linewidth]{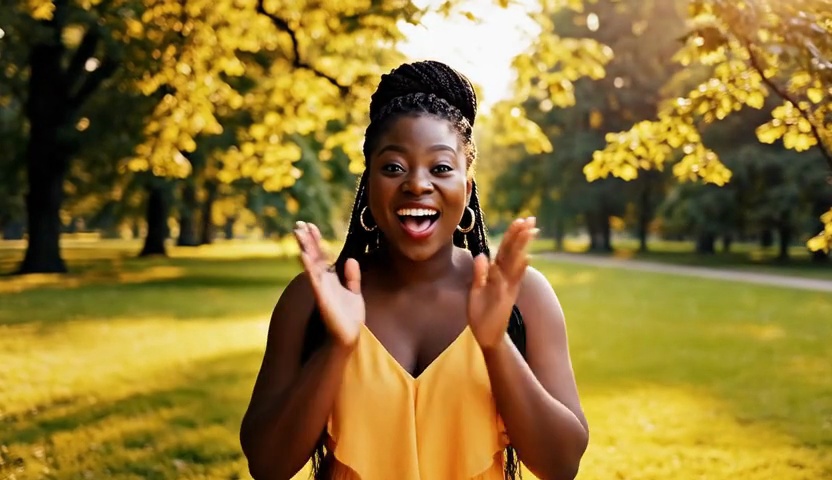} & 
    \includegraphics[width=0.16\linewidth]{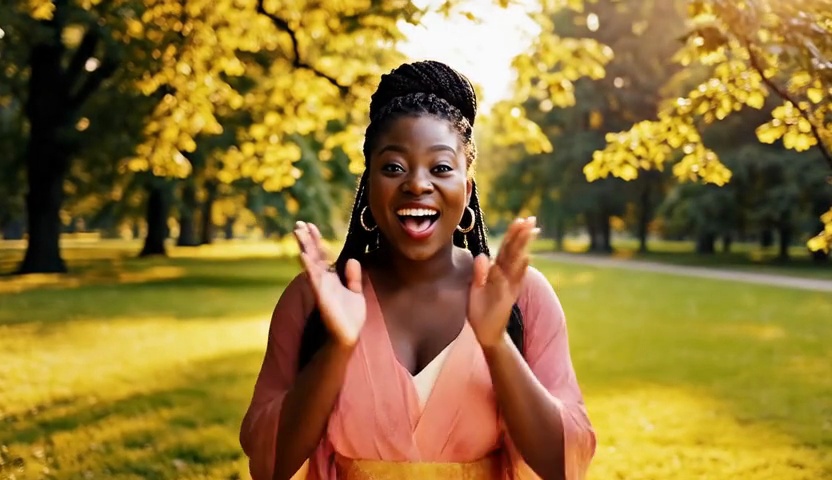} & 
    \includegraphics[width=0.16\linewidth]{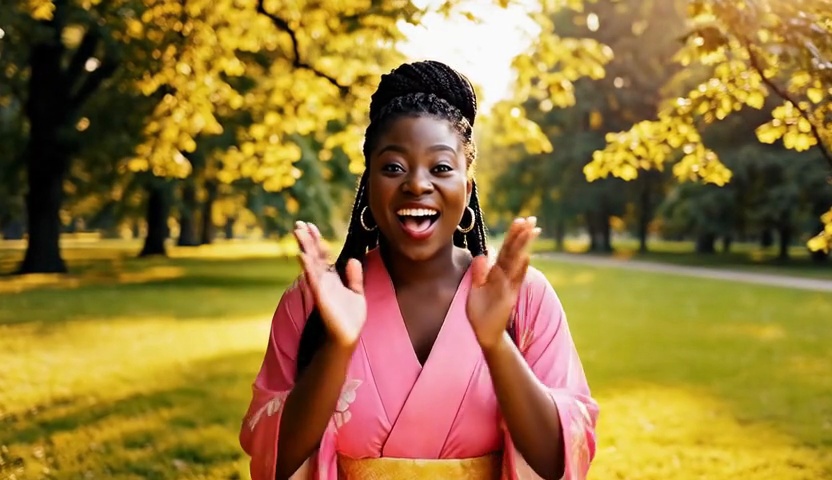} & 
    \includegraphics[width=0.16\linewidth]{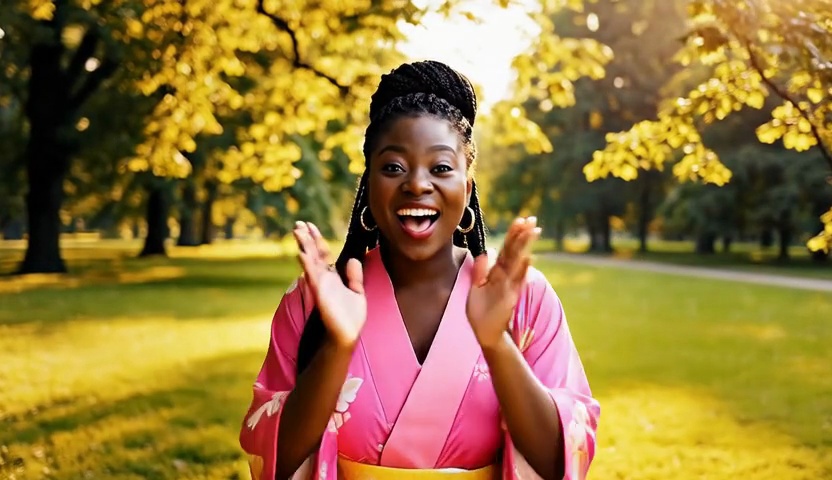} & 
    \includegraphics[width=0.16\linewidth]{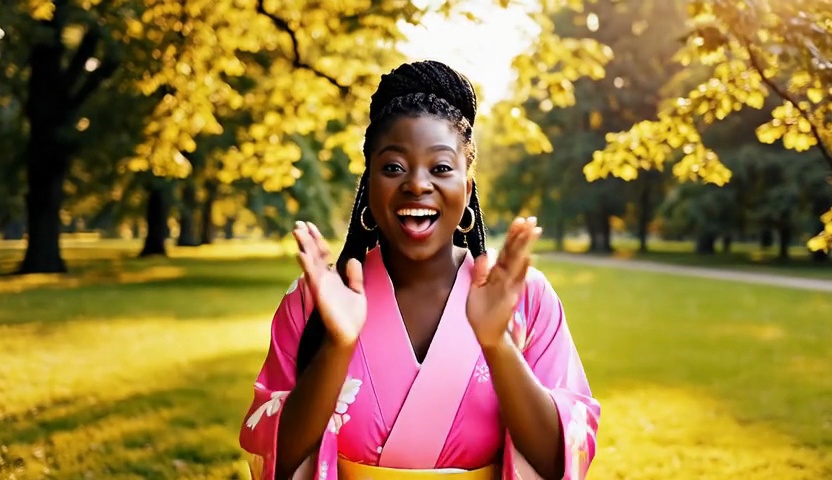} \\[-3.5pt]
    &
    \includegraphics[width=0.16\linewidth]{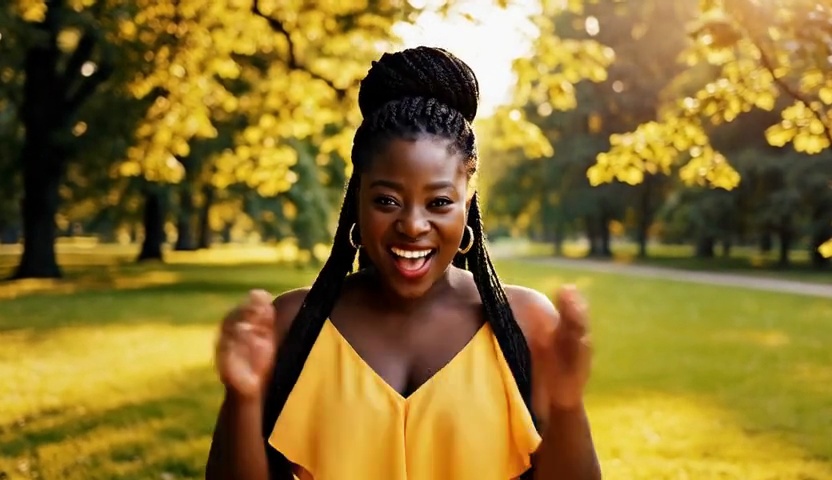} & 
    \includegraphics[width=0.16\linewidth]{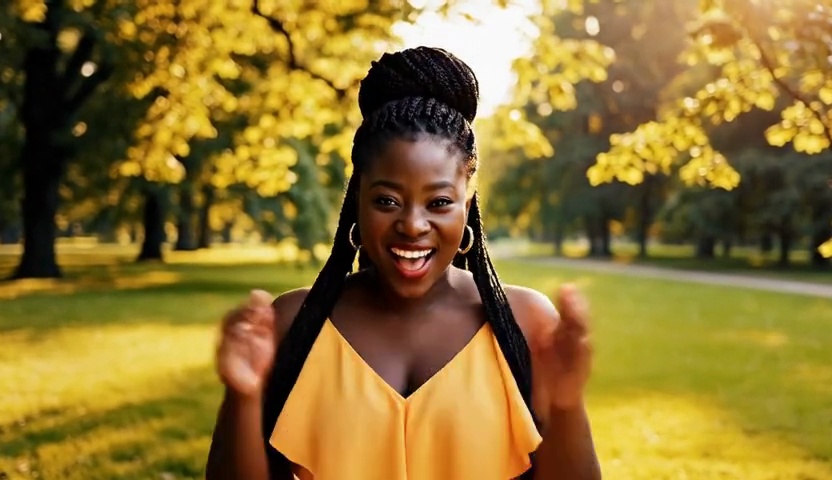} & 
    \includegraphics[width=0.16\linewidth]{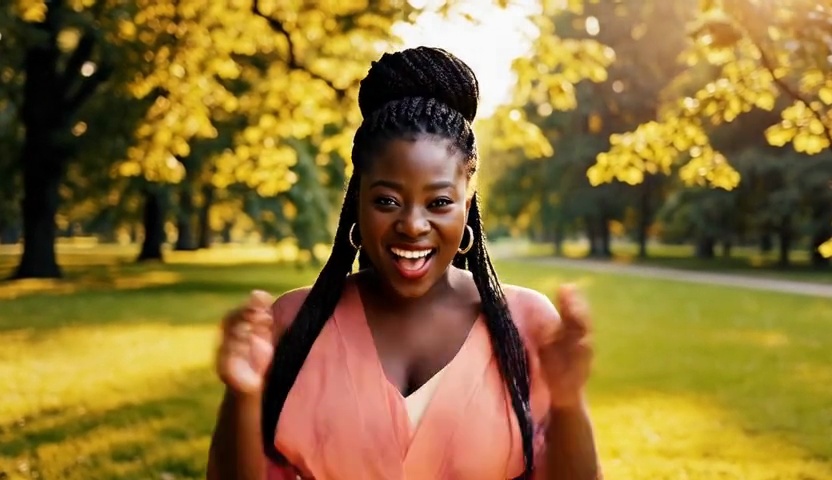} & 
    \includegraphics[width=0.16\linewidth]{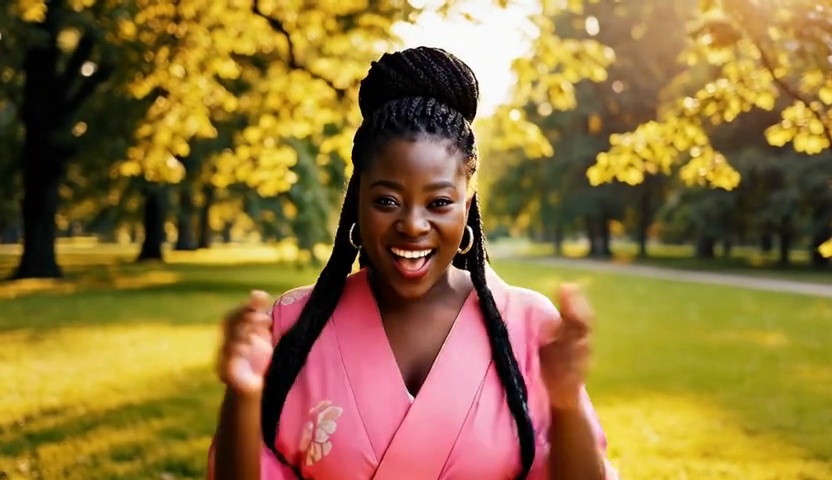} & 
    \includegraphics[width=0.16\linewidth]{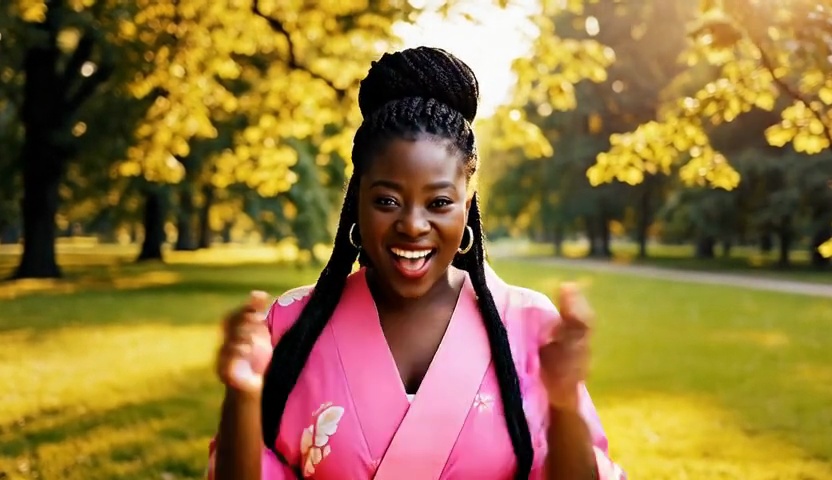} & 
    \includegraphics[width=0.16\linewidth]{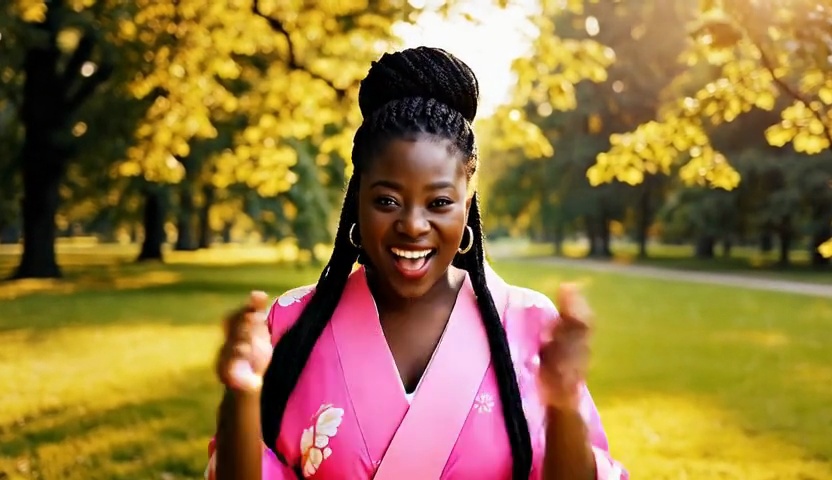} \\

    & 0.000 & 0.194 & 0.387 & 0.613 & 0.806 & 1.000

    \end{tabular}
    \label{fig:more_results_video}
\end{figure*}

%% file: supp/details.tex
\section{Additional Details}

\subsection{Benchmarks}
\label{app:benchmarks}

\paragraph{\textbf{Simplified PIE-Bench}}
Following the protocol of Kontinuous Kontext~\cite{parihar2025kontinuous}, we evaluate our continuous editing method on a curated subset of PIE-Bench~\cite{ju2023direct}. Specifically, we filter out the roughness, transparency, and style categories, resulting in a dataset of 540 tuples consisting of a source image and an editing instruction. 
Since PIE-Bench originally contains many samples requiring multiple simultaneous attribute changes, which are less suitable for continuous editing, we employed an LLM to simplify the instructions. Specifically, the LLM was prompted to revise the instructions to target modifications of only one or two attributes in the source image.

\paragraph{\textbf{SAEdit Benchmark}}
To compare against SAEdit~\cite{kamenetsky2025saedit} and Concept Sliders~\cite{gandikota2023conceptslidersloraadaptors}, we adopt the evaluation benchmark proposed in SAEdit. Our benchmark focuses on three specific attribute categories: curly hair, beard, and angry. For each category, an LLM was used to generate pairs of prompts: one describing the source image and the other incorporating the target attribute corresponding to the category. Source images were then generated from the source prompts using FLUX~\cite{flux} across multiple random seeds. This benchmark comprises 116 samples in total.

\subsection{Metrics}
\label{app:metrics}

We generate a sequence of $N=6$ edits, denoted as $\{I_i\}_{i=1}^{N}$, with editing strengths uniformly distributed between 0 and 1. We assess the performance of each method across the following dimensions:
\begin{enumerate}[leftmargin=*]
\item \textbf{Smoothness and Linearity.}
We evaluate the smoothness of the edit sequence using the $\delta_\text{smooth}$ metric~\cite{parihar2025kontinuous}, which measures second-order smoothness.
To further quantify the uniformity of the transition, we introduce a \textit{Linearity} metric. We compute the Coefficient of Variation (CV) of the stepwise LPIPS~\cite{zhang2018perceptual} distances. Formally, given the set of distances $S = \{\text{LPIPS}(I_{i+1}, I_i)\}$, we calculate $CV = \sigma(S)/\mu(S)$. Lower values indicate a linear, evenly paced progression, while higher values indicate “jumpy” or irregular transitions.
\item \textbf{Text Alignment Consistency.} 
We assess whether the semantic change at each step is consistently aligned with the intended text transformation. We introduce the \textit{Normalized CLIP-Dir} metric, which computes the average cosine similarity between the stepwise image direction and the text direction, normalized by the global image-text alignment. This is computed as:
\begin{equation*}
    \frac{1}{N} \sum_{i=1}^{N-1} \frac{
        \cos(\Delta \mathbf{v}_{img}^{(i)}, \Delta \mathbf{v}_{text})
    }{
        \cos(\Delta \mathbf{v}_{img}^{(global)}, \Delta \mathbf{v}_{text})
    },
\end{equation*}
where $\Delta \mathbf{v}_{img}^{(i)} = \text{CLIP}(I_{i+1}) - \text{CLIP}(I_{i})$ is the local edit direction, $\Delta \mathbf{v}_{img}^{(global)} = \text{CLIP}(I_{N-1}) - \text{CLIP}(I_{0})$ is the total edit direction, and $\Delta \mathbf{v}_{text} = \text{CLIP}(c_\text{edit}) - \text{CLIP}(c_\text{src})$ is the text direction.
\item \textbf{Perceptual Trajectory Consistency.} 
To ensure the editing process follows a direct perceptual path rather than wandering through arbitrary semantic states, we measure the cosine similarity between the direction of each editing step and the global edit direction in the DreamSim~\cite{fu2023dreamsim} embedding space. Formally:
\begin{equation*}
    \frac{1}{N-1} \sum_{i=0}^{N-2} \cos \left( 
        \text{DS}(I_{i+1}) - \text{DS}(I_{i}), \quad \text{DS}(I_{N-1}) - \text{DS}(I_{0}) 
    \right),
\end{equation*}
where $\text{DS}(\cdot)$ denotes the DreamSim~\cite{fu2023dreamsim} embedding.
\end{enumerate}

%% file: supp/additional_results.tex
\section{Additional Experiments}

\subsection{Additional Qualitative Results}

We present more qualitative comparison results in \Cref{fig:comparison_120,fig:comparison_331,fig:comparison_374,fig:comparison_saedit_fjords_leather_jacket_seed990,fig:comparison_saedit_renaissance_painter_courtyard_seed990}.

\subsection{Qwen As Backbone}
\label{app:results_with_qwen}
The main backbone model used in our work for both image and video editing is Lucy-Edit~\cite{decart2025lucyedit}. However, our method is not specific to the Lucy-Edit architecture and can be readily adapted to other backbone models. We demonstrate this generality by implementing our method using Qwen-Image-Edit~\cite{wu2025qwen}. Specifically, we incorporate the \REC{} instruction by training a LoRA module on top of Qwen-Image to learn the identity instruction. All other method details of AdaOr remain unchanged. We present qualitative results in \Cref{fig:qwen-lora-qualitative}.

\subsection{Analytical \REC{} Instruction}

In Section 3.3 of the main paper, we demonstrated that the \REC{} prediction can be analytically approximated by:
\begin{equation*}
\label{eq:rec_analytical_pred}
\epsilon(\mathbf{z_t}; c_I, \text{\REC}, t) \approx \frac{\mathbf{z_t} - c_I}{\sigma_t}.
\end{equation*}
Here, we examine whether this analytical approximation can replace the learned \REC{} prediction employed in AdaOr. Recall that our AdaOr guidance is defined as:
\begin{equation*}
    \epsilon^{w, \alpha}(\mathbf{z_t};c_I, c_T, t) =  \mathcal{O}(\alpha) + \alpha \cdot w (\epsilon(\mathbf{z_t};c_I, c_T, t) - \epsilon(\mathbf{z_t};c_I,\varnothing, t)),
\end{equation*}
where the adaptive origin $\mathcal{O}(\alpha)$ is:
\begin{equation*}
    \mathcal{O}(\alpha) = s(\alpha) \epsilon(\mathbf{z_t};c_I,c_T=\varnothing, t) + (1-s(\alpha)) \epsilon(\mathbf{z_t};c_I,c_T=\text{\REC}, t).
\end{equation*}
We evaluate the effect of replacing the learned term $\epsilon(\mathbf{z_t};c_I,c_T=\text{\REC}, t)$ with the analytical form $(\mathbf{z_t} - c_I) / \sigma_t$ using Lucy-Edit as the backbone. The results are shown in \Cref{fig:analytical_supp}, where the first row shows our results, and the second row shows the analytic version.

As observed in the figure, while this substitution produces smooth and continuous sequences, the intermediate outputs lack realism and appear to drift off the natural image manifold. This discrepancy can be attributed to the fundamental difference between Euclidean and manifold interpolation. The analytical prediction essentially pulls the latent linearly toward the input in Euclidean space. In high-dimensional data spaces, such linear paths often traverse low-density regions that do not correspond to valid images. In contrast, the learned \REC{} prediction leverages the model's learned prior, which captures the underlying data distribution. This allows the editing trajectory to traverse the image manifold, ensuring that intermediate steps remain within the space of valid natural images and resulting in significantly more realistic sequences.

\input{supp/figures/ours_vs_others_piebench/figure_1}

\input{supp/figures/ours_vs_others_saedit/figure_1}

\input{supp/figures/qwen-lora/qwen-lora}

\input{supp/figures/analytical_supp/analytical_supp}

%% file: supp/figures/ours_vs_others_piebench/figure_1.tex
\newcommand{\renderComparisonRow}[3]{
\begin{figure*}
    \centering
    \setlength{\tabcolsep}{0pt}
    \begin{tabular}{cccc cc cccccc}
        \multicolumn{12}{c}{\textit{''#2''}} \\
        \raisebox{26pt}{\rotatebox[origin=t]{90}{{Kontinuous}}} & { } &
        \raisebox{26pt}{\rotatebox[origin=t]{90}{{Kontext}}} & { } &
        \includegraphics[width=0.125\linewidth]{supp/figures/ours_vs_others_piebench/kontionouos-kontext/image_sample_#1/source.jpg} & { } &
        \includegraphics[width=0.125\linewidth]{supp/figures/ours_vs_others_piebench/kontionouos-kontext/image_sample_#1/0.0.jpg} &
        \includegraphics[width=0.125\linewidth]{supp/figures/ours_vs_others_piebench/kontionouos-kontext/image_sample_#1/0.2.jpg} &
        \includegraphics[width=0.125\linewidth]{supp/figures/ours_vs_others_piebench/kontionouos-kontext/image_sample_#1/0.4.jpg} &
        \includegraphics[width=0.125\linewidth]{supp/figures/ours_vs_others_piebench/kontionouos-kontext/image_sample_#1/0.6.jpg} &
        \includegraphics[width=0.125\linewidth]{supp/figures/ours_vs_others_piebench/kontionouos-kontext/image_sample_#1/0.8.jpg} &
        \includegraphics[width=0.125\linewidth]{supp/figures/ours_vs_others_piebench/kontionouos-kontext/image_sample_#1/1.0.jpg} \\
        && \raisebox{26pt}{\rotatebox[origin=t]{90}{{FreeMorph}}} & { } &
        \includegraphics[width=0.125\linewidth]{supp/figures/ours_vs_others_piebench/lucy_freemorph/#1/source.jpg} & { } &
        \includegraphics[width=0.125\linewidth]{supp/figures/ours_vs_others_piebench/lucy_freemorph/#1/0.0.jpg} &
        \includegraphics[width=0.125\linewidth]{supp/figures/ours_vs_others_piebench/lucy_freemorph/#1/0.2.jpg} &
        \includegraphics[width=0.125\linewidth]{supp/figures/ours_vs_others_piebench/lucy_freemorph/#1/0.4.jpg} &
        \includegraphics[width=0.125\linewidth]{supp/figures/ours_vs_others_piebench/lucy_freemorph/#1/0.6.jpg} &
        \includegraphics[width=0.125\linewidth]{supp/figures/ours_vs_others_piebench/lucy_freemorph/#1/0.8.jpg} &
        \includegraphics[width=0.125\linewidth]{supp/figures/ours_vs_others_piebench/lucy_freemorph/#1/1.0.jpg} \\
        && \raisebox{26pt}{\rotatebox[origin=t]{90}{{AdaOr (Ours)}}} & { } &
        \includegraphics[width=0.125\linewidth]{supp/figures/ours_vs_others_piebench/ours/#1/src.jpg} & { } &
        \includegraphics[width=0.125\linewidth]{supp/figures/ours_vs_others_piebench/ours/#1/0.0.jpg} &
        \includegraphics[width=0.125\linewidth]{supp/figures/ours_vs_others_piebench/ours/#1/0.2.jpg} &
        \includegraphics[width=0.125\linewidth]{supp/figures/ours_vs_others_piebench/ours/#1/0.4.jpg} &
        \includegraphics[width=0.125\linewidth]{supp/figures/ours_vs_others_piebench/ours/#1/0.6.jpg} &
        \includegraphics[width=0.125\linewidth]{supp/figures/ours_vs_others_piebench/ours/#1/0.8.jpg} &
        \includegraphics[width=0.125\linewidth]{supp/figures/ours_vs_others_piebench/ours/#1/1.0.jpg} \\
        &&&& Input && 0.0 & 0.2 & 0.4 & 0.6 & 0.8 & 1.0
    \end{tabular}
    \vspace{-8pt}
    \caption{#3}
    \label{fig:comparison_#1}
\end{figure*}
}
\renderComparisonRow{120}{Transform the woman’s brown hair into \textbf{\underline{straight pink hair}}.}{\textbf{Qualitative comparison.}
    We compare our method (bottom) against Kontinuous Kontext (top) and FreeMorph \cite{cao2025freemorph} (middle).}

\renderComparisonRow{331}{\input{supp/figures/ours_vs_others_piebench/kontionouos-kontext/image_sample_331/instruction.txt}}{\textbf{Qualitative comparison.}
    We compare our method (bottom) against Kontinuous Kontext (top) and FreeMorph \cite{cao2025freemorph} (middle).}

\renderComparisonRow{374}{Modify the vintage camera to resemble a \textbf{\underline{wooden toy}} and change travel photography to a plastic theme.}{\textbf{Qualitative comparison.}
    We compare our method (bottom) against Kontinuous Kontext (top) and FreeMorph \cite{cao2025freemorph} (middle).}

%% file: supp/figures/ours_vs_others_piebench/kontionouos-kontext/image_sample_331/instruction.txt
Change the lavender to red and the vase to green.

%% file: supp/figures/ours_vs_others_saedit/figure_1.tex
\newcommand{\renderSAEditComparison}[3]{
\begin{figure*}
    \centering
    \setlength{\tabcolsep}{0pt}
    \begin{tabular}{cc c cccc}
    \multicolumn{7}{c}{\textit{#3}} \\

    \raisebox{45pt}{\rotatebox[origin=t]{90}{\small{Concept Sliders}}} & { } &
    \includegraphics[width=0.195\linewidth]{supp/figures/ours_vs_others_saedit/concept-sliders/#1/#2_factor0.0.jpg} &
    \includegraphics[width=0.195\linewidth]{supp/figures/ours_vs_others_saedit/concept-sliders/#1/#2_factor0.2.jpg} & 
    \includegraphics[width=0.195\linewidth]{supp/figures/ours_vs_others_saedit/concept-sliders/#1/#2_factor0.5.jpg} & 
    \includegraphics[width=0.195\linewidth]{supp/figures/ours_vs_others_saedit/concept-sliders/#1/#2_factor1.0.jpg} & 
    \includegraphics[width=0.195\linewidth]{supp/figures/ours_vs_others_saedit/concept-sliders/#1/#2_factor2.0.jpg}\\
    \raisebox{45pt}{\rotatebox[origin=t]{90}{\small{SAEdit}}} & { } &
    \includegraphics[width=0.195\linewidth]{supp/figures/ours_vs_others_saedit/saedit/#1/#2_factor8.0.jpg} & 
    \includegraphics[width=0.195\linewidth]{supp/figures/ours_vs_others_saedit/saedit/#1/#2_factor10.0.jpg} & 
    \includegraphics[width=0.195\linewidth]{supp/figures/ours_vs_others_saedit/saedit/#1/#2_factor11.0.jpg} & 
    \includegraphics[width=0.195\linewidth]{supp/figures/ours_vs_others_saedit/saedit/#1/#2_factor12.0.jpg} & 
    \includegraphics[width=0.195\linewidth]{supp/figures/ours_vs_others_saedit/saedit/#1/#2_factor13.0.jpg} \\
    \raisebox{45pt}{\rotatebox[origin=t]{90}{\small{AdaOr (Ours)}}} & { } &
    \includegraphics[width=0.195\linewidth]{supp/figures/ours_vs_others_saedit/ours/#1/0.0/inference_output.jpg} & 
    \includegraphics[width=0.195\linewidth]{supp/figures/ours_vs_others_saedit/ours/#1/0.2/inference_output.jpg} & 
    \includegraphics[width=0.195\linewidth]{supp/figures/ours_vs_others_saedit/ours/#1/0.4/inference_output.jpg} & 
    \includegraphics[width=0.195\linewidth]{supp/figures/ours_vs_others_saedit/ours/#1/0.8/inference_output.jpg} & 
    \includegraphics[width=0.195\linewidth]{supp/figures/ours_vs_others_saedit/ours/#1/1.0/inference_output.jpg}  \\
    && \multicolumn{5}{l}{\small{Edit Intensity} $\xrightarrow{\hspace{440pt}}$} \\

    \end{tabular}
    \caption{\textbf{Qualitative comparison.} Comparison of our method (bottom) against Concept Sliders (top) and SAEdit (middle).}
    \label{fig:comparison_saedit_#2}
\end{figure*}
}

\renderSAEditComparison{beard_edits_fjords_leather_jacket_seed990}{fjords_leather_jacket_seed990}{``Add beard''}
\renderSAEditComparison{curly_hair_renaissance_painter_courtyard_seed990}{renaissance_painter_courtyard_seed990}{``Add curly hair''}

%% file: supp/figures/qwen-lora/qwen-lora.tex
\newcommand{\renderQwenLoraRow}[1]{
    \multicolumn{5}{c}{\textit{``\input{supp/figures/qwen-lora/#1/instruction.txt}''}} \\
    \includegraphics[width=0.15\linewidth]{supp/figures/qwen-lora/#1/0.0.jpg} &
    \includegraphics[width=0.15\linewidth]{supp/figures/qwen-lora/#1/0.2.jpg} &
    \includegraphics[width=0.15\linewidth]{supp/figures/qwen-lora/#1/0.6.jpg} &
    \includegraphics[width=0.15\linewidth]{supp/figures/qwen-lora/#1/0.8.jpg} &
    \includegraphics[width=0.15\linewidth]{supp/figures/qwen-lora/#1/1.0.jpg} \\
}

\begin{figure*}
    \centering
    \setlength{\tabcolsep}{0pt} %
    
    \begin{tabular}{cccccc}
        
    \multicolumn{5}{c}{\textit{``\input{supp/figures/qwen-lora/19/instruction.txt}''}} \\
    \includegraphics[width=0.15\linewidth]{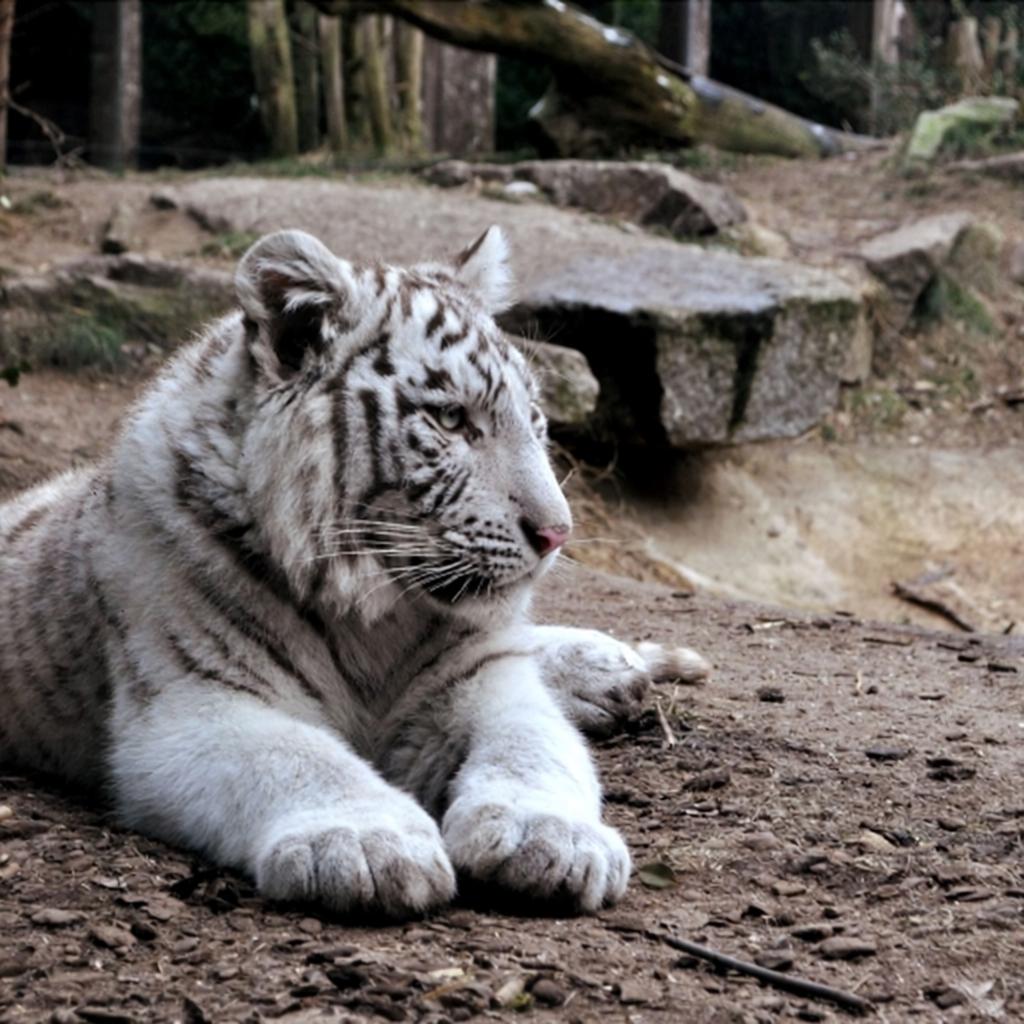} &
    \includegraphics[width=0.15\linewidth]{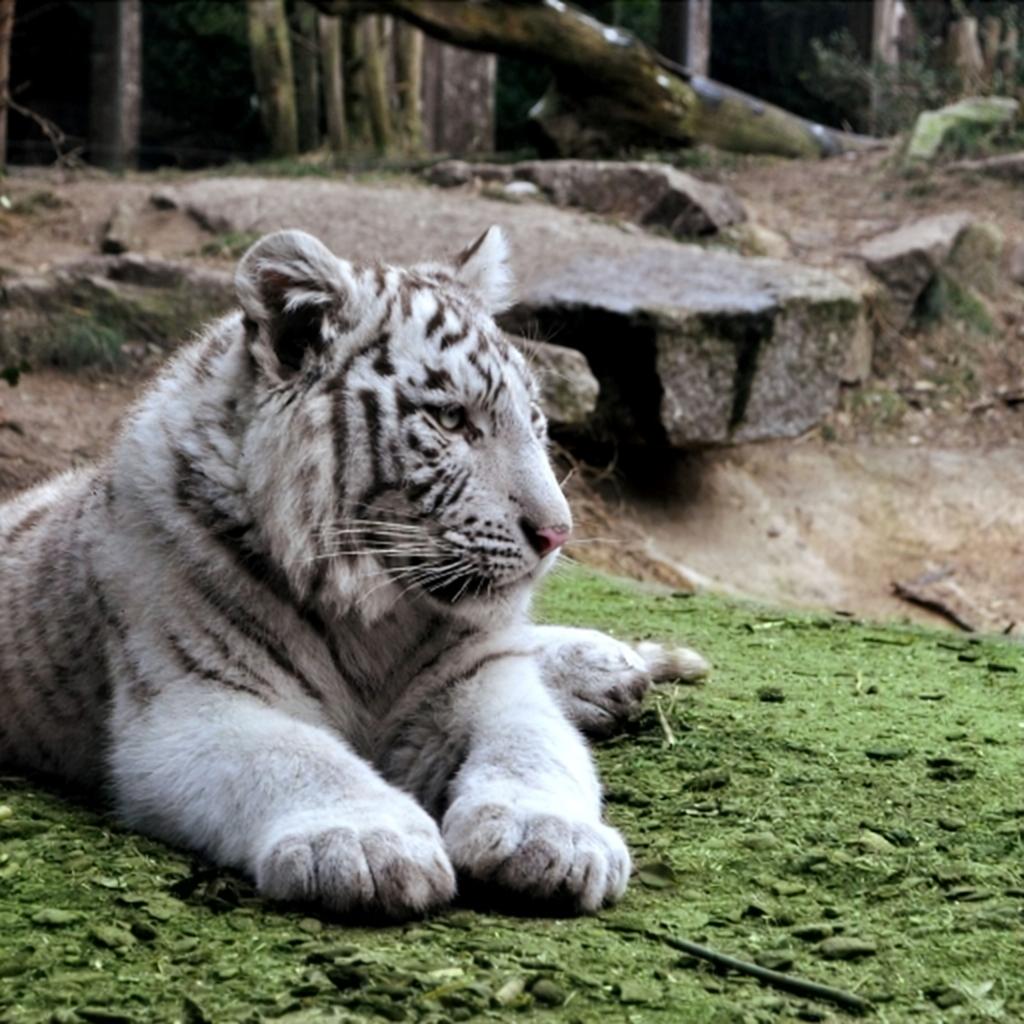} &
    \includegraphics[width=0.15\linewidth]{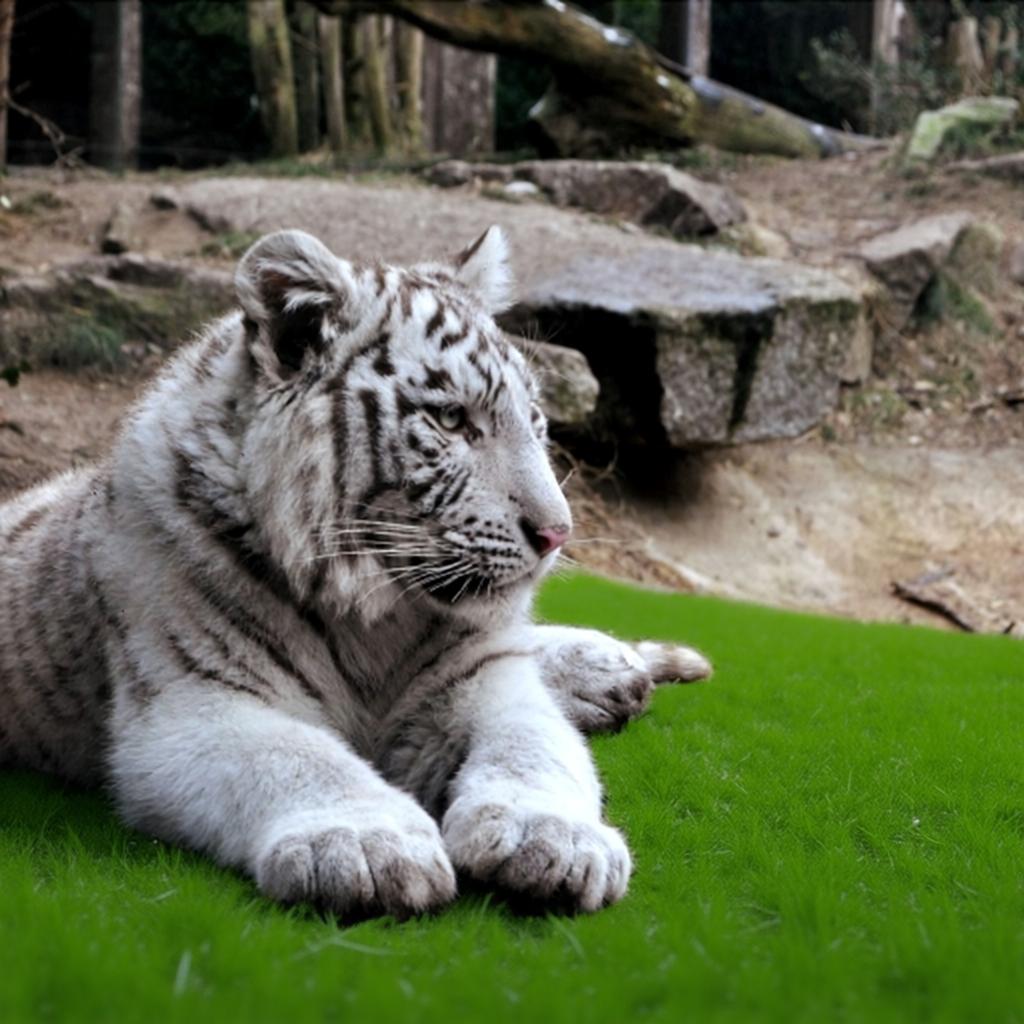} &
    \includegraphics[width=0.15\linewidth]{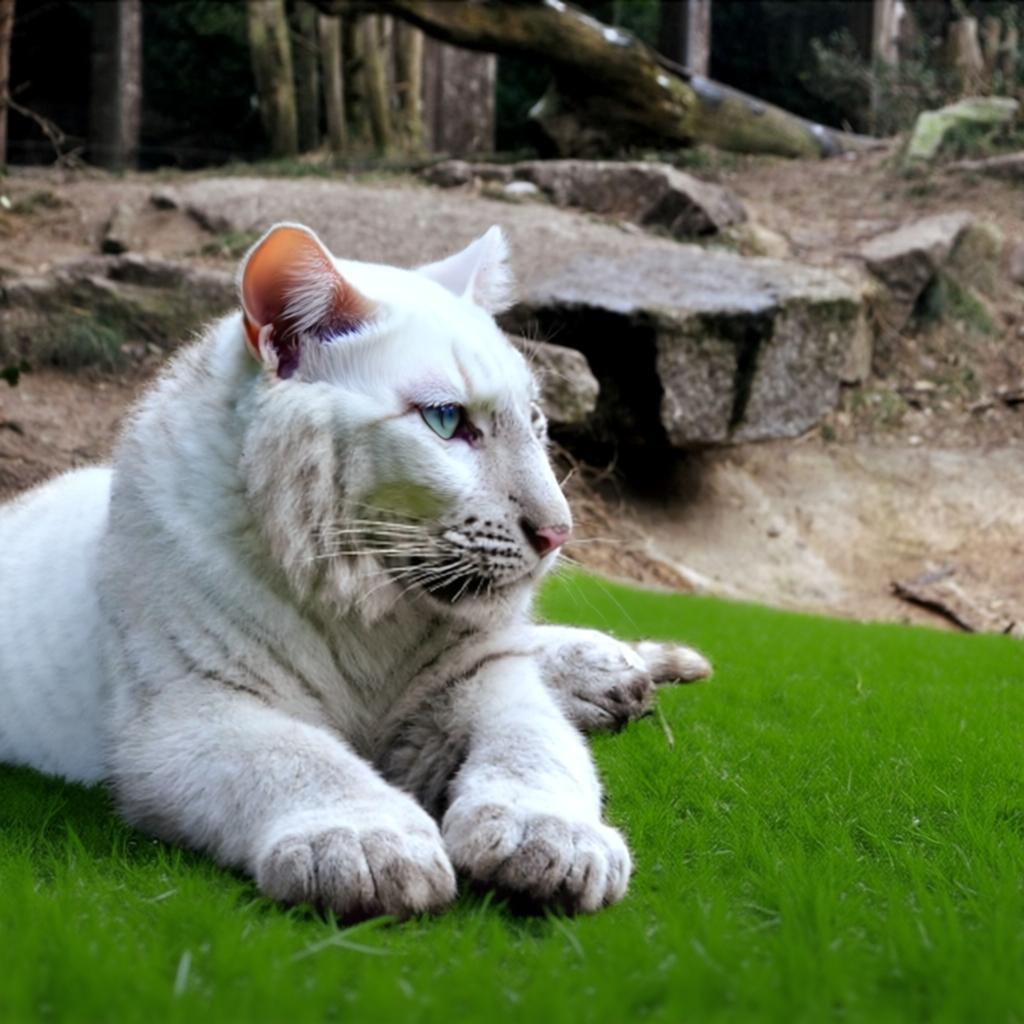} &
    \includegraphics[width=0.15\linewidth]{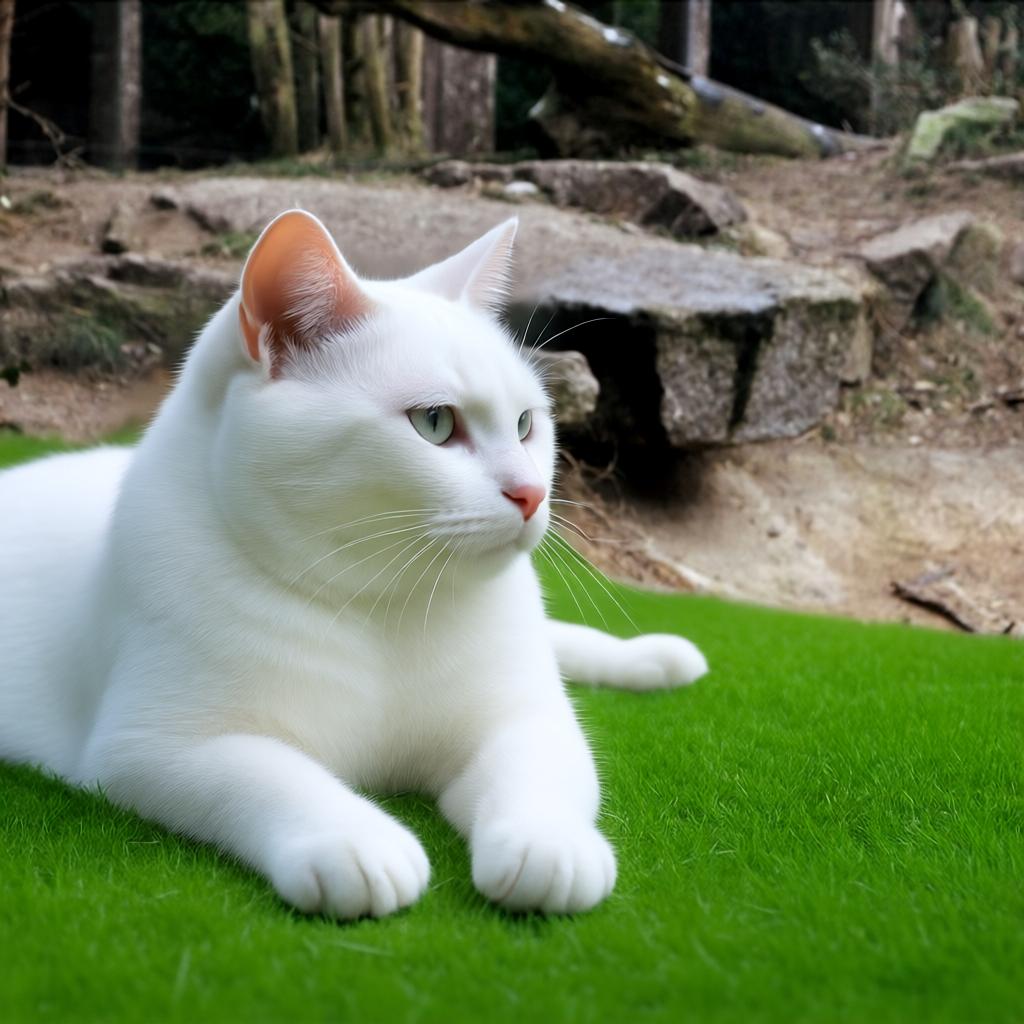} \\

    \multicolumn{5}{c}{\textit{``\input{supp/figures/qwen-lora/23/instruction.txt}''}} \\
    \includegraphics[width=0.15\linewidth]{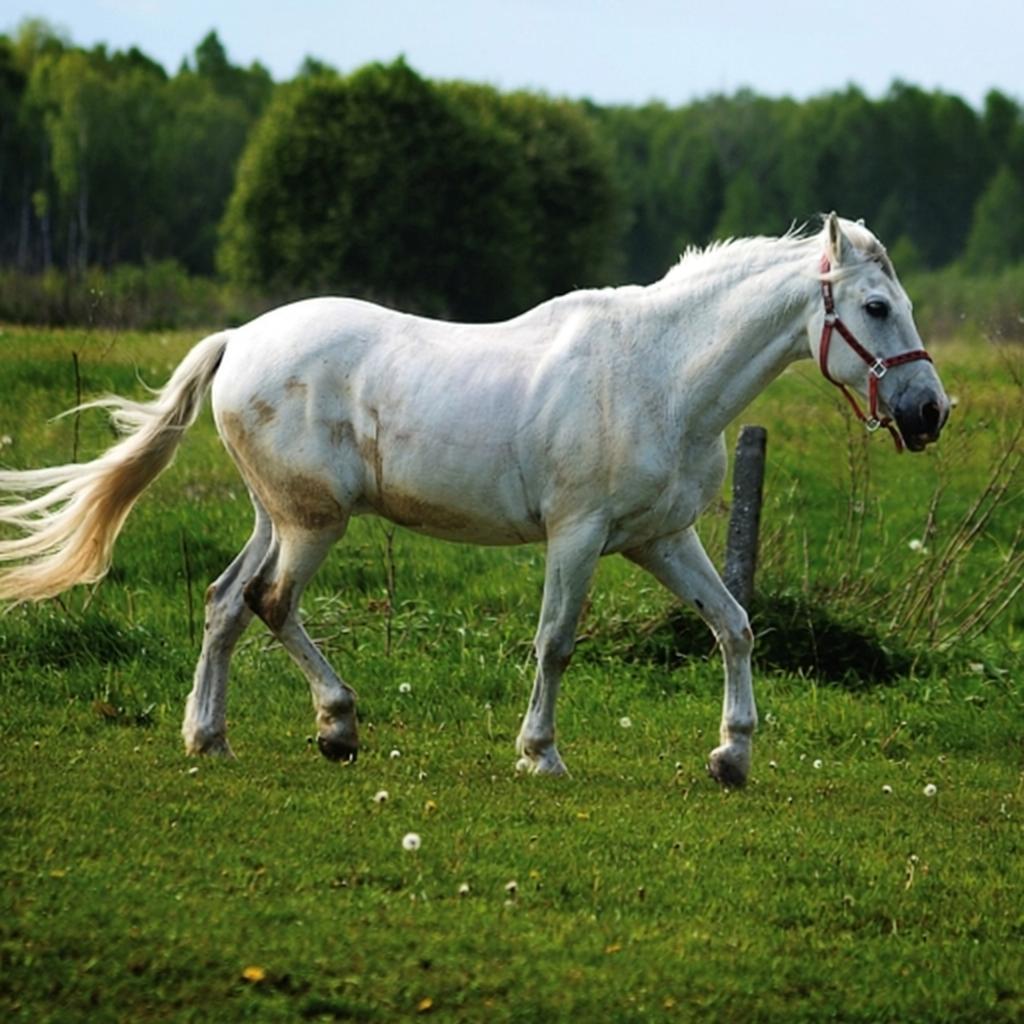} &
    \includegraphics[width=0.15\linewidth]{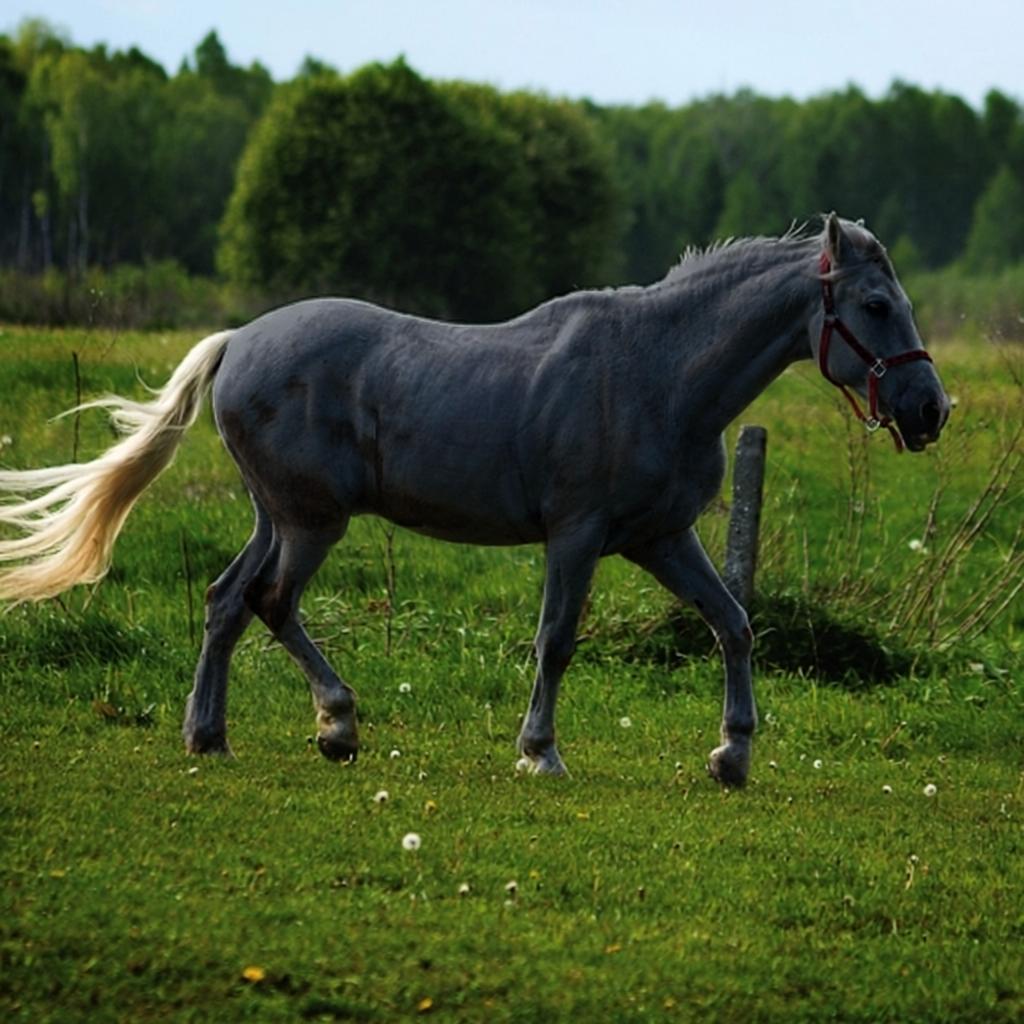} &
    \includegraphics[width=0.15\linewidth]{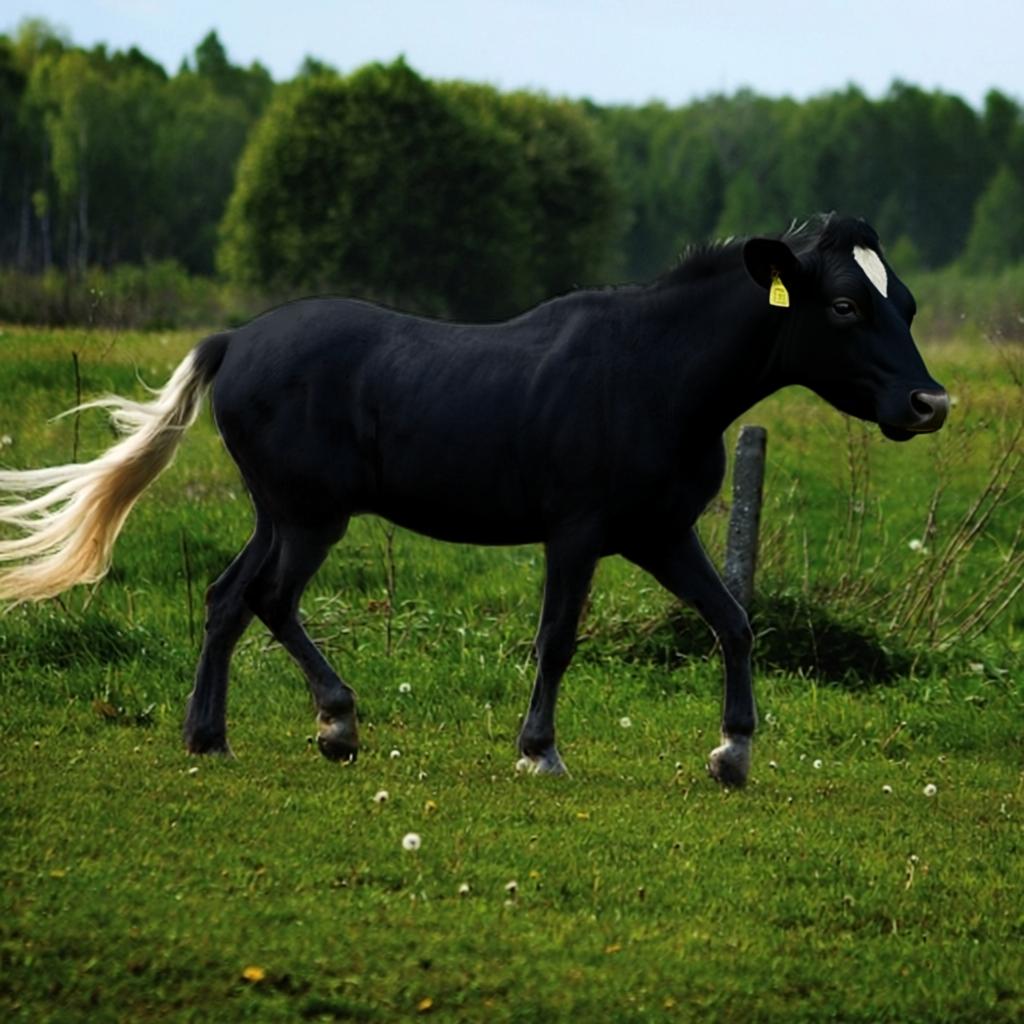} &
    \includegraphics[width=0.15\linewidth]{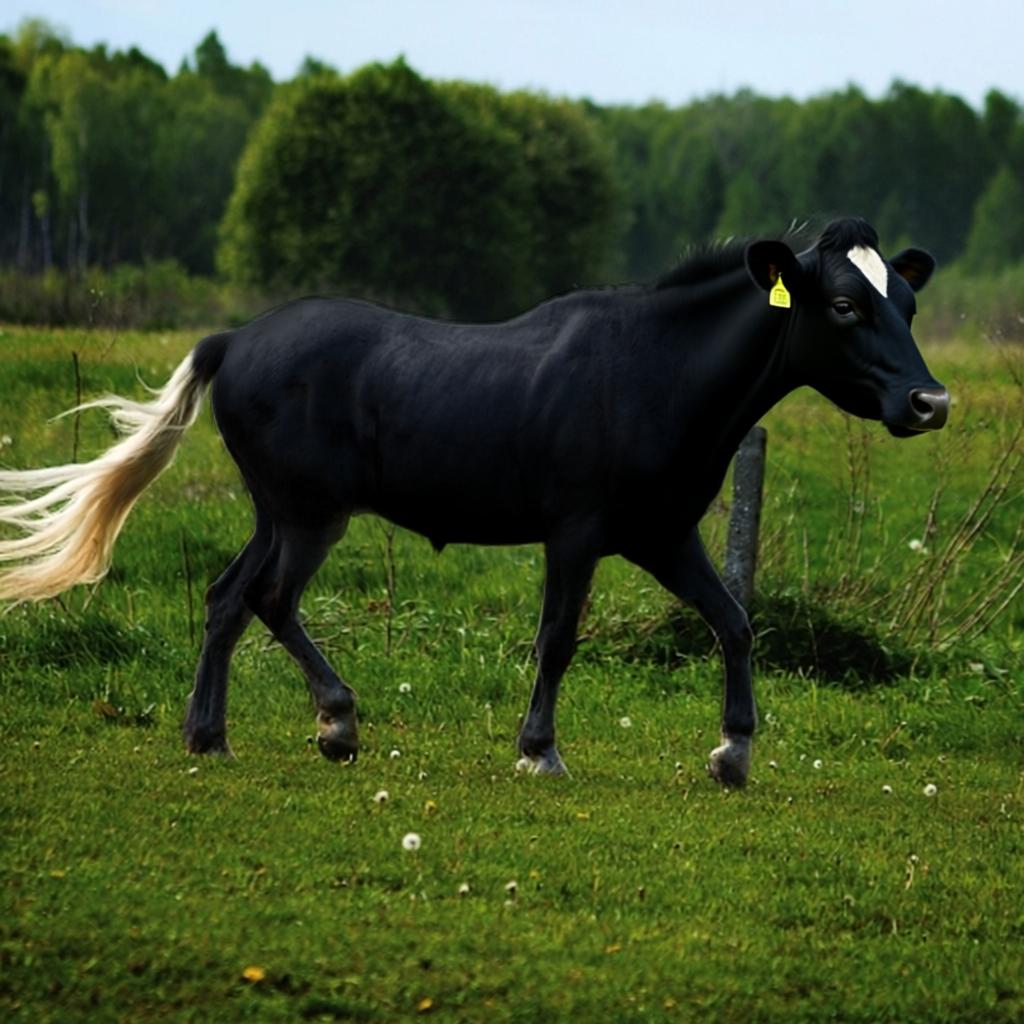} &
    \includegraphics[width=0.15\linewidth]{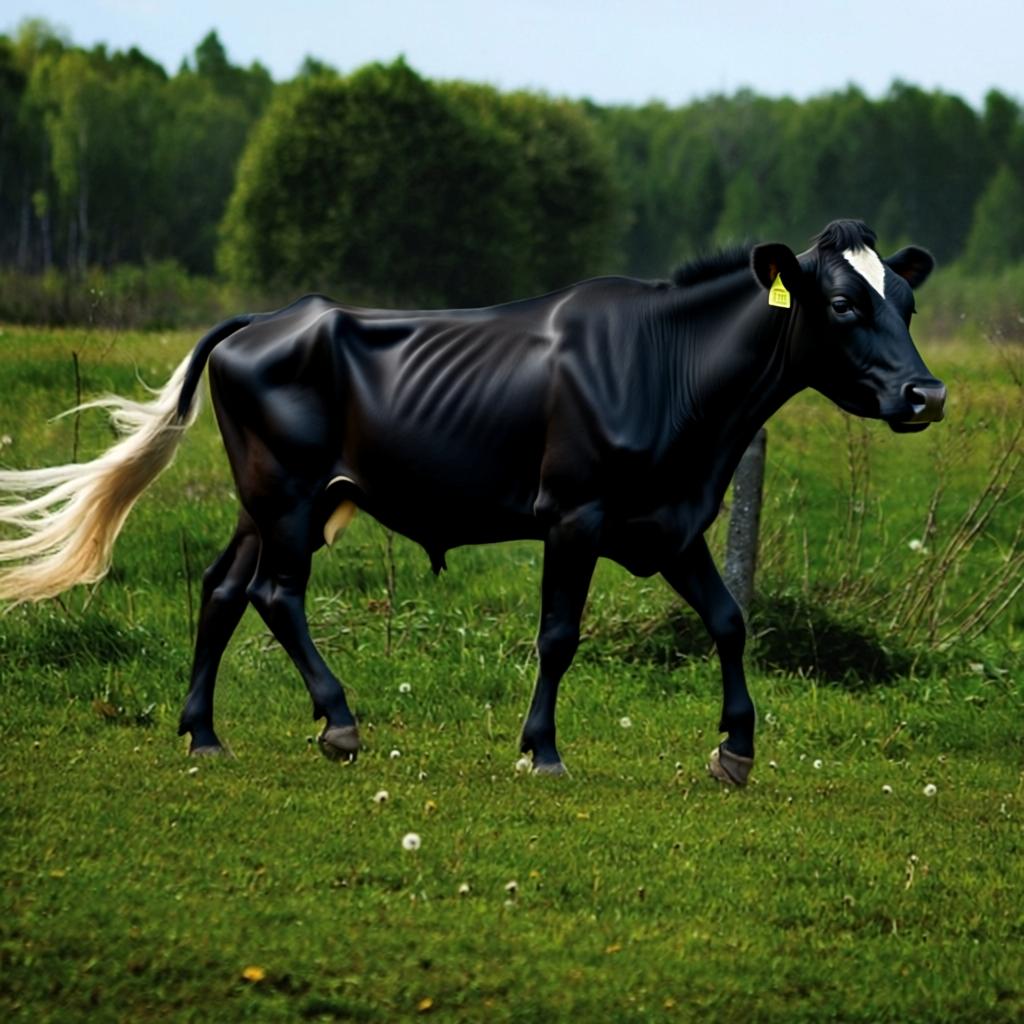} \\

        \multicolumn{5}{l}{Edit Intensity $\xrightarrow{\hspace{320pt}}$} \\
        
    \end{tabular}
    \caption{\textbf{Qualitative results with Qwen-Image-Edit as backbone.} 
    }
    \label{fig:qwen-lora-qualitative}
\end{figure*}

%% file: supp/figures/qwen-lora/19/instruction.txt
Change the white tiger into a white cat and change the brown ground to green grass.

%% file: supp/figures/qwen-lora/23/instruction.txt
Transform the gray horse into a black cow.

%% file: supp/figures/analytical_supp/analytical_supp.tex
\begin{figure*}
    \centering
    \setlength{\tabcolsep}{0pt} %
    
    \begin{tabular}{cc cccccc}
        \multicolumn{8}{c}{\textit{``Change the frosting to strawberry''}} \\

        \raisebox{28pt}{\rotatebox[origin=t]{90}{{AdaOr (Ours)}}} & { } &
        \includegraphics[width=0.14\linewidth]{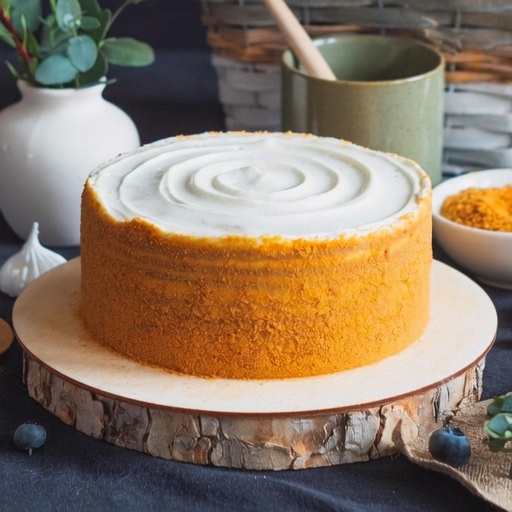} &
        \includegraphics[width=0.14\linewidth]{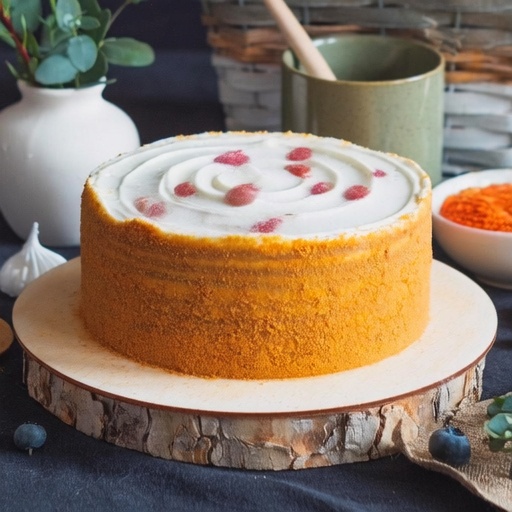} &
        \includegraphics[width=0.14\linewidth]{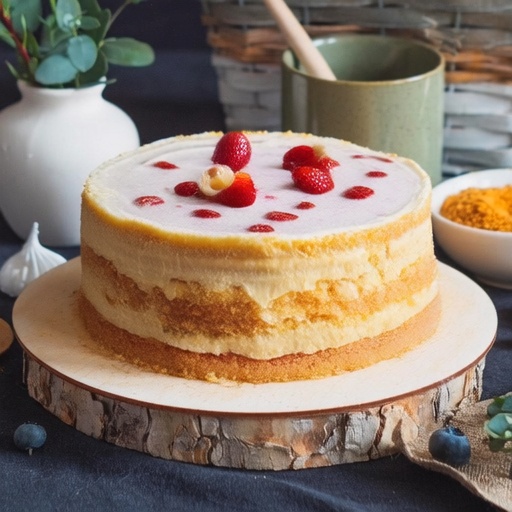} &
        \includegraphics[width=0.14\linewidth]{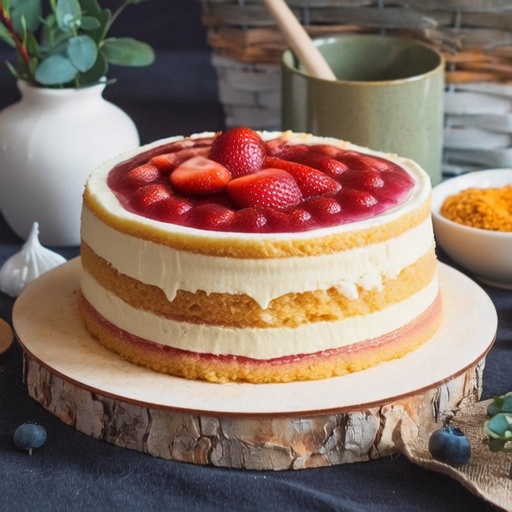} &
        \includegraphics[width=0.14\linewidth]{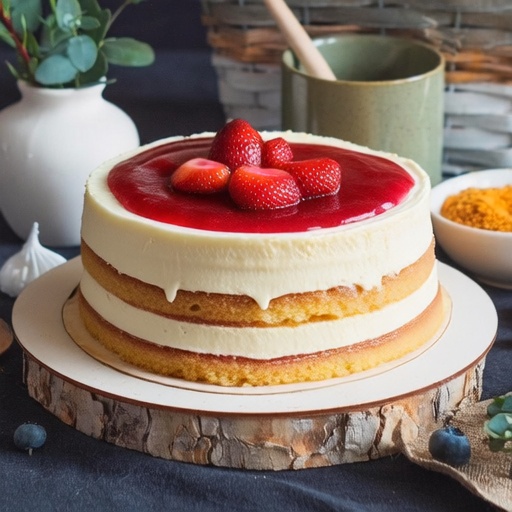} &
        \includegraphics[width=0.14\linewidth]{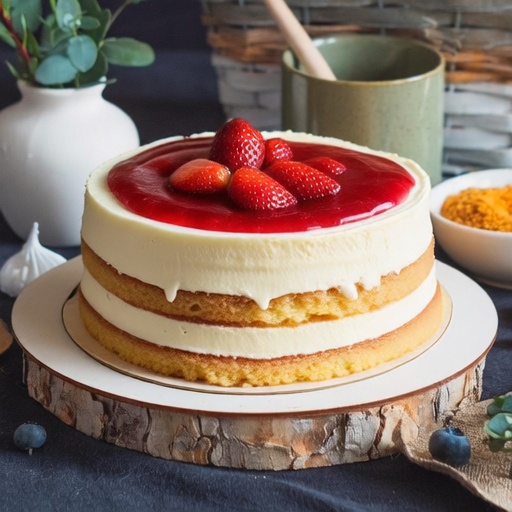} \\

        \raisebox{35pt}{\rotatebox[origin=t]{90}{{Analytic \REC}}} & { } &
        \includegraphics[width=0.14\linewidth]{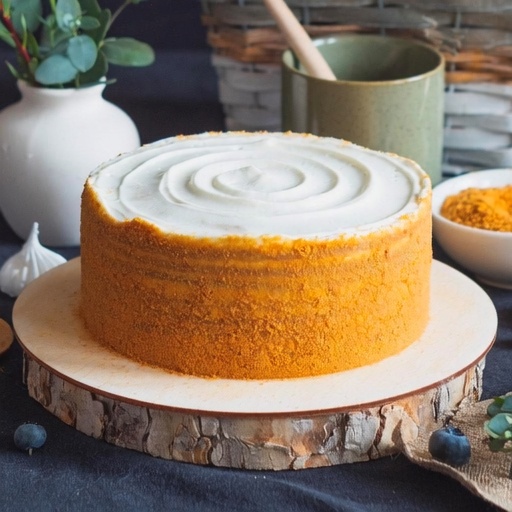} &
        \includegraphics[width=0.14\linewidth]{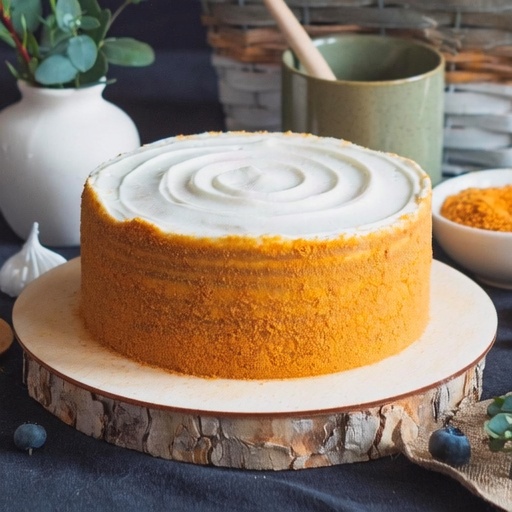} &
        \includegraphics[width=0.14\linewidth]{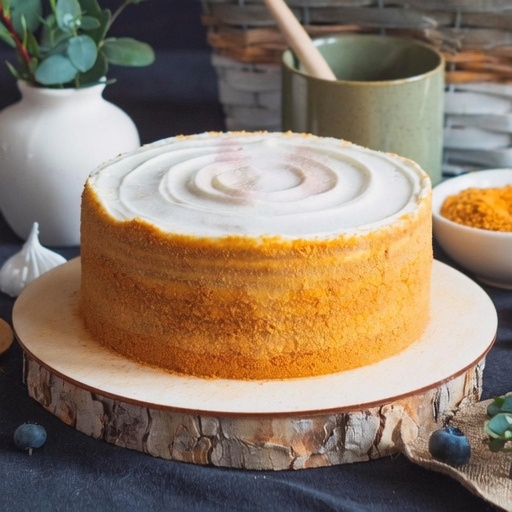} &
        \includegraphics[width=0.14\linewidth]{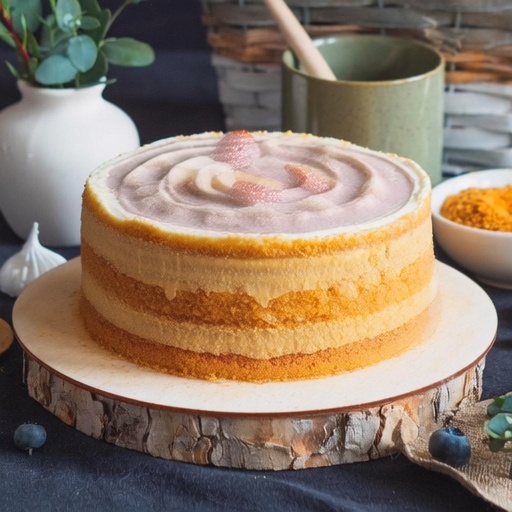} &
        \includegraphics[width=0.14\linewidth]{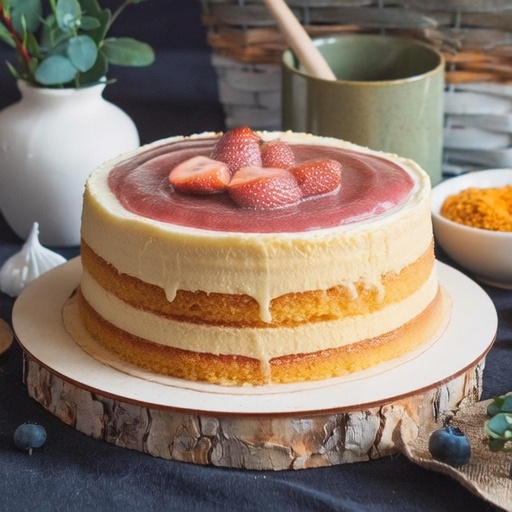} &
        \includegraphics[width=0.14\linewidth]{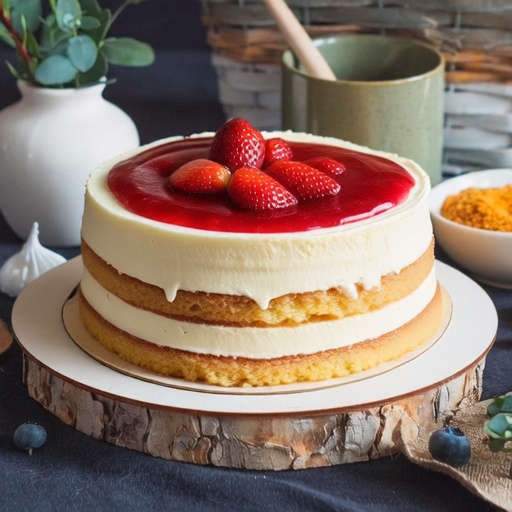} \\

        & \multicolumn{7}{l}{Edit Intensity $\xrightarrow{\hspace{370pt}}$} \\
        
    \end{tabular}
    \caption{\textbf{Qualitative results comparing analytical \REC{} prediction with our learned \REC.} Using analytical \REC{} prediction results in off-manifold intermediate images.
    }
    \label{fig:analytical_supp}
\end{figure*}